\documentclass[10pt,journal,compsoc]{IEEEtran}
\usepackage[T1]{fontenc}
\usepackage{amsmath,amssymb,amsthm,mathtools}
\usepackage{graphicx,booktabs,array,tabularx,microtype}
\usepackage{xcolor,pgfplots}
\pgfplotsset{compat=1.18}
\usepackage[hidelinks]{hyperref}
\usepackage{cite}
\usepackage{enumitem}
\hypersetup{pdftitle={Spectral Thresholds and Observation Masks for Factor Agreement in Low-Rank PSD Learning},pdfauthor={Enliang Hu},pdfsubject={Exactness of asymmetric low-rank PSD factorization},pdfkeywords={symmetric matrix factorization, penalty exactness, alternating minimization}}
\newcounter{algorithm}
\makeatletter
\def\fps@algorithm{tbp}
\def\ftype@algorithm{4}
\def\ext@algorithm{loa}
\def\fnum@algorithm{Algorithm~\thealgorithm}
\newenvironment{algorithm}{\@float{algorithm}}{\end@float}
\makeatother
\newtheorem{theorem}{Theorem}
\newtheorem{lemma}[theorem]{Lemma}
\newtheorem{proposition}[theorem]{Proposition}
\newtheorem{corollary}[theorem]{Corollary}
\newtheorem{assumption}{Assumption}
\theoremstyle{remark}

\DeclareMathOperator{\sym}{sym}
\DeclareMathOperator{\skewop}{skew}

\DeclareMathOperator{\dom}{dom}
\DeclareMathOperator{\dist}{dist}

\allowdisplaybreaks[1]
\title{Spectral Thresholds and Observation Masks for Factor Agreement in Low-Rank PSD Learning}
\author{Enliang Hu\thanks{Email: \texttt{ynel.hu@gmail.com}.}}
\begin{document}
\maketitle
\begin{abstract}
A quadratic penalty encourages the two factors in an asymmetric low-rank positive semidefinite (PSD) model to agree. We study when agreement holds at every joint critical point and how to bound disagreement at an approximate solution. For a transpose-symmetric convex loss and a shared convex regularizer, we bound the distance between the factors by the difference of their stationarity residuals divided by a spectral margin. For a strongly convex loss, the transpose difference strictly improves the pointwise threshold exactly when the midpoint has full column rank and its range has trivial intersection with the top eigenspace of the symmetric part of the selected loss subgradient. For fully observed least squares with unrestricted factors and a PSD target, every positive penalty enforces agreement. A symmetric binary mask preserves this property for every PSD target exactly when its observed entries form disjoint fully observed principal blocks. We also give a curvature bound for smooth nonconvex losses and an absolute-loss example in which an exact one-sided update stops at a point of strict symmetric descent. For fixed positive penalty, sufficient decrease and relative block residual control give whole-sequence convergence under the stated boundedness, regularity and Kurdyka--\L{}ojasiewicz assumptions. Experiments on masked Gram matrices measure the information lost by a one-sided update and the cost of solving its inner systems.

\end{abstract}

\section{Introduction}
\label{sec:introduction}
Positive semidefinite (PSD) representations encode kernels, similarities, and learned metrics. Their usefulness in pattern analysis is limited by the cost of storing and optimizing a dense matrix. Low-rank factorization replaces a matrix variable by $XX^\top$, reducing storage from quadratic to linear in the number of objects when the factor rank is small. This reparameterization also makes the optimization nonconvex, even when the loss is convex in the matrix.

Asymmetric factorization introduces a second factor and solves
\begin{equation}
\label{eq:main-split}
F_\gamma(X,Y)
=f(XY^\top)+\tfrac12h(X)+\tfrac12h(Y)
+\tfrac{\gamma}{2}\|X-Y\|_F^2.
\end{equation}
For $\gamma\ge0$ and convex $f,h$, fixing either factor gives a convex problem. The discrepancy penalty is intended to recover the symmetric objective
\begin{equation}
\label{eq:main-symmetric}
\Phi(X)=f(XX^\top)+h(X).
\end{equation}
In the coordinates $Y=X+\Delta$, zero displacement gives $F_\gamma(X,X)=\Phi(X)$. This identity raises two questions. At a computed pair, how close is the midpoint to stationarity for $\Phi$? At a symmetric pair, does a zero one-sided step imply that no common-factor direction decreases $\Phi$?

Quadratic splitting has been used in symmetric NMF and low-rank matrix learning~\cite{snmf-nips,snmf-tkde,hu2019,han2026lorads}. Hu et al.~\cite{hu2026abmf} derived a sufficient spectral condition for agreement and a penalty bound that depends on the initialization. Their rank-two SymNMF example shows that a positive penalty alone can leave the factors unequal (Appendix~\ref{app:symnmf-counterexample}).

Our first result retains the curvature between $XY^\top$ and $YX^\top$ in the block stationarity equations. Theorem~\ref{thm:geometry} characterizes strict improvement of the spectral bound and also controls disagreement at a pair with nonzero residuals. The bound depends on the midpoint and the loss subgradient at that pair. Theorem~\ref{thm:signed-curv} extends the calculation to signed curvature. Without a restriction on the region, even a coercive smooth loss can admit unequal critical factors at every positive penalty (Proposition~\ref{prop:no-global-gamma}).

Partial observation changes the answer even when the target is PSD. A principal submatrix of a PSD kernel is PSD, but an entrywise mask can produce an indefinite matrix $W\odot A$. Its smallest eigenvalue determines exactly when a nonzero pair $(P,-P)$ can be critical. This observation yields the classification of binary masks in Theorem~\ref{thm:mask-universal} and the trace penalty that eliminates this family in Corollary~\ref{cor:hinge-branch}. At a specified midpoint the weighted least-squares objective is an even quartic in the factor difference. Theorem~\ref{thm:fiber} uses its coefficients to characterize agreement on that slice, also when the factors must be nonnegative.

The second question first arises with nonsmooth losses. For an entrywise absolute loss on a two-by-two positive definite matrix, the exact one-sided update is zero although moving both factors together decreases the loss to first order. Section~\ref{sec:zero-step} identifies the observed pairs responsible for this discrepancy and gives a test at a stalled iterate. For a smooth loss, Section~\ref{sec:kl-fixed} gives a fixed-penalty convergence proof with relative tolerances on the block solves. Appendix~\ref{sec:adaptive} treats a separate stagewise algorithm that checks the residuals after each penalty trial.

The experiments use digit and wine-feature Gram matrices to test when the one-sided and symmetric models choose different updates. On sparse digit similarities we also compare the cost of direct and iterative row solves at a common midpoint-gradient tolerance.

\section{Related Work}
\label{sec:related}

\paragraph{Exact penalties for symmetric factorization.}
Low-rank PSD parametrization and direct SymNMF provide the original single-factor models~\cite{Burer05,snmf-gopt}. Zhu et al.~\cite{snmf-nips} and the journal treatment by Li et al.~\cite{snmf-tkde} exploit a penalized two-factor formulation of SymNMF and obtain convergence to symmetric critical points under sufficient penalization. Hu and Kwok~\cite{hu2019} use a quadratic factor-difference penalty and biconvex surrogates for low-rank matrix learning. Hu et al.~\cite{hu2026abmf} give the general transpose-symmetric convex-loss spectral criterion and the counterexample to exactness for every positive penalty in unrestricted choices of SymNMF input. Yang et al.~\cite{yang2026gsmf} establish finite-penalty exactness for generalized symmetric factorization under their conditions and develop nonmonotone inexact alternating updates with a KL analysis. Liu et al.~\cite{liu2026masked} obtain sufficient penalization for masked SymNMF on incomplete networks. Our mask theorem instead quantifies over every PSD input and every positive penalty for a specified mask and unrestricted factors. At a given pair, Corollary~\ref{cor:radial-margin} converts a signed curvature margin into a bound on how fast the split objective decreases when the factors move together.

\paragraph{Descent directions for the symmetric objective.}
First-order surrogate models and successive approximation use agreement at the base point to construct directions~\cite{mairal2013surrogates,yang2017successive}. Ochs and Malitsky~\cite{ochs2019model} combine model improvement with an Armijo-type search. Prox-linear methods linearize the inner product map in composite low-rank recovery~\cite{charisopoulos2021composite}. These are antecedents for the acceptance test in Section~\ref{sec:zero-step}. Gradient sampling also seeks robust descent by pooling nearby gradient information~\cite{burke2005sampling}. Our tolerance-band test uses the signs of observed residuals, with a finite-step bound determined by the first possible sign change outside the band. A one-sided factor update perturbs $VV^\top$ by $HV^\top$; a common-factor update perturbs it by $HV^\top+VH^\top$. At a nonsmooth loss the two models can have different slopes even at exactly equal factors. Tseng~\cite{tseng2001bcd} identifies regularity conditions under which coordinatewise minimization implies stationarity. For the transpose-symmetric factorization we identify the two projected subdifferentials that determine every lost slope. Moutier et al.~\cite{moutier2020off} study off-diagonal SymNMF with an entrywise absolute loss and direct coordinate updates. Our result for that loss family locates the discrepancy on exactly fitted observed pairs and tests a zero step using shared symmetric subgradients. At finite displacement Proposition~\ref{prop:absolute-ball} characterizes the largest ball of identical models.

\paragraph{Gauss--Newton and Levenberg--Marquardt updates.}
Liu et al.~\cite{liu2015gn} derive inexpensive Gauss--Newton updates for symmetric low-rank product approximation. GNMR linearizes both factors jointly in matrix sensing and completion~\cite{zilber2022gnmr}. For full-observation squared loss the unconstrained one-sided half-step coincides with the unprojected inexact LM direction in Behling et al.~\cite[Eq.~(7)]{behling2026pilm}, who prove descent for that direction and use safeguards after projection. Inverse-Gram preconditioning is also well established~\cite{tong2021scaledgd,xu2023scaled}. Appendix~\ref{sec:curvature} identifies the additional skew curvature induced by an extension off the symmetric matrices and the observation patterns under which the resulting linear system separates across rows. The directional comparison of Section~\ref{sec:zero-step} also covers losses with no least-squares residual to linearize.

\paragraph{Stationarity of the factor midpoint.}
Han et al.~\cite{han2026lorads} transfer approximate KKT conditions for a low-rank ADMM splitting to the midpoint. Their smooth analysis already contains a term quadratic in the disagreement. Our finite-output bounds use the difference of the two block residuals to bound disagreement under a positive margin and use their sum to recover symmetric stationarity. For nonsmooth shared regularizers the latter step uses approximate subgradients or a proximal-gradient mapping. Gram balance for ridge-regularized low-rank factorization is already known~\cite{li2019nonconvexgeometry}; Appendix~\ref{sec:unregularized} uses it to study equality of the factors. Direct SymNMF coordinate descent also uses univariate quartics~\cite{vandaele2016coordinate}. In Section~\ref{sec:fiber} the scalar variable scales a given factor difference at fixed midpoint. The quartic coefficients across all such directions characterize the feasible slice. Standard hinge proximal formulas~\cite{parikh2014proximal} enter the implementation of the trace correction.

\paragraph{Convergence of alternating updates.}
Sufficient decrease and a relative subgradient error lead to KL convergence for a fixed objective~\cite{abs2013framework,Frankel2014SplittingMW}, including block and proximal alternating schemes~\cite{xu13,bolte2014palm}. Section~\ref{sec:kl-fixed} verifies these hypotheses for inexact alternating updates with a common endpoint residual. Its limit solves the split stationarity conditions. A positive exactness margin provides the further conclusion of agreement. Search along a certified direction on the original objective and changes to the penalty define different sequences. The warm-started spectral estimate is a practical proposal; the variable-penalty acceptance conditions appear separately in Appendix~\ref{sec:adaptive}.

\section{Model and Spectral Exactness}
\label{sec:preliminaries}
Let $X,Y\in\mathbb R^{n\times r}$. We use the Frobenius inner product, write $\operatorname{sym}Z=(Z+Z^\top)/2$ and $\operatorname{skew}Z=(Z-Z^\top)/2$, and let $\|\cdot\|_2$ denote the matrix spectral norm. A finite loss is \emph{transpose symmetric} if $f(Z)=f(Z^\top)$ for every square $Z$. For a differentiable loss this gives $\nabla f(Z^\top)=\nabla f(Z)^\top$; for a finite convex loss the analogous identity holds for its subdifferential.

Equivalently, in anchor and displacement coordinates $Y=X+\Delta$ the split objective is
\[
\begin{aligned}
 \widehat F_\gamma(X,\Delta)={}&f(XX^\top+X\Delta^\top)+\tfrac\gamma2\|\Delta\|_F^2\\
 &+\tfrac12[h(X)+h(X+\Delta)].
\end{aligned}
\]
It satisfies $\widehat F_\gamma(X,0)=\Phi(X)$ and $\widehat F_\gamma(X,\Delta)=\widehat F_\gamma(X+\Delta,-\Delta)$ by transpose symmetry. We use $(X,Y)$ for separate endpoint residuals and a fixed midpoint for joint contractions.

In this section the shared regularizer $h$ is proper, closed, and convex, and may include an indicator of a closed convex set. Ordinary convexity corresponds to strong-convexity modulus zero. A joint critical point of \eqref{eq:main-split} is a pair for which a common $G\in\partial f(XY^\top)$ and $H_X\in\partial h(X)$, $H_Y\in\partial h(Y)$ satisfy
\begin{equation}
\label{eq:joint-critical}
GY+\gamma(X-Y)+H_X/2=0,\quad
G^\top X-\gamma(X-Y)+H_Y/2=0.
\end{equation}
Unrelated loss-subgradient selections in the two equations do not establish joint criticality. For smooth losses there is no ambiguity.

At a given pair define $U=(X+Y)/2$, $D=X-Y$, $Z=XY^\top$, and $S=\operatorname{sym}G$. Let $e_X,e_Y$ denote the left sides of \eqref{eq:joint-critical} at that same pair and write $e_-=e_X-e_Y$. For a factor-shaped test matrix $H$, set
\[
 L_UH=HU^\top-UH^\top,\qquad \mathcal S[H]=SH.
\]
The identity $Z-Z^\top=L_UD$ expresses the transpose difference in terms of the factor disagreement. The adjoint $L_U^*$ and the following operator eigenvalue use Frobenius inner products.

\begin{theorem}[When transpose curvature lowers the penalty threshold]
\label{thm:geometry}
Let $f$ be finite, transpose symmetric and $\sigma_f$-strongly convex, and let $h$ be proper, closed and $\sigma_h$-strongly convex, with $\sigma_f,\sigma_h\ge0$. For a common $G\in\partial f(Z)$ and selections $H_X\in\partial h(X)$, $H_Y\in\partial h(Y)$, put
\begin{equation}
 \kappa_U=\lambda_{\max}\bigl(\mathcal S-
                       \tfrac{\sigma_f}{2}L_U^*L_U\bigr).
 \label{eq:allrank-threshold}
\end{equation}
If $m=2\gamma+\sigma_h/2-\kappa_U>0$, then
\begin{equation*}
 \|X-Y\|_F\le \frac{\|e_X-e_Y\|_F}{m}.
\end{equation*}
In particular, a jointly critical pair with this margin has equal factors, and their common value is critical for \eqref{eq:main-symmetric}.

One always has $\kappa_U\le\lambda_{\max}(S)$. For $\sigma_f>0$, let $E$ denote the entire top eigenspace of $S$. Then $\kappa_U<\lambda_{\max}(S)$ if and only if
\begin{equation}
 \operatorname{rank}(U)=r
 \quad\text{and}\quad E\cap\operatorname{range}(U)=\{0\}.
 \label{eq:allrank-strict}
\end{equation}
\end{theorem}

To see where the improvement comes from, subtract the two block residuals and pair with $D$. Strong monotonicity at $Z,Z^\top$ and at $X,Y$ gives
\begin{equation*}
\begin{aligned}
 \langle D,e_-\rangle\ \ge\ &
 (2\gamma+\sigma_h/2)\|D\|_F^2-\langle D,SD\rangle\\
 &+\tfrac{\sigma_f}{2}\|L_UD\|_F^2.
\end{aligned}
\end{equation*}
The Rayleigh bound gives the stated margin. Strict improvement requires the correction to be positive on every nonzero factor direction whose columns lie in $E$. Appendix~\ref{app:proof-spectral-improvement} characterizes this nullspace and proves \eqref{eq:allrank-strict}.

\paragraph{Comparison of penalty thresholds.}
Dropping the last nonnegative term at a joint critical pair recovers the sufficient condition of Hu et al.~\cite{hu2026abmf},
\begin{equation}
\label{eq:spectral-margin}
2\gamma+\sigma_h/2>\lambda_{\max}(S).
\end{equation}
At the evaluated pair, Theorem~\ref{thm:geometry} reduces the right-hand side of the sufficient inequality for $\gamma$ by $(\lambda_{\max}(S)-\kappa_U)/2$. When $\sigma_f=0$ the two spectral tests coincide. A masked loss may still have curvature on its observed entries; Appendix~\ref{sec:consensus-geometry} gives the corresponding operator, a lower-dimensional spectral estimate, and an example. A penalty valid on an entire region requires control of the margin throughout that region. Section~\ref{sec:framework} treats signed curvature and initialization-dependent bounds.

For exact alternating minimization, convex blocks and a positive penalty give sufficient descent~\cite{hu2026abmf}. Sections~\ref{sec:delta-solve} and~\ref{sec:kl-fixed} treat inexact updates at fixed penalty. Missing observations can leave factor sublevels unbounded, so Appendix~\ref{sec:adaptive} states a separate level-set bound for penalty changes.

\section{Problem Class and General Penalty Theory}
\label{sec:framework}
Both factors have width $r$, the same feasible domain and the same regularizer. The data loss is specified on square matrices, including nonsymmetric products, and obeys $f(Z^\top)=f(Z)$. This covers full or symmetrically masked similarity fitting, symmetric linear measurements with a specified transpose-invariant loss, and nonnegative factors via $h=\delta_+$. An asymmetric observation protocol needs a justified symmetric extension. A rectangular target, two distinct factor domains, or a changing outer multiplier requires additional modeling. The required regularity depends on the conclusion.

\begin{itemize}
\item For the convex spectral certificate, $f$ is finite convex and transpose symmetric, and $h$ is proper closed convex. A nonsmooth loss uses the common selection in \eqref{eq:joint-critical}.
\item For the signed-curvature certificate below, $f$ is differentiable and can be nonconvex. It suffices to control curvature on transpose pairs, and the subgradient pairing of $h$ on factor pairs. Weak convexity is one sufficient condition.
\item For the fixed-objective convergence result, the smooth coupling has locally Lipschitz gradient, and the composite objective has the KL property on its cluster set. The theorem allows additional nonconvex penalties under its explicit value-continuity and update conditions. Stationarity, agreement, and global optimality are different conclusions.
\end{itemize}
Examples covered by the latter two sets of assumptions include smooth robust losses on symmetric measurements, a weakly convex sparsity penalty, and a smooth double-well factor penalty. The one-block formulas below require a tractable proximal map or a suitable inner solver. For the nonsmooth joint loss in the first level, two independent block subgradients cannot be substituted into the smooth-coupling convergence proof.

\subsection{A signed curvature margin}
Write $D=X-Y$, $Z=XY^\top$, $B=Z-Z^\top$, $U=(X+Y)/2$, $S=\sym G$ and $G=\nabla f(Z)$, with $f$ continuously differentiable and transpose symmetric. For $H_X\in\partial h(X)$, $H_Y\in\partial h(Y)$, denote by $e_X,e_Y$ the left sides of \eqref{eq:joint-critical}; $\partial$ is the limiting subdifferential when $h$ is nonconvex. Throughout this section $h$ is proper and lower semicontinuous. Use constants $\mu_T,\mu_h\in\mathbb R$ such that, at the pairs under consideration,
\begin{align}
 \langle G-G^\top,B\rangle&\ge\mu_T\|B\|_F^2,\nonumber\\
 \langle H_X-H_Y,D\rangle&\ge\mu_h\|D\|_F^2.\label{eq:signed-pair}
\end{align}
For a uniform result require these inequalities for all relevant pairs and selections. For example $h-\mu_h\|\cdot\|_F^2/2$ convex implies the second inequality, also for $\mu_h<0$. Differentiable $\rho_f$-weakly convex losses admit $\mu_T=-\rho_f$, by convexifying $f$ and comparing $Z,Z^\top$. Often much larger $\mu_T$ is available.

\begin{theorem}[Penalty and disagreement with signed curvature]\label{thm:signed-curv}
Under \eqref{eq:signed-pair}, set $d=\|D\|_F$. When $d>0$, let
\[
 q=\langle D,SD\rangle/d^2,\qquad
 m=2\gamma+\mu_h/2-q+\mu_T\|B\|_F^2/(2d^2).
\]
Then $m d^2\le\langle D,e_X-e_Y\rangle$. Positive $m$ implies $d\le\|e_X-e_Y\|_F/m$. If $\|U\|_2\le R$ and $\lambda_{\max}(S)\le M$ throughout a region, every critical pair there has equal factors whenever
\begin{equation}
 2\gamma+\mu_h/2>M+2[-\mu_T]_+R^2 .
 \label{eq:uniform-weak}
\end{equation}
Every diagonal joint critical pair $(U,U)$ satisfies $0\in\partial\Phi(U)$, including for proper lower semicontinuous nonconvex $h$.
\end{theorem}
\begin{proof}
Expanding both residuals gives the identity
\begin{align*}
 \langle D,e_X-e_Y\rangle
 &=2\gamma d^2-\langle D,SD\rangle\\
 &\quad+\tfrac12\langle G-G^\top,B\rangle
       +\tfrac12\langle H_X-H_Y,D\rangle.
\end{align*}
Apply \eqref{eq:signed-pair} and Cauchy--Schwarz. Since $B=DU^\top-UD^\top$, $\|B\|_F\le2R d$. For negative $\mu_T$ this gives a lower bound for its contribution; for nonnegative $\mu_T$ one can omit the term. At a diagonal critical pair, $G=\nabla f(UU^\top)$ is symmetric. Either block equation gives $-2GU\in\partial h(U)$ (indeed $H_X=H_Y=-2GU$). The smooth sum rule gives $\partial\Phi(U)=2GU+\partial h(U)$, proving stationarity.
\end{proof}
This diagonal argument uses the uniqueness of the loss gradient. For a convex nonsmooth loss, the residual identity still holds for a common selection $G$, since $G^\top\in\partial f(Z^\top)$. At agreement Theorem~\ref{thm:geometry} uses convexity of $h$ to average the endpoint selections.

\begin{corollary}[Descent toward agreement at fixed midpoint]
\label{cor:radial-margin}
Let $f$ be continuously differentiable and transpose symmetric. Suppose $h-\mu_h\|\cdot\|_F^2/2$ is proper closed convex, with $\mu_h$ of either sign. Use $m$ from Theorem~\ref{thm:signed-curv}, with the stated transpose-pair bound at the current feasible pair $X\ne Y$. Set $X_t=X-tD/2$, $Y_t=Y+tD/2$, and $E(t)=F_\gamma(X_t,Y_t)$ for $0\le t\le1$. Then
\[
 E'(0+)\le -\tfrac12 m\|D\|_F^2.
\]
In particular a positive margin certifies strict local descent toward agreement. If the margin is at least $m_0>0$ all along $0\le t<1$, then
\begin{equation}
 F_\gamma(X,Y)-\Phi(U)\ge\tfrac14 m_0\|D\|_F^2 .
 \label{eq:radial-integral}
\end{equation}
\end{corollary}
\begin{proof}
The derivative of the smooth part along $(-D/2,D/2)$ is minus one half of its gradient-difference pairing in Theorem~\ref{thm:signed-curv}. For the regularizer, the two weak-convexity chord bounds in directions $-D$ at $X$ and $D$ at $Y$ sum to at most $-\mu_h\|D\|_F^2$. Their coefficient in $E'(0+)$ is $1/4$. Together these estimates give the first inequality even when endpoint normals are unavailable. At time $t<1$ the disagreement is $(1-t)D$. Converting its inward direction back to the time variable gives $E'(t+)\le-m_0(1-t)\|D\|_F^2/2$. Integrate on compact subintervals of $(0,1)$ and pass to the endpoints. The restricted convexified regularizer is finite and continuous on this segment by convexity and closedness.
\end{proof}
Holding $X$ fixed changes the sign test. For $f(z)=(z-2)^2/2$, $h=0$, $\gamma=1$, $X=1$ and $Y=5/4$, moving $Y$ toward $X$ with velocity $X-Y$ has derivative $+1/8$. Moving both factors toward their midpoint has derivative $-11/128$. The target is positive and the pointwise margin is $11/4$.

The first conclusion tests a short simultaneous move of both factors. A full replacement by the midpoint requires control over the segment, as illustrated by the exact squared-loss identity in Section~\ref{sec:fiber}.

\begin{corollary}[A penalty selected before the first sweep]\label{cor:initial-smooth}
Assume the hypotheses of Theorem~\ref{thm:signed-curv}, with \eqref{eq:signed-pair} holding uniformly on the initial sublevel. Let $X_0=Y_0$, $E_0=\Phi(X_0)$, $h\ge h_*$ and $f\ge f_*$. Suppose $\nabla f$ is globally $L$-Lipschitz, $L>0$, $\gamma>0$, and $\mu_T\ge0$. Then a sufficient condition on the entire initial sublevel is
\[
 2\gamma+\mu_h/2>\sqrt{2L(E_0-h_*-f_*)}.
\]
\end{corollary}
\begin{proof}
On the sublevel, $f(Z)\le E_0-h_*$. The smooth descent lemma at $Z-G/L$, together with $f\ge f_*$, gives $\|G\|_F^2\le2L(f(Z)-f_*)$, even for a nonconvex $f$. Use $\lambda_{\max}(S)\le\|G\|_F$ in Theorem~\ref{thm:signed-curv}. The equal initialization makes $E_0$ independent of $\gamma$.
\end{proof}
The pointwise bound uses local information. The corollary is an a priori sufficient threshold; its constants can make it conservative. With negative transpose curvature the regional version additionally requires a radius bound. A positive quadratic agreement penalty by itself cannot bound common-factor directions on an unobserved subspace.

\subsection{Curvature in the directions that matter for agreement}
For any differentiable $\psi$ on symmetric matrices and any $\beta\in\mathbb R$, set
\[
 f(Z)=\psi(\sym Z)+\tfrac\beta2\|\skewop Z\|_F^2.
\]
Here $G-G^\top=\beta B$ identically: arbitrary nonconvex curvature of $\psi$ leaves $\mu_T=\beta$. It still affects $S$ and the difficulty of the block solve. More generally adding $\tau\|\skewop Z\|_F^2/2$ to any smooth extension preserves the symmetric gradient $S$ at each evaluation pair and preserves the objective on all PSD products. It increases the transpose pairing by $\tau\|B\|_F^2/2$ in the residual identity. At $X=-Y$ the product is symmetric and this correction has zero gradient. Section~\ref{sec:mask-classification} identifies an obstruction that such a modification cannot remove.

For example, let $C_i=C_i^\top$ be sensing matrices, let $b_i$ be measured scores and put
\[
 \psi(S)=\tfrac1N\sum_{i=1}^N \log\big(1+(\langle C_i,S\rangle-b_i)^2\big).
\]
The scalar loss has negative curvature away from the origin and its derivative has magnitude at most one. With $h$ the indicator of a common closed factor ball and $\beta\ge0$, one obtains $\lambda_{\max}(S_G)\le N^{-1}\sum_i\|C_i\|_2$ for $S_G=\sym\nabla f(Z)$. Thus $2\gamma>N^{-1}\sum_i\|C_i\|_2$ suffices for exactness at every joint critical point. The ball bounds all factors. The smooth loss is definable in the real exponential structure and the indicator is semi-algebraic, so the sum is definable and KL~\cite{abs2013framework}. A projected-gradient fallback therefore gives a concrete robust nonconvex sensing model covered by both the exactness and convergence results.

A scalar example gives the exact boundary for a nonconvex regularizer. Take $f=0$, $h(u)=(u^2-1)^2/4$, so $h''\ge-1$. On $x=-y=t$ the critical equations give $t(t^2-1+4\gamma)=0$. Moreover, subtracting the two equations at an arbitrary critical pair gives
\[
 (x-y)(4\gamma-1+x^2+xy+y^2)=0 .
\]
All critical factors agree exactly when $\gamma\ge1/4$ for $\gamma>0$. The general strict bound recovers every $\gamma>1/4$; this example separately resolves the boundary. It also shows why a convergence theorem for fixed positive $\gamma$ cannot dispense with a distinct exactness condition.

\begin{proposition}[A fixed coercive loss with no global exact penalty]\label{prop:no-global-gamma}
There exists a single real-analytic, coercive scalar loss with globally Lipschitz derivative and $h=0$ for which every finite $\gamma>0$ admits an unequal critical pair. This holds even though both the split objective and the original objective have the KL property.
\end{proposition}
\begin{proof}
Set $t=\log(1+z^2)$, $c=z^2/(1+z^2)$, and
\[
 f(z)=z^2(5/4+\sin t),\qquad f'(z)=2z(5/4+\sin t+c\cos t).
\]
Thus $f(z)\ge z^2/4$. The derivative of the parentheses times $z$ is bounded since $zt'=2c$ and $zc'=2z^2/(1+z^2)^2$ are bounded. Hence $f''$ is bounded, giving global smoothness and weak convexity. On the negative half-line choose $z=-\sqrt{\exp(t)-1}$. At $t=2\pi j$ the bracket is positive. At $t=2\pi j+5\pi/4$ it tends to $5/4-\sqrt2<0$. The corresponding positive values of $f'(z)$ tend to infinity. Continuity gives $f'(z)=2\gamma$ at some $z<0$ for each proposed penalty. Then $x=\sqrt{-z}$ and $y=-x$ satisfy both split equations. Both objectives are real analytic, hence locally KL at each critical point~\cite{abs2013framework}. Transpose symmetry in this scalar example is automatic.
\end{proof}
This example concerns all critical points over the entire factor space. It leaves room for an initialization-dependent certificate on a bounded loss sublevel as in Corollary~\ref{cor:initial-smooth}. KL convergence and even global smoothness of the loss alone cannot provide a global exact penalty for nonconvex data terms.

\section{Penalty Conclusions for Structured Models}
\label{sec:classes}
\paragraph{Unrestricted factors.}
Setting $h=0$ yields $X^\top X=Y^\top Y$ at every joint critical pair for $\gamma\ne0$, by subtracting $X^\top e_X$ and $e_Y^\top Y$. With full observations, $f(Z)=\|Z-A\|_F^2/2$ and $A=A^\top$, this implies $(A+2\gamma I)D+DT=0$, where $T=X^\top X=Y^\top Y$. For $\gamma>0$, all critical factors agree if and only if $A+2\gamma I\succeq0$, including equality at the spectral boundary (Theorem~\ref{thm:hzero-sharp}). For PSD $A$ every positive penalty therefore suffices. Gram balance alone gives no such general result for partial observations.
\paragraph{Shared nonnegativity constraint.}
For $h=\delta_+$ and the full Frobenius loss, Corollary~2 of~\cite{hu2026abmf} yields equal factors under either of two conditions: the columns within each factor are orthogonal and $\gamma\ge-\lambda_{\min}(A)$, or $r=1$ and $A$ is entrywise nonnegative, in both cases with $\gamma>0$. Appendix~\ref{app:symnmf-counterexample} records the rank-two failure for an entrywise nonnegative indefinite target. The antipodal mechanism below concerns unrestricted factors. For constrained factors the feasible midpoint directions instead enter Theorem~\ref{thm:fiber}.

\section{Observation Masks That Preserve Factor Agreement}
\label{sec:mask-classification}
Consider the weighted PSD approximation problem with unrestricted real factors ($h=0$),
\begin{equation}
 F_\gamma(X,Y)=\tfrac12\|XY^\top-A\|_W^2
                +\tfrac{\gamma}{2}\|X-Y\|_F^2,
 \label{eq:weighted-main}
\end{equation}
where $\|Z\|_W^2=\sum_{ij}w_{ij}Z_{ij}^2$, $A=A^\top$, $W=W^\top$ is entrywise nonnegative, and $\gamma>0$. An observation graph has an edge $ij$ for $i<j$ with $w_{ij}>0$; a diagonal entry records whether the self-similarity was observed. The following classification concerns all critical points, including saddles.

\begin{theorem}[Universal exactness of a binary mask]
\label{thm:mask-universal}
Fix any $r\ge1$ and symmetric $W\in\{0,1\}^{n\times n}$. The following conditions are equivalent.
\begin{enumerate}[leftmargin=*,itemsep=0pt]
\item For every $A\succeq0$ and $\gamma>0$, every critical point of \eqref{eq:weighted-main} has $X=Y$.
\item $W\succeq0$.
\item A common row and column permutation makes $W$ block diagonal with all-ones square blocks and an optional all-zero block.
\end{enumerate}
With all diagonal entries observed, this says that every connected component is a clique. With no diagonal observations, any observed edge rules out the universal assertion. On every failing mask the choice $A=\mathbf1\mathbf1^\top$ yields unequal critical factors whenever $0<2\gamma<-\lambda_{\min}(W)$. Scaling this rank-one target gives a counterexample at any prescribed positive penalty.
\end{theorem}

\begin{lemma}[Exact threshold for opposite factors]
\label{lem:antipodal}
For any weights as in \eqref{eq:weighted-main}, let $M=W\odot A$.
There exists a nonzero critical pair $(P,-P)$ if and only if
\begin{equation}
                  \lambda_{\min}(M)+2\gamma<0 .
 \label{eq:antipodal-threshold}
\end{equation}
\end{lemma}
\begin{proof}
On opposite factors the objective minus its value at the origin is
\[
 \langle P,(M+2\gamma I)P\rangle
       +\tfrac12\|PP^\top\|_W^2 .
\]
For a negative eigenvalue of $M+2\gamma I$, a small multiple of its eigenvector in one column lowers this value. The restricted objective is coercive because $F_\gamma(P,-P)\ge 2\gamma\|P\|_F^2$, and hence attains a nonzero minimum. Variations of the midpoint contribute zero first derivative there: in midpoint and difference coordinates, changing the sign of the midpoint transposes the product and leaves the objective invariant. Thus the restricted minimum is a critical point in both variables. Conversely, at a restricted critical point the derivative in the direction $P$ is twice the sum of the quadratic form and $\|PP^\top\|_W^2$. If $M+2\gamma I\succeq0$ both terms must vanish. Zero quartic implies $w_{ij}(PP^\top)_{ij}=0$ for every entry, so $\langle P,MP\rangle=0$ and the remaining quadratic is $2\gamma\|P\|_F^2$. This forces $P=0$, including at the spectral boundary.
\end{proof}

\begin{proof}[Proof of Theorem~\ref{thm:mask-universal}]
The implication (1)$\Rightarrow$(2) follows from Lemma~\ref{lem:antipodal} with $A=\mathbf1\mathbf1^\top$. For (2)$\Rightarrow$(3), write a PSD mask as a Gram matrix. A zero diagonal gives a zero vector and row. All other vectors are unit vectors: inner product one identifies the same vector and inner product zero gives orthogonality. Their equivalence classes produce the claimed blocks, each of which is PSD. This is the familiar clique characterization of positive semidefinite binary Gram matrices; the factorization implication requires a separate argument.

For (3)$\Rightarrow$(1), \eqref{eq:weighted-main} separates over the nonzero blocks and the unobserved rows. Within each block the target is a PSD principal submatrix. To see why its critical factors coincide, write them as $X_b,Y_b$ and the target as $A_b$. Multiplying the two gradient equations on the left by $X_b^\top$ and, after transposition, on the right by $Y_b$, and subtracting gives $X_b^\top X_b=Y_b^\top Y_b=T_b$. Subtracting the gradient equations themselves yields
\[
 (A_b+2\gamma I)(X_b-Y_b)+(X_b-Y_b)T_b=0 .
\]
Its inner product with $X_b-Y_b$ forces equality. Each unobserved row has only the penalty and obeys the same conclusion. At equal factors the split equations give symmetric stationarity.
\end{proof}

\paragraph{A minimal witness.}
With all diagonals observed, a connected component that is not complete contains an induced three-vertex path. On these vertices the mask has eigenvalue $1-\sqrt 2$. The target $aa^\top$, where $a$ indicates the three vertices, thus produces opposite critical factors for $0<\gamma<(\sqrt 2-1)/2$. If an observed edge has both diagonals missing, the two endpoints already suffice: take the two-by-two all-ones target, and $X=-Y=\sqrt{1-2\gamma}\,(1,-1)^\top$ for $0<\gamma<1/2$. The symmetric ground truth achieves zero objective for these rank-one targets. Existence of another critical point does not imply a solver reaches it from a given start.

\begin{figure}[t]
\centering
\begin{tikzpicture}
\begin{axis}[width=0.96\columnwidth,height=4.1cm,
  xlabel={Opposite-factor amplitude $t$}, ylabel={Objective change},
  xmin=-0.9,xmax=0.9,ymin=-0.053,ymax=0.21,xtick={-0.8,-0.4,0,0.4,0.8},
  ytick={0,0.1,0.2},tick label style={font=\scriptsize},label style={font=\scriptsize},
  legend style={at={(0.5,1.06)},anchor=south,legend columns=3,draw=none,font=\scriptsize},
  axis lines=left, clip=true]
\addplot[forget plot,black!40,thin,domain=-0.9:0.9] {0};
\addplot[blue!75!black,very thick,samples=81,domain=-0.9:0.9] {(0.2+1-sqrt(2))*x^2+7/16*x^4};\addlegendentry{$\gamma=.10$}
\addplot[black,thick,dashed,samples=81,domain=-0.9:0.9] {7/16*x^4};\addlegendentry{$\gamma=\gamma_*$}
\addplot[red!70!black,thick,dashdotted,samples=81,domain=-0.9:0.9] {(0.6+1-sqrt(2))*x^2+7/16*x^4};\addlegendentry{$\gamma=.30$}
\end{axis}
\end{tikzpicture}
\caption{Three vertices with observed diagonal and edges $12,23$. For the all-ones target, the plotted change is $F_\gamma(tv,-tv)-F_\gamma(0,0)$ with $v=(1,-\sqrt2,1)^\top/2$ and $\gamma_*=(\sqrt2-1)/2$. Below the threshold a small opposite split lowers the objective. The curve is a directional slice; Lemma~\ref{lem:antipodal} supplies joint criticality by minimization over the entire opposite-factor subspace.}
\label{fig:path-obstruction}
\end{figure}

For a fixed target, \eqref{eq:antipodal-threshold} detects one family of unequal critical points. Its failure excludes that family but does not prove full exactness. For real-valued weights, PSD of $W$ excludes this obstruction for every PSD target by the Schur product property; the clique decomposition and the sufficient direction of Theorem~\ref{thm:mask-universal} use binary weights. For nonnegative factors, opposite feasible factors must both vanish. Theorem~\ref{thm:fiber} treats the feasible directions for that model directly.

\subsection{Loss extensions and opposite critical factors}
\label{sec:extension-obstruction}
The symmetric objective determines the loss at PSD products only. Could a smooth change elsewhere eliminate the opposite-factor obstruction? The behavior of the loss at the origin already settles this question.

\begin{proposition}[Obstruction for smooth extensions]
\label{prop:smooth-no-go}
Let $\widetilde f$ be bounded below, differentiable on square matrices and transpose symmetric. Suppose it agrees with the weighted loss in \eqref{eq:weighted-main} on every $t vv^\top$ for $t\ge0$. If $\lambda_{\min}(W\odot A)+2\gamma<0$, the unrestricted split problem using $\widetilde f$ has a nonzero critical pair $(P,-P)$.
\end{proposition}
\begin{proof}
Transpose symmetry makes $\nabla\widetilde f(0)$ symmetric. The right derivative on each PSD rank-one ray gives $v^\top\nabla\widetilde f(0)v=-v^\top(W\odot A)v$, hence $\nabla\widetilde f(0)=-W\odot A$. On opposite factors the objective change near zero is $\langle P,(W\odot A+2\gamma I)P\rangle+o(\|P\|_F^2)$. A small step in a negative direction lowers it. The lower bound on $\widetilde f$ and $\gamma>0$ ensure coercivity on this subspace. At a nonzero minimizer the missing midpoint derivatives vanish by transpose symmetry, as in Lemma~\ref{lem:antipodal}.
\end{proof}

One convex nonsmooth extension changes exactly this derivative on negative traces. With $\theta\ge0$ add $\theta[-\operatorname{tr}Z]_+$ to the weighted loss. For $Z=VV^\top$ this addition vanishes. Each block remains convex since the added term is $\theta[-\langle X,Y\rangle]_+$.
\begin{corollary}[Exact coefficient for the opposite-factor branch]
\label{cor:hinge-branch}
A nonzero opposite critical pair exists for the modified split problem if and only if
\[
        2\gamma+\theta+\lambda_{\min}(W\odot A)<0 .
\]
Consequently $\theta_*=[-\lambda_{\min}(W\odot A)-2\gamma]_+$ is the smallest nonnegative coefficient eliminating that entire branch.
\end{corollary}
\begin{proof}
At $X=-Y=P\ne0$ the trace term is smooth. On this subspace it is $\theta\|P\|_F^2$; its first variation in the midpoint is zero. The proof of Lemma~\ref{lem:antipodal} applies with quadratic coefficient $M+(2\gamma+\theta)I$. At equality a vanishing quartic again gives $(2\gamma+\theta)\|P\|_F^2=0$.
\end{proof}
This rules out a specified family of critical points. Other unequal stationary pairs, including points on the trace-zero kink, require a separate analysis. If nonnegativity is imposed on both factors the hinge vanishes identically.

\paragraph{A sharp mask-only coefficient for normalized kernels.}
Suppose $A\succeq0$ and $A_{ii}=1$. Schur-multiplying $W-\lambda_{\min}(W)I\succeq0$ by $A$ yields $W\odot A\succeq\lambda_{\min}(W)I$. Equality of the smallest eigenvalues is attained at $A=\mathbf1\mathbf1^\top$. It follows that
\begin{equation}
 \sup_{\substack{A\succeq0\\\operatorname{diag} A=\mathbf1}}
       \theta_*(A,W,\gamma)=[-\lambda_{\min}(W)-2\gamma]_+.
 \label{eq:mask-hinge-uniform}
\end{equation}
The same worst-case positive part holds for $0\le A_{ii}\le1$: when $\lambda_{\min}(W)<0$, multiplying this scalar by $\operatorname{diag}(A)$ only increases the lower bound; when $W\succeq0$, $W\odot A\succeq0$. Thus one mask eigensolve covers every normalized kernel with this observation pattern. Replacing $-\lambda_{\min}(W)$ by $\max_i(\sum_{j\ne i}w_{ij}-w_{ii})$ gives a cheaper, conservative row-sum bound. Both statements concern removal of the opposite-factor family.

\paragraph{Exact block update.}
For fixed $Y$, write $x_i,y_i$ for transposed rows and let
\[
 K_i=\gamma I+\sum_j w_{ij}y_jy_j^\top,\quad
 b_i=\gamma y_i+\sum_jw_{ij}a_{ij}y_j .
\]
Compute $x_i^0=K_i^{-1}b_i$ and $p_i=K_i^{-1}y_i$ using the same factorizations. The exact minimizer is
\begin{equation}
\begin{aligned}
 x_i&=x_i^0+\alpha p_i, \qquad
 c_0=\sum_i y_i^\top x_i^0,\quad c_1=\sum_i y_i^\top p_i,\\
 \alpha&=\operatorname{clip}_{[0,\theta]}(-c_0/c_1)
                    \quad(c_1>0).
\end{aligned}
\label{eq:hinge-block}
\end{equation}
Take $\alpha=0$ when $c_1=0$, which forces $Y=0$. A hinge subgradient has coefficient $-\alpha$, with $\alpha=0$ at positive trace, $\alpha=\theta$ at negative trace, and any intermediate value at zero trace. The resulting linear equations are $K_ix_i=b_i+\alpha y_i$. Substitution gives trace $c_0+\alpha c_1$ and the stated clipping formula. If $c_0\ge0$ the second right-hand side can be skipped. The scalar hinge proximal calculation is standard; its role here is to change the proven obstruction while keeping the weighted block solves. The cost of obtaining $\theta_*$ or a certified upper bound must also be counted in a runtime comparison. We make no joint convergence claim for cyclic updates at a nonsmooth coupling.

\section{The Objective at a Fixed Midpoint}
\label{sec:fiber}
Let $U=(X+Y)/2$, $D=X-Y$, and $L_UH=HU^\top-UH^\top$. We allow unrestricted factors or a common entrywise nonnegativity constraint. Fix a feasible midpoint. The feasible differences form the entire factor space in the first case and the box $|D_{ij}|\le2U_{ij}$ in the second. Write $\mathcal V_U$ for the entire space or, for the box, for the linear subspace $H_{ij}=0$ wherever $U_{ij}=0$. Each direction of this subspace is feasible near zero in both signs. Define
\begin{align}
 S_U &= W\odot(UU^\top-A),\quad b(H)=\|HH^\top\|_W^2,\nonumber\\
 a_U(H)&=2\gamma\|H\|_F^2-\langle H,S_UH\rangle+
                     \tfrac12\|L_UH\|_W^2, \label{eq:fiber-coeff}\\
 k(U)&=\max_{\substack{H\in\mathcal V_U\\\|H\|_F=1}}
       \left(\langle H,S_UH\rangle-\tfrac12\|L_UH\|_W^2\right).
 \label{eq:fiber-thresh}
\end{align}
Set $k(U)=-\infty$ on a zero-dimensional feasible subspace.

\begin{theorem}[Exact agreement at a specified midpoint]
\label{thm:fiber}
For \eqref{eq:weighted-main} with $\gamma>0$ and either factor domain, let $\phi_U(D)=F_\gamma(U+D/2,U-D/2)$. Then
\begin{equation}
 \phi_U(tH)-\Phi(U)=\tfrac14 t^2 a_U(H)+\tfrac1{32}t^4 b(H)
 \label{eq:fiber-quartic}
\end{equation}
for every feasible $tH$. The following are equivalent: $2\gamma\ge k(U)$; $D=0$ is the unique global minimizer at this midpoint; $D=0$ is the only constrained stationary difference at this midpoint. If $2\gamma<k(U)$ the fixed-midpoint problem has nonzero global minimizers in opposite pairs. For unrestricted factors, in a direction with $a_U(H)<0$ the optimizing magnitude and objective gain are
\[
 \begin{aligned}
 |t_*|&=\sqrt{-4a_U(H)/b(H)},\\
 \Phi(U)-\phi_U(t_*H)&=a_U(H)^2/(2b(H)).
 \end{aligned}
\]
\end{theorem}
\begin{proof}
The product is $UU^\top+tL_UH/2-t^2HH^\top/4$. Symmetric and skew-symmetric matrices are orthogonal under the symmetric weights. Expanding the square yields \eqref{eq:fiber-quartic}. If $b(H)=0$, the support of $W$ sees no entry of $HH^\top$, so $\langle H,S_UH\rangle=0$ and $a_U(H)\ge2\gamma\|H\|_F^2$. When $2\gamma\ge k(U)$, $a_U(H)\ge0$; for each nonzero feasible $H$ at least one polynomial coefficient is strictly positive. The feasible inward ray therefore gives strict descent at every nonzero $D$, including on a box face. At $D=0$ the first derivative vanishes. If $2\gamma<k(U)$ a feasible small ray has strictly smaller value. Existence follows from coercivity of the penalty at fixed $U$ (or compactness of the box); evenness gives the opposite minimizer. Minimizing the polynomial in $t^2$ proves the final formula.
\end{proof}
A nonzero minimum with fixed $U$ need not be critical when $U$ is free. Lemma~\ref{lem:antipodal} exploits the exceptional midpoint $U=0$, where first-order stationarity in $U$ is automatic on unrestricted factors.

\paragraph{Nonnegativity can change the threshold.}
Take $r=1$, full observations, $U=(1,0)^\top$, and $A=(1,-1)^\top(1,-1)\succeq0$. The operator in \eqref{eq:fiber-thresh} is the matrix with rows $(0,1)$ and $(1,-2)$. Its unrestricted top eigenvalue is $\sqrt2-1$, whereas its restriction to the feasible line $\mathbb R(1,0)^\top$ is zero. The midpoint is also critical for the nonnegative symmetric problem. Directions that split a zero entry into opposite signs have no role in its constrained stability.

\paragraph{Objective change at a computed midpoint.}
For unrestricted factors let $e_X,e_Y$ be the current split gradients. Differentiating \eqref{eq:fiber-quartic} at $t=1$, $H=D$, and eliminating $a_U(D)$ gives
\begin{equation}
 F_\gamma(X,Y)-\Phi(U)=
 \tfrac14\langle D,e_X-e_Y\rangle-\tfrac1{32}\|DD^\top\|_W^2.
 \label{eq:fiber-work}
\end{equation}
For the nonnegative constraint, adding the endpoint normal-cone selections (with the $1/2$ block weighting) to the gradients replaces equality by $\le$ in \eqref{eq:fiber-work}, since their difference has nonnegative inner product with $D$. At an unequal joint critical pair, replacing both factors by their average therefore raises the objective by at least $\|DD^\top\|_W^2/32$; the unrestricted formula is exact. If all diagonal weights are positive this bound is strictly positive whenever $D\ne0$. A test based only on $\|D\|_F$ cannot give the sign of the objective change at a finite iterate.

\subsection{A fixed-penalty correction}
\label{sec:fiber-correction}
After a pair of block updates, denote the resulting point by $X,Y$. At the same midpoint, minimize \eqref{eq:fiber-quartic} over $0\le t\le1$. For $b(D)>0$ choose
\begin{equation}
 t_*=\sqrt{\operatorname{clip}_{[0,1]}(-4a_U(D)/b(D))}.
 \label{eq:fiber-correction}
\end{equation}
If $b(D)=0$, take $t_*=0$. Keep $D$ if
$\phi_U(D)-\phi_U(t_*D)<\eta(1-t_*)^2\|D\|_F^2$ for fixed $\eta>0$; otherwise use $t_*D$. This correction preserves feasibility and the fixed-penalty descent. It leaves the output midpoint unchanged at that sweep; any change in symmetric accuracy must come from subsequent sweeps. The sums for $a_U,b$ use only observed entries and row inner products at $O(|\Omega|r)$ work, where $|\Omega|$ counts observed entries. The implementation in Section~\ref{sec:digits-fiber} uses dense masked arrays at its small problem size.

\begin{proposition}[Fixed-penalty accumulation points]
\label{prop:fiber-iterate}
Use one projected-gradient step per block with step $0.9/(\gamma+\|W\|_{\max}\|Y\|_2^2)$ for the first block and the analogous step using the updated first factor for the second, followed by \eqref{eq:fiber-correction} and its acceptance test. If the initial sublevel is bounded, every accumulation point is a joint critical point of the fixed-penalty problem. Positive diagonal weights suffice for boundedness.
\end{proposition}
\begin{proof}
The denominators bound the block-gradient Lipschitz constants; here $\|W\|_{\max}=\max_{ij}w_{ij}$. The descent lemma and the projection inequality give a decrease bounded below by a positive constant times the squared change in each block, uniformly since $\gamma>0$. The accepted correction also pays for its squared factor change, $(1-t_*)^2\|D\|_F^2/2$. Summing the decreases proves that all three changes tend to zero. On a bounded sublevel the step sizes are bounded away from zero. Along any convergent subsequence the intermediate pairs have the same limit; continuity of the gradients, step sizes, and projections gives both projected-gradient fixed-point equations there. For the sublevel claim, bounded objective controls $\|D\|_F$ and each $x_i^\top y_i$ through its observed diagonal. The identity $\|u_i\|_2^2=x_i^\top y_i+\|d_i\|_2^2/4$ then bounds both factors.
\end{proof}

\section{Certificates at a Computed Pair}
\label{sec:general-bridge}
The clique classification uses unrestricted factors and a binary observation pattern. The polynomial along a fixed midpoint uses a squared loss. For the convex special case a residual argument applies to a finite, transpose-symmetric convex loss $f$ and a shared convex regularizer $h$. At a current pair take a common $G\in\partial f(XY^\top)$, endpoint subgradients $H_X,H_Y$ of $h$, and let $e_X,e_Y$ be the two left sides of \eqref{eq:joint-critical}. For $D=X-Y\ne0$ and the Frobenius inner product, put
\[
 m=2\gamma+\langle D,GY-G^\top X\rangle/\|D\|_F^2 .
\]
Monotonicity of the subgradients of $h$ gives $m\|D\|_F^2\le\langle D,e_X-e_Y\rangle$. Positive $m$ therefore certifies consensus to accuracy $\|e_X-e_Y\|_F/m$. Both gradient products are available at the same endpoint. Transpose symmetry accounts for the part of this bound lost when the transpose-difference term is omitted in Lemma~1 of~\cite{hu2026abmf}. Appendix~\ref{sec:finite-accuracy} gives a computable bound on the spectral improvement and a proximal-gradient certificate for the reported midpoint when $h$ is nonsmooth.

The bounds use different information. The mask and input tests in Theorem~\ref{thm:mask-universal} and Lemma~\ref{lem:antipodal} apply before choosing factors. Theorem~\ref{thm:fiber} compares every split at one specified midpoint. The residual certificate addresses an actual endpoint. The certificate can be evaluated at any current penalty. Appendix~\ref{sec:adaptive} gives conditions under which a sequence of penalty trials returns a certified midpoint in finite work. For full-observation PSD approximation, Appendix~\ref{sec:unregularized} gives a direct input-based exactness test and a finite-residual bound from Gram balance. Appendix~\ref{sec:curvature} treats the cost of a one-sided block update relative to a symmetric LM system.

\section{One-Sided Displacements and Inexact Block Solves}
\label{sec:delta-solve}
Hold a factor $V$ fixed, and write the updated factor as $V+\Delta$. By transpose symmetry the same subproblem applies to either block:
\begin{equation}
 q_V(\Delta)= f((V+\Delta)V^\top)
          +\tfrac12 h(V+\Delta)+\tfrac\gamma2\|\Delta\|_F^2.
 \label{eq:q-delta}
\end{equation}
The omitted term $h(V)/2$ is constant. A sweep updates the first factor with $V=Y^k$, then the second with $V=X^{k+1}$. For the second block $\Delta=Y^{k+1}-X^{k+1}$. In a call with old free factor $T$, initialize $\Delta_0=T-V$ and write $s=\Delta-\Delta_0$. The distance that controls inner accuracy is $\|s\|_F$, the change of the updated block. Initializing every call at zero would discard the old point used in the descent proof.

\subsection{What a zero displacement says about the original objective}
\label{sec:zero-step}
At a symmetric anchor $V$ write $S=VV^\top$ and let $q_V$ be the one-sided model in \eqref{eq:q-delta}. Zero displacement recovers the original value, $q_V(0)+h(V)/2=\Phi(V)$. We ask whether a nonzero direction $H$ at this base point decreases $\Phi$, and whether an optimal zero block step certifies stationarity of $\Phi$. Both questions concern movement of the anchor, distinct from contraction toward an existing midpoint in Section~\ref{sec:fiber}.

\begin{theorem}[Directional information at a symmetric anchor]
\label{thm:anchor-slope}
Let $h$ be proper closed convex, $V\in\dom h$, and $f$ be transpose symmetric. For differentiable $f$, including nonconvex $f$,
\begin{equation}
 \Phi'(V;H)=2q_V'(0;H).
 \label{eq:anchor-equality}
\end{equation}
For finite convex $f$, possibly nonsmooth,
\begin{equation}
 \Phi'(V;H)\le 2q_V'(0;H).
 \label{eq:anchor-major}
\end{equation}
Here primes denote right directional derivatives. In the second case define the compact convex sets
\begin{equation}
 \begin{aligned}
 C_V&=\{GV:G\in\partial f(S)\},\\
 A_V&=\{(\sym G)V:G\in\partial f(S)\}.
 \end{aligned}
 \label{eq:anchor-sets}
\end{equation}
Then $A_V\subseteq C_V$. The loss terms on the two sides of \eqref{eq:anchor-major} agree in every direction if and only if $C_V=A_V$. In direction $H$ the excess slope of the doubled one-sided loss is exactly
$2[\sigma_{C_V}(H)-\sigma_{A_V}(H)]$, where $\sigma_C(H)=\max_{c\in C}\langle c,H\rangle$.
\end{theorem}
\begin{proof}
Set $B=HV^\top$. Moving the symmetric factor by $tH$ changes the product by $t(B+B^\top)+t^2HH^\top$, whereas the one-sided product changes by $tB$. Differentiability, or local Lipschitz continuity of finite convex $f$, makes the quadratic term immaterial to the derivative. At $S=S^\top$, a differentiable loss has symmetric gradient, so $\langle\nabla f(S),B+B^\top\rangle=2\langle\nabla f(S),B\rangle$. The shared regularizer gives the same $h'(V;H)$ after doubling and the quadratic penalty has zero derivative. For convex $f$, the loss derivative is the support function of $\partial f(S)$. Transpose invariance and convexity put $(G+G^\top)/2$ in this same set. The two loss slopes are consequently $2\sigma_{A_V}(H)$ and $2\sigma_{C_V}(H)$, respectively. A compact convex set is determined by its support function. The comparisons of loss slopes are finite even when the regularizer derivative is extended valued.
\end{proof}

If $C_V=A_V$ and $q_V$ is convex, zero solves the block exactly when $\Phi$ is directionally stationary at $V$ (its right derivative is nonnegative on every feasible ray). This includes a smooth loss and a convex block. In the nonsmooth case a surplus one-sided slope can conceal descent. The next example uses the ordinary entrywise absolute loss, with every matrix entry observed.

\begin{proposition}[An exact zero step at a nonstationary PSD fit]
\label{prop:anchor-kink}
Let $h=0$, $r=1$, $f(Z)=\sum_{ij}|Z_{ij}-A_{ij}|$, and take
\[
 A=\begin{pmatrix}1/2&1\\1&3\end{pmatrix}\succ0,\qquad
 V=(1,1)^\top,\qquad H=(-1,1)^\top.
\]
For every $\gamma>0$, the unique minimizer of $q_V$ is zero, while $\Phi'(V;H)=-4$. The example also holds with nonnegative factors.
\end{proposition}
\begin{proof}
At $S=VV^\top$ the two diagonal residuals have opposite signs and both off-diagonal residuals vanish. For $d=(d_1,d_2)^\top$,
\[
 q_V'(0;d)=d_1-d_2+|d_1|+|d_2|\ge0.
\]
Convexity and the positive ridge give optimality and uniqueness. The symmetric product perturbation in direction $H$ has diagonal $(-2,2)$ and zero off-diagonal entries to first order. In fact $\Phi(V+tH)=5/2-4t+2t^2$ for all sufficiently small $t\ge0$. The anchor is interior to the nonnegative orthant. Since $\det A=1/2$ and $A_{11}>0$, the target is positive definite.
\end{proof}
Here an alternating solver initialized with both factors equal stays at $V$ for every penalty. For $\gamma>1$, inequality~\eqref{eq:spectral-margin} holds uniformly, since all loss subgradients have spectral norm at most two. The margin concerns joint critical pairs; two optimal block steps at this kink can use different loss subgradients. Proposition~\ref{prop:kink-stall} illustrates the related problem of stopping at unequal factors.

\begin{corollary}[Which observations hide a descent direction?]
\label{cor:edge-slope}
For $f(Z)=\sum_{ij}w_{ij}|Z_{ij}-A_{ij}|$, let $W=W^\top\ge0$ entrywise, $A=A^\top$, and let $\mathcal E_0$ consist of pairs $i<j$ with $w_{ij}>0$ and $S_{ij}=A_{ij}$. Set $b_{ij}=\langle H_i,V_j\rangle$, using row vectors. The excess of the doubled one-sided loss slope over the symmetric loss slope is
\begin{equation}
 2\sum_{(i,j)\in\mathcal E_0}w_{ij}
 \big(|b_{ij}|+|b_{ji}|-|b_{ij}+b_{ji}|\big).
 \label{eq:edge-slope}
\end{equation}
For unrestricted directions the two slopes agree for every $H$ if and only if each edge in $\mathcal E_0$ has an endpoint with zero factor row.
\end{corollary}
\begin{proof}
At any nonzero residual the two ordered entries have the same sign, so their linear contributions cancel in the difference of slopes. Diagonal entries cancel as well. At an exactly fitted edge, the absolute value gives the displayed triangle gap. If $V_i$ or $V_j$ vanishes this gap is zero. Otherwise choose $H_i=V_j$ and $H_j=-V_i$. The two scalar terms then have opposite strict signs and every other edge has nonnegative gap.
\end{proof}
The condition characterizes equality of all slopes; failure at one edge alone does not imply an algorithm will stop. It also gives an implementable check for $h=0$. On a zero block step, set $T=T^\top$, fix $T_{ij}=w_{ij}\operatorname{sign}(S_{ij}-A_{ij})$ on nonzero residuals, and impose $|T_{ij}|\le w_{ij}$ on zero residuals. Solve the box-constrained least-squares problem
\[
 T_*\in\arg\min_T\|2TV\|_F^2,\qquad g_*=2T_*V .
\]
This uses one shared variable for each symmetric fitted entry, including a fitted diagonal. The feasible images $2TV$ form the set $2A_V$. Consequently $g_*=0$ certifies directional stationarity, and otherwise $\Phi'(V;-g_*)=-\|g_*\|_F^2$ by projection of the origin onto this compact convex set. In particular one can test a stalled fit without changing the loss extension for the regular block updates. For an approximate feasible $T$ one may evaluate the actual directional slope of $-2TV$ and accept a strict descent direction. A claim of exact stationarity requires the exact condition above. Replacing exact ties by a numerical active-set tolerance changes the test.

\paragraph{Certifying descent with a floating-point tolerance.}
For the same weighted absolute loss with $h=0$, enlarge the set of fitted observations to include entries with $|(VV^\top-A)_{ij}|\le\tau$. Use a symmetric sign outside this band and allow a shared sign in $[-1,1]$ on each positive-weight unordered observation inside it. Stack the factor entries into a vector. The resulting trial gradients form
\[
 \mathcal G_\tau=\{g_0+Mz:\ |z_k|\le1\}.
\]
Here $g_0$ is twice the fixed signed residual weights times $V$. A column of $M$ contributes $2w_{ij}V_j$ in row $i$ and $2w_{ij}V_i$ in row $j$ for an off-diagonal observation. A diagonal column contributes $2w_{ii}V_i$ once. For $\tau>0$, some of these signs can differ from the true loss signs. Nevertheless $2A_V\subseteq\mathcal G_\tau$ whenever the band contains every actual tie and signs outside it are correct.

\begin{proposition}[An inexact test with a band of near-zero residuals]
\label{prop:band-descent}
For any feasible $z$, put $g=g_0+Mz$ and $e(z)=\langle z,M^\top g\rangle+\|M^\top g\|_1$. Then
\[
 \Phi'(V;-g)\le -\|g\|^2+e(z).
\]
If $g\ne0$ and $e(z)\le\theta\|g\|^2$ for $0\le\theta<1$, $-g$ is a strict descent direction and $\dist(0,2A_V)\ge(1-\theta)\|g\|$.
\end{proposition}
\begin{proof}
The upper slope for any $H$ is the support function of $\mathcal G_\tau$, namely $\bar s_\tau(H)=\langle g_0,H\rangle+\|M^\top H\|_1$. Substituting $-g$ proves the first inequality. For every $p\in2A_V$, $\langle p,-g\rangle\le\bar s_\tau(-g)$. Cauchy--Schwarz gives the distance bound. The quantity $e$ is the usual Frank--Wolfe gap~\cite{jaggi2013fw} for minimizing $\|g_0+Mz\|^2/2$ over the box.
\end{proof}
The test permits early termination of projected-gradient iterations. A zero vector obtained using the enlarged band is inconclusive for exact stationarity. Section~\ref{sec:robust-checkpoints} compares the band-aware direction with the genuine steepest direction for exactly represented data and measures the cost of this test.

\paragraph{Step lengths with guaranteed decrease.}
Write $R=VV^\top-A$, $B=HV^\top+VH^\top$, and $C=\sum_{ij}w_{ij}|(HH^\top)_{ij}|$. Among positive-weight entries outside the band with $R_{ij}B_{ij}<0$, let $t_s$ be the smallest $|R_{ij}/B_{ij}|$ (an empty minimum is $+\infty$). For $0\le t\le t_s$, the linearized residuals outside the band retain their signs up to a possible zero at the endpoint. Inside the band the triangle inequality gives $|R_{ij}+tB_{ij}|-|R_{ij}|\le t|B_{ij}|$. Restoring $t^2HH^\top$ and applying the triangle inequality once more gives
\begin{equation}
 \Phi(V+tH)-\Phi(V)\le t\bar s_\tau(H)+t^2 C .
 \label{eq:finite-band-step}
\end{equation}
Consequently, if $\eta=-\bar s_\tau(H)>0$, every $0<t\le\min(t_s,\eta/(2C))$ decreases $\Phi$ by at least $t\eta/2$, with the second limit omitted for $C=0$. With $\tau=0$, $t_s$ can be arbitrarily small even for the exact steepest direction. A positive band can remove the first sign crossing from this restriction at the cost of a more conservative slope. This bound describes a guaranteed interval; larger accepted steps are possible.

\paragraph{Comparing one-sided and symmetric steps at finite displacement.}
For the weighted absolute loss with $W=W^\top\ge0$ entrywise, take $h=0$, $R=VV^\top-A$ and $H=\Delta/2$. Write $b=HV^\top$ and $\ell_W(Z)=\sum_{ij}w_{ij}|Z_{ij}|$. Compare the one-sided block and the symmetric prox-linear subproblem in the same units:
\begin{equation}
\begin{aligned}
 m_o(H)&=\ell_W(R+2b)+\tfrac\lambda2\|H\|_F^2,\\
 m_s(H)&=\ell_W(R+b+b^\top)+\tfrac\lambda2\|H\|_F^2,
 \qquad \lambda=4\gamma>0 .
\end{aligned}
\label{eq:two-absolute-models}
\end{equation}
Let $[a]_+=\max(a,0)$. For any rank, their exact difference is
\begin{equation}
\begin{aligned}
 \mathcal E_V(H)&=m_o(H)-m_s(H)\\
 &=2\sum_{i<j}w_{ij}
 [|b_{ij}-b_{ji}|-|R_{ij}+b_{ij}+b_{ji}|]_+ .
\end{aligned}
\label{eq:finite-sign-cost}
\end{equation}
Indeed, on the ordered entries of a pair the one-sided residuals have midpoint $p=R_{ij}+b_{ij}+b_{ji}$ and half-difference $q=b_{ij}-b_{ji}$. Their absolute values sum to $2\max(|p|,|q|)$, whereas the symmetric loss contributes $2|p|$. Thus a pair contributes strictly precisely when the two one-sided predicted residuals have opposite strict signs and positive weight. Diagonals agree identically.

Let $H_o,H_s$ minimize the respective models. Adding their strong-convexity inequalities yields
\begin{equation}
 \lambda\|H_o-H_s\|_F^2\le \mathcal E_V(H_s)-\mathcal E_V(H_o).
 \label{eq:finite-sign-distance}
\end{equation}
In particular the optimal values agree if and only if $\mathcal E_V(H_s)=0$, in which case the updates agree. The converse about values follows from $m_o\ge m_s$ and uniqueness. Equations~\eqref{eq:finite-sign-cost} and~\eqref{eq:finite-sign-distance} describe the finite-step counterpart of Corollary~\ref{cor:edge-slope}. A nonzero residual can begin contributing once the proposed step crosses its sign.

\begin{proposition}[Largest ball of identical one-sided and symmetric models]
\label{prop:absolute-ball}
For each observed off-diagonal pair put $a_{ij}=\max(\|V_i\|,\|V_j\|)$ and define
\[
 s_{ij}=\begin{cases}
  |R_{ij}|/(2a_{ij}) & R_{ij}\ne0,\ a_{ij}>0,\\
  0 & R_{ij}=0,\ \|V_i\|\,\|V_j\|>0,\\
  +\infty & \text{otherwise}.
 \end{cases}
\]
Let $s_*=\min_{i<j:\,w_{ij}>0}s_{ij}$, with $\min\varnothing=+\infty$. Then $m_o(H)=m_s(H)$ for every $\|H\|_F\le t$ if and only if $0\le t\le s_*$, for radii $t\ge0$.
\end{proposition}
\begin{proof}
If $R_{ij}\ne0$, each of $2|b_{ij}|$ and $2|b_{ji}|$ is at most $2a_{ij}t$. Up to and including $s_{ij}$ the two residuals therefore cannot have opposite strict signs. For any larger radius and $a_{ij}>0$, change only the row multiplying a largest-norm partner and choose its sign to flip that one residual. The other residual retains its original nonzero sign. At an exact tie, any positive radius permits opposite signs when both partner rows are nonzero, by taking small steps in the two rows independently. If either partner vanishes at a tie, one predicted residual stays zero. Pairs with two zero partners and diagonal observations never contribute. Equation~\eqref{eq:finite-sign-cost} sums nonnegative terms and gives both implications.
\end{proof}
For $s_*>0$ the result also yields a sufficient damping condition for equality of the two updates. If $g\in\partial m_s(0)$ and $\|g\|_F\le\lambda s_*$, strong monotonicity gives $\|H_s\|_F\le\|g\|_F/\lambda\le s_*$. Then $\mathcal E_V(H_s)=0$. This concerns two update models at one anchor. It gives no penalty threshold for agreement of all stationary factor pairs.

A computable test also works for inexact inner solutions. A dual iterate $U$ of the one-sided box problem obeys $|U_{ij}|\le w_{ij}$. Its symmetric part $T=(U+U^\top)/2$ is feasible for the symmetric dual. For an arbitrary candidate $H$ let
\[
 B_s(H,U)=m_s(H)-\langle R,T\rangle+
                  \frac{\|2TV\|_F^2}{2\lambda}.
\]
The box representation of $\ell_W$ and minimization of its affine lower bound plus the ridge give
\[
 \tfrac\lambda2\|H-H_s\|_F^2
 \le m_s(H)-\min m_s\le B_s(H,U).
\]
A small bound certifies transfer of an inexpensive block step to the symmetric model. If a feasible symmetric trial $P$ satisfies $m_s(H)-m_s(P)>\varepsilon$, it instead rules out $\varepsilon$-optimality. A large upper bound by itself is inconclusive. More generally, if $\widetilde H_j$ have certified model gaps $e_j$ for $j=o,s$, the triangle inequality gives
\[
 \|H_o-H_s\|_F\ge\|\widetilde H_o-\widetilde H_s\|_F
            -\sqrt{2e_o/\lambda}-\sqrt{2e_s/\lambda}.
\]
A positive lower bound distinguishes two genuinely different exact updates from errors of the numerical inner solvers.

\paragraph{Choosing a sign band for a specified step length.}
For $\|H\|_F\le1$, the exact norm of the linear functional $H\mapsto (HV^\top+VH^\top)_{ij}$ is
\[
 L_{ij}=\sqrt{\|V_i\|^2+\|V_j\|^2}\quad(i\ne j),
 \qquad L_{ii}=2\|V_i\| .
\]
This follows from Cauchy--Schwarz on the two affected rows, with equality along their representing vector. For a requested radius $t>0$, keep the signs free on $|R_{ij}|\le t L_{ij}$ and fixed elsewhere. No linearized residual outside this band changes sign before $t$, for any unit direction. Construct the upper slope $\bar s_t$ as in Proposition~\ref{prop:band-descent} using these entry-dependent bands. The proof of \eqref{eq:finite-band-step} yields the lower bound
\[
 \Phi(V)-\Phi(V+tH)\ge -t\bar s_t(H)
     -t^2\sum_{ij}w_{ij}|(HH^\top)_{ij}| .
\]
The implementation compares four radii by this bound and takes the candidate with the largest strictly positive guaranteed decrease. This sufficient test can be conservative. Its cost and effect on unobserved entries are measured in Section~\ref{sec:masked-rank}.

\paragraph{Restoring directional equality.}
When the data specify the loss only on symmetric predictions, one may use
\[
 f_{\rm s}(Z)=f(\sym Z)+\tfrac\beta2\|\skewop Z\|_F^2,\qquad \beta\ge0 .
\]
This preserves every original prediction value. At $S$, $f_{\rm s}'(S;B)=f'(S;\sym B)$ for finite convex $f$, so $2f_{\rm s}'(S;B)=f'(S;B+B^\top)$. Thus zero is a faithful stationarity test for the exact convex block of this extension. With $h=0$, $\beta=0$ and $P=\Delta/2$, the block is precisely the prox-linear model for $V\mapsto VV^\top$ with ridge $2\gamma\|P\|_F^2$. This option can couple rows that were separable for an entrywise extension. It repairs first-order information on the diagonal. The off-diagonal exactness conditions in Section~\ref{sec:framework} are a separate requirement.

\paragraph{A test for an inexact descent direction.}
Suppose \eqref{eq:anchor-major} or \eqref{eq:anchor-equality} holds and $q_V$ is $\mu$-strongly convex, $\mu>0$. For any computed feasible $d\ne0$ define
\begin{equation}
 c_V(d)=2[q_V(0)-q_V(d)]+\mu\|d\|_F^2 .
 \label{eq:anchor-cert}
\end{equation}
Scalar convexity of $q_V(td)-\mu t^2\|d\|_F^2/2$ yields $\Phi'(V;d)\le -c_V(d)$. If a computed $e\in\partial q_V(d)$ satisfies $\|e\|_F\le\eta\|d\|_F$, strong convexity at $d$ also gives $c_V(d)\ge2(\mu-\eta)\|d\|_F^2$. Whenever $c_V(d)>0$, backtracking from $t=1$ along $V+td$ terminates with $\Phi(V+td)\le\Phi(V)-\alpha t c_V(d)$ for $0<\alpha<1$. This follows from the definition of the right derivative. A call can use the value test without solving the block to a relative residual. An undamped unit step can increase $\Phi$: for $f(z)=(z-1)^2/2$, $h=0$, $V=1/4$ and $\gamma=1/16$, the exact displacement is $15/8$, giving $V+d=17/8$. The acceptance test here uses the original symmetric value and the zero-step baseline. Ordinary alternating updates use the old free factor and the split objective in \eqref{eq:block-dec}; their fixed-penalty KL theorem concerns that trajectory.

\subsection{Computing and accepting an alternating block update}
\paragraph{Loss-dependent implementations.}
Let $g_V(T)=f(TV^\top)+\gamma\|T-V\|_F^2/2$ and $r(T)=h(T)/2$. When $f$ is differentiable, $\nabla g_V(T)=\nabla f(TV^\top)V+\gamma(T-V)$. For unweighted square loss and $h=0$,
\[
 \Delta(V^\top V+\gamma I)=AV-V(V^\top V).
\]
One small Cholesky factorization serves all rows. For the masked square loss with symmetric weights, let $v_j$ denote transposed rows of $V$. In row $i$, solve $B_i\delta_i=b_i$ with
\[
 B_i=\gamma I+\sum_j w_{ij}v_jv_j^\top,\qquad
 b_i=\sum_j w_{ij}(A_{ij}-v_i^\top v_j)v_j .
\]
Rows sharing weights share a factorization. For large $r$, preconditioned CG can apply the row matrices by products with the observed factors and stop using $B_i\delta_i-b_i$. Sum squared residuals over rows for the block test below. A separable nonsmooth $h$ preserves independent row subproblems; a coupled penalty can instead use a block proximal map. If the row systems have spectrum in $[\gamma,M_i]$, standard CG bounds give the reduction $2((\sqrt{\kappa_i}-1)/(\sqrt{\kappa_i}+1))^j$ in the energy-norm error after $j$ iterations, $\kappa_i=M_i/\gamma$; direct factorization is often preferable at small rank. Initialize each inner solve at its old displacement. For a linear sensing map $\mathcal A$ and squared measurement loss the Hessian action on $P$ is $\mathcal A^*\mathcal A(PV^\top)V+\gamma P$. These formulas require neither a dense $n\times n$ Jacobian nor symmetry of the one-sided perturbation.

\paragraph{When inner acceleration applies.}
Suppose $f$ and $h$ are respectively $\rho_f$- and $\rho_h$-weakly convex on their full domains (use zero for ordinary convexity). Then $q_V$ has lower curvature
\[
 m_V=\gamma-\rho_f\|V\|_2^2-\rho_h/2 .
\]
This follows by affine composition and addition of the ridge. For smooth $f$ with gradient Lipschitz constant $L_f$, the smooth part is $(L_f\|V\|_2^2+\gamma)$-smooth. If needed, add $\lambda\|s\|_F^2/2$ to the inner objective. This preserves its first derivative at the old free factor and raises $m_V$ by $\lambda$. One explicit convex splitting uses the smooth term
\[
 f(TV^\top)+\tfrac{\rho_f\|V\|_2^2}{2}\|T\|_F^2
\]
and puts the remaining quadratics, $h(T)/2$, and any linear terms in the proximal term. The latter has modulus at least $m'=m_V+\lambda$ and the smooth term is convex with Lipschitz gradient bound $L'=(L_f+\rho_f)\|V\|_2^2$. Its proximal map reduces to a rescaled and shifted prox of $h$ when the resulting quadratic is positive. For $m'>0$, $L'>0$ and a tractable proximal map, a convex accelerated epoch of length $p$ obeys the FISTA gap bound $2L'\|T_0-T_*\|_F^2/(p+1)^2$ \cite{beck2009}. Strong convexity gives a gap contraction by $4L'/(m'(p+1)^2)$ per restart when $p+1>2\sqrt{L'/m'}$. This inner complexity concerns the block objective. The acceptance test below is evaluated separately at the proposed output.

\paragraph{Two implementable acceptance tests.}
Fix constants $a>0$, $0\le b<\infty$, common to all accepted blocks. With $T'=V+\Delta$, require a computable $v\in\partial r(T')$ such that
\begin{align}
 (g_V+r)(T)-(g_V+r)(T')&\ge a\|s\|_F^2,\label{eq:block-dec}\\
 \|\nabla g_V(T')+v\|_F&\le b\|s\|_F.\label{eq:block-res}
\end{align}
An objective-gap tolerance or a fixed number of accelerated iterations alone does not establish either test. The following two implementations exhibit usable constants.

\emph{Relative residual for a regularized block.}
Suppose $m_V\ge m_0$ along the run, $0\le\lambda\le\bar\lambda$, and the inner solver returns
$\|\nabla g_V(T')+v+\lambda s\|_F\le\eta\|s\|_F$.
The subgradient inequality with curvature $m_V$ gives
\[
 (g_V+r)(T)-(g_V+r)(T')\ge
 (\lambda+m_0/2-\eta)\|s\|_F^2.
\]
Thus $b=\bar\lambda+\eta$ works if the coefficient on the right has a uniform positive lower bound $a$. Choose $m_V+\lambda>0$ to make the regularized inner problem strongly convex. In the convex setting $\gamma>0$ gives $m_0\ge\gamma$; with $\lambda=0$ any $0<\eta<\gamma/2$ suffices. Exact AM is included with $\eta=0$. For a nonsmooth term, a final proximal-gradient step from an accelerated candidate $y$ yields an explicit subgradient $v=M(y-T')-\nabla g_V(y)$ for an unregularized proximal step with model parameter $M$; include $\lambda(y-T)$ in the smooth gradient for a regularized step. A vanishing relative tolerance at $s=0$ requires verification of stationarity.

\emph{One proximal-gradient fallback.}
Let $L_B$ bound the Lipschitz constant of $\nabla g_V$ on the step segments and $M\in[L_B+2a,\bar M]$. Set
\[
 T'\in\arg\min_W r(W)+\langle\nabla g_V(T),W-T\rangle
                   +\tfrac M2\|W-T\|_F^2 .
\]
Assume the proximal subproblem attains its minimum. Comparing to $W=T$ gives \eqref{eq:block-dec} by the smooth descent lemma. Fermat's rule gives $v=-\nabla g_V(T)-Ms$, hence \eqref{eq:block-res} with $b=L_B+\bar M$. For weakly convex $h$, $M>\rho_h/2$ also ensures uniqueness. A finite backtracking search can majorize the smooth term with a strict margin on a bounded region where a common Lipschitz bound holds. A rejected accelerated proposal may be replaced by this step from the old factor. No inertia is carried across accepted block changes in the theorem below.

\begin{proposition}[Exact block solves can stop outside joint stationarity]\label{prop:kink-stall}
Let $n=r=1$, $h=0$, $f(z)=|z-2|$, and $\gamma=3/4$. Every joint critical pair has equal factors. Exact alternating minimization from $x^0=y^0=1$, updating $x$ first, reaches $(2,1)$ after one sweep and stays there.
\end{proposition}
\begin{proof}
Every loss subgradient satisfies $|G|\le1$. Inequality~\eqref{eq:spectral-margin} is strict since $2\gamma=3/2>1$. With $y=1$, the convex block objective $|x-2|+3(x-1)^2/8$ has unique minimizer $x=2$. With $x=2$, the convex block objective $|2y-2|+3(2-y)^2/8$ has unique minimizer $y=1$. At the kink, stationarity in the first coordinate requires $G=-3/4$ and in the second requires $G=3/8$. These both belong to $[-1,1]$, but no common selection exists. The objective is semi-algebraic and hence KL; its blockwise stopping point is not jointly stationary.
\end{proof}
This example isolates the smooth-coupling hypothesis in the next theorem. Even the strict general penalty bound and exact inner solutions do not replace the common endpoint residual condition for a nonsmooth coupled loss.

\section{Whole-Sequence Convergence for Fixed Penalty}
\label{sec:kl-fixed}
Let $z^k=(X^k,Y^k)$ denote a completed sweep, $z^{k+1/2}=(X^{k+1},Y^k)$ its intermediate point, and $t_k=\|z^{k+1}-z^k\|$ the product Frobenius step length. Put $p(X,Y)=f(XY^\top)+\gamma\|X-Y\|_F^2/2$.

\begin{theorem}[Fixed-$\gamma$ inexact alternating displacements]\label{thm:fixed-kl}
Fix $\gamma>0$. Assume $f$ is differentiable and transpose symmetric, its gradient is locally Lipschitz, and $h$ is proper lsc. Start with finite objective, assume $F_\gamma$ bounded below along the algorithm, and assume the generated factors are bounded. Suppose $h$ is continuous relative to its effective domain and cluster factors belong to that domain, or instead that $h$ is weakly convex. Each block satisfies \eqref{eq:block-dec}--\eqref{eq:block-res} with uniform $a,b$. Assume $F_\gamma$ is a Kurdyka--\L{}ojasiewicz (KL) function on the cluster set. Then
\[
 \begin{gathered}\sum_{k=0}^\infty t_k<\infty,\qquad z^k\longrightarrow z^*,\\
 0\in\partial F_\gamma(z^*),\quad F_\gamma(z^k)\longrightarrow F_\gamma(z^*).\end{gathered}
\]
Intermediate points have the same limit. If an exactness condition from Section~\ref{sec:framework} holds at every potential unequal limit, the factors agree and their common value is stationary for $\Phi$. Semi-algebraic $F_\gamma$ supplies the KL hypothesis, including polynomial losses with nonnegative or box indicators. All constants and objective values in this theorem use one fixed penalty.
\end{theorem}
\begin{proof}
\emph{Descent.} Adding the two block decreases gives $F_k-F_{k+1}\ge a t_k^2$, where $F_k=F_\gamma(z^k)$. Thus $F_k\downarrow\ell$ and $t_k\to0$.

\emph{A joint endpoint subgradient.} Let $u_X,u_Y$ be the two accepted block residuals, measured at $z^{k+1/2}$ and $z^{k+1}$ respectively. At the completed sweep,
\[
 \begin{gathered}v_c=\nabla_Xp(z^{k+1})-\nabla_Xp(z^{k+1/2}),\\
 w^{k+1}=(u_X+v_c,\ u_Y)
 \in\partial F_\gamma(z^{k+1}).\end{gathered}
\]
This uses the smooth sum rule and separability of $h(X)+h(Y)$. On the bounded factor region the mixed gradient is Lipschitz with constant $L_c$ (enclose the segments in a compact ball). The two residual tests and the triangle inequality give $\|w^{k+1}\|\le (b+L_c)t_k=:C t_k$; enlarge $C$ to a positive constant if necessary.

\emph{Cluster values and stationarity.} The set of cluster points is compact and nonempty, the distance to it tends to zero, and $w^k\to0$. Relative continuity gives value convergence on any convergent subsequence in the first case. In the weakly convex case, bounded subgradients of $h$ at each endpoint follow by subtracting the bounded smooth gradients from $w^k$. A proper lsc weakly convex function has a quadratic-affine minorant. Hence both regularizer values are bounded below on a bounded set, and the bound on their sum gives individual upper bounds. Lower semicontinuity puts the subsequential limits in the domain. Applying the weak subgradient inequality from the endpoints to their limit gives the reverse limsup inequality for each $h$ value. Thus all cluster points have objective $\ell$. Value-attentive closedness of the limiting subdifferential gives stationarity at each cluster point.

\emph{Finite length.} The uniform KL inequality on the compact cluster set gives an increasing concave desingularizer $\varphi$, $\varphi(0)=0$, and $\varphi'(F_k-\ell)\dist(0,\partial F_\gamma(z^k))\ge1$ for all sufficiently large $k$ with $F_k>\ell$. Uniformization follows from a finite cover, the minimum of the local value radii and the sum of the corresponding desingularizers~\cite{abs2013framework,bolte2014palm}. Write $u_k=F_k-\ell$. If a gap reaches zero, all later steps vanish. Otherwise the endpoint residual bound gives $t_{k-1}>0$ and concavity yields
\[
 \varphi(u_k)-\varphi(u_{k+1})
 \ge \frac{a\,t_k^2}{C\,t_{k-1}} .
\]
By the arithmetic--geometric mean inequality,
$2t_k\le t_{k-1}+(C/a)(\varphi(u_k)-\varphi(u_{k+1}))$.
Summation bounds every tail partial sum by $t_{K-1}+(C/a)\varphi(u_K)$ for sufficiently large $K$. Completeness gives convergence of $z^k$ to one critical point. Intermediate points lie within $t_k$ of the full endpoints. The conclusion about equal factors follows from Theorem~\ref{thm:signed-curv}.
\end{proof}
The proof verifies sufficient decrease, relative error and value continuity in the abstract descent framework of Attouch, Bolte and Svaiter~\cite{abs2013framework}. The block residual tolerances control the joint subgradient at the end of each sweep. The successive anchors of one shared implementation can alternate between two different factors when the exactness margin fails.

\begin{corollary}[Finite-work agreement certificate]\label{cor:inexact-margin}
Suppose the signed directional margin of Theorem~\ref{thm:signed-curv} is at least $m_*>0$ at the non-diagonal endpoints in question. Then
\[
 \|X^{k+1}-Y^{k+1}\|_F\le \sqrt{2}\,C t_k/m_* .
\]
If this bound on the margin holds at all non-diagonal endpoints, at least one of the first $K$ endpoints satisfies
\[
 \|X^{k+1}-Y^{k+1}\|_F\le
 \frac{\sqrt2 C}{m_*}\sqrt{\frac{F_0-\inf F_\gamma}{aK}}
\]
when the displayed infimum is finite. For a specified endpoint, replace the theoretical mixed-gradient constant by the measured drift from the first to the second half-step and the two accepted residuals in the construction of $w^{k+1}$.
\end{corollary}
\begin{proof}
Use $\|w_X-w_Y\|_F\le\sqrt2\|w\|$ in Theorem~\ref{thm:signed-curv}. Sum the descent bound and take the least $t_k$ among the first $K$. A diagonal endpoint already has zero disagreement.
\end{proof}
The output bound addresses agreement. It can be paired with the separate original-problem residual of Appendix~\ref{sec:finite-accuracy}; a small disagreement alone does not imply that the midpoint minimizes the PSD objective.

\section{Experiments}
\label{sec:experiments}
We test the mask classification, compare the one-sided and symmetric update models on partially observed Gram matrices, and measure the cost of inexact row solves.

\subsection{Masked similarity learning on handwritten digits}
\label{sec:digits-fiber}
We used the 1,797-example $8\times8$ optical handwritten-digit set distributed with scikit-learn, the test portion of the UCI collection~\cite{ucidigits}. For six seeds we sampled 24 images from each of ten classes, normalized the 64 pixel intensities to unit Euclidean norm, and formed the cosine Gram matrix. We observed all diagonals and randomly retained either 10\% or 30\% of the unordered off-diagonal pairs, using nested masks within seed. Labels were used for stratification and retrieval evaluation only. Timing excludes data generation and off-line held-out evaluation. The digit-kernel protocols used one BLAS thread on an AMD EPYC 9V74, binary weights, and $\gamma=0.5$. The ancillary files provide trial-level results, the data hash, package versions, and the executable protocols.

\paragraph{Unrestricted factors: predicted versus observed obstruction.}
Across the six masks at 10\%, $-\lambda_{\min}(W\odot A)/2$ ranged from $2.684$ to $2.847$. At 30\% it ranged from $4.238$ to $4.585$. Thus all twelve matrices predict a nonzero opposite critical pair at the chosen penalty, despite having PSD targets. In each case we minimized the antipodal rank-one objective from a scaled minimum eigenvector. Its value fell below the origin and the relative single-block gradient norm at the resulting opposite pair was at most $2.4\times10^{-7}$, using the original block gradients. Appending zero columns embeds these witnesses at any greater width.

For each of the twelve matrices we also ran 40 exact rank-one AM sweeps, with $\theta=0$ or $\theta=\theta_*+0.1$ from Corollary~\ref{cor:hinge-branch}. The starting pairs were the computed witness, the witness with a random perturbation of relative norm 1\% in $Y$, and an aligned positive pair. The median first-sweep displacement relative to the starting factor norm was $2.04\times10^{-8}$ versus $1.52$ at the witness, and $0.00991$ versus $1.53$ with the perturbation. On all 12 positive starts the hinge remained inactive throughout; the two methods gave identical iterates. All 12 unmodified runs from the approximate witnesses also exited negative trace by sweep 40. These tests establish a detectable change at the predicted critical branch; they give no evidence that generic AM starts are trapped there. The zero midpoint of an opposite pair has zero symmetric gradient even when the reconstructed kernel is poor, illustrating why a stationarity tolerance alone cannot characterize recovery quality.

\paragraph{Nonnegative factors: time and retrieval accuracy.}
On the same matrices we fitted rank eight, with matched positive initial factors. The trace hinge vanishes on this domain. We compared projected alternating block steps (Alt), the same steps plus the exact safeguarded contraction (Alt+C), and direct symmetric projected gradient with backtracking (Sym). All three methods check at every sweep the original midpoint residual
\[
 \|U-[U-2(W\odot(UU^\top-A))U]_+\|_F/\max(1,\|U\|_F)
 \le 0.02 .
\]
One sweep comprises both block updates or one symmetric update. Method order rotates across seeds. The first experiment used the spectral block step of Proposition~\ref{prop:fiber-iterate} with a cap of 500 sweeps. Both alternating variants reached the cap on all twelve matrices. Sym reached tolerance on two of six runs at 10\% and one of six at 30\%. Each capped output is included in the trial records.

A second experiment raised the cap to 5,000 sweeps and used a rowwise diagonal majorizer for both alternating methods. For nonnegative $Y$ the block row Hessian $K_i$ has nonnegative entries. The diagonal matrix of its row sums majorizes $K_i$ since their difference is a weighted graph Laplacian. We scale each coordinate by $0.9$ times the inverse of this diagonal before projection. On the bounded sublevel the scaling matrices are bounded above and below, so the sufficient-decrease and projected-stationarity proof of Proposition~\ref{prop:fiber-iterate} also applies in this diagonal metric. We kept the symmetric method, sampled matrices, starts, tolerance, and task metrics identical across protocols. Both the sweep limit and the majorizer differ between protocols, so their individual effects on the success rate cannot be identified.

\begin{table}[t]
\centering
\caption{Nonnegative digit-kernel reconstruction with a 5,000-sweep limit. Each cell is the median of six paired runs. All 36 runs reached the same residual tolerance. P@5 is the fraction of same-digit labels among the five highest-scoring unobserved neighbors per image.}
\label{tab:digits-fiber}
\small\setlength{\tabcolsep}{3pt}
\begin{tabular}{llrrrr}
\toprule
Observed & Method & Sweeps & Time (s) & Test RMSE & P@5\\
\midrule
10\% & Alt & 449.5 & .193 & .0973 & .342\\
     & Alt+C & 449.0 & .345 & .0972 & .339\\
     & Sym & 537.5 & .258 & .0965 & .364\\
30\% & Alt & 677.5 & .283 & .0301 & .751\\
     & Alt+C & 677.0 & .501 & .0301 & .751\\
     & Sym & 627.5 & .304 & .0299 & .751\\
\bottomrule
\end{tabular}
\end{table}

The contraction collapsed the difference on a median of 448 and 673 sweeps at the two densities. Repeated collapse barely changed the iteration count and increased the recorded time. The uncorrected block method had the lowest median time here; direct symmetric optimization had the lowest median test RMSE at both densities. Wall times concern these dense, single-thread implementations and include every residual check and correction. Six paired samples from one image collection cannot establish a general solver ranking.

\paragraph{Checks of the classification.}
We exhaustively tested the mask PSD criterion against the block characterization on all $1,024$ five-vertex binary masks with observed diagonal and all $1,024$ four-vertex masks with free diagonal. In each enumeration exactly 52 masks passed both tests. We separately checked the scalar quartic and work identities on 200 random instances, and the closed-form block KKT equations at ranks one through four in 150 cases covering the three hinge regimes equally. These computations check formulas, not stationarity at all points of a continuous space.

\subsection{Descent at small one-sided steps}
\label{sec:robust-checkpoints}
We tested the absolute-loss model with $h=0$ on complete rank-one fits to Gram matrices from UCI optical digits and Wine~\cite{ucidigits,uciwine}. No labels entered the fit. We used integer pixel intensities for digits. For Wine we min--max scaled each feature over the collection and rounded it to one of 257 levels. We divided the integer Gram matrices by $4096$ and $262144$, respectively. The inputs and reported factors are therefore amenable to exact arithmetic checks on their binary representations. A second condition independently corrupts each unordered off-diagonal entry with probability $0.1$ by adding $+0.5$ or $-0.5$ with equal probability and mirroring the change. Clean targets are PSD. The corrupt targets need only be symmetric.

\paragraph{An exact two-object test.}
For a two-object principal submatrix with diagonal $a,c$ and off-diagonal $b>0$, set both anchor entries to $\sqrt b$. The one-sided zero step is optimal for every $\gamma>0$: the free sign on each fitted off-diagonal entry can cancel its row's diagonal sign. The symmetric objective has a strict descent direction exactly when $\min(a,c)<b<\max(a,c)$. Indeed the two symmetric row equations require one shared off-diagonal sign; it exists precisely when the two diagonal subgradient intervals intersect. Integer Gram comparisons found this strict obstruction in $42{,}423/1{,}613{,}706$ digit pairs and $7{,}680/15{,}753$ Wine pairs. Each count describes an explicitly specified two-object initialization. On unit-diagonal kernels this particular obstruction is impossible, because the diagonals are equal. The counts do not estimate a solver's failure probability.

\paragraph{Stopping points reached from spectral initialization.}
For each collection, condition and $n\in\{32,64\}$ we sampled ten subsets using seeds 0--9. Within each condition all comparisons use the same leading-eigenpair start and $\gamma=n/4$. At each symmetric anchor $V$ we computed the one-sided rank-one block by sorting row breakpoints $a_{ij}/v_j$ and solving the resulting scalar optimality conditions. We backtracked from the unit step on $\Phi$ and stopped at relative block step $10^{-10}$, at the actual decrease floor $10^{-12}\max(1,\Phi)$, or after 300 proposals. This trajectory keeps a single symmetric anchor. All 80 runs reached the small-step test. Exact integer comparisons of binary inputs, including the interval of admissible row subgradients, found zero mathematically optimal block steps and zero exactly fitted off-diagonal entries at these checkpoints. In contrast, the prespecified band $10^{-10}\max(1,\|A\|_{\max})$ contained a median of 35.5 unordered off-diagonal entries. Separately, ordinary split AM from the same starts reached the two-block change tolerance on all 80 runs. None reached relative disagreement $10^{-10}$ (median $0.0107$). The checkpoint tests below concern the symmetric-anchor trajectory.

\begin{table}[t]
\centering\small\setlength{\tabcolsep}{3pt}
\caption{One extra direction at the same 80 checkpoints. The indicated bandwidth is scaled by $\max(1,\|A\|_{\max})$. Gain is relative to the pre-checkpoint objective. The final column gives the median guaranteed step from \eqref{eq:finite-band-step}, capped at one for a unit direction.}
\label{tab:band-stop}
\begin{tabular}{lrrr}
\toprule
Band & Gain $\ge10^{-6}$ & Gain (\%) & Step bound\\
\midrule
$0$          & 0/80 & 0       & $1.29\,10^{-17}$\\
$10^{-14}$   & 55/80 & 0.00200 & $4.95\,10^{-5}$\\
$10^{-10}$   & 78/80 & 0.00482 & $3.74\,10^{-4}$\\
$10^{-6}$    & 80/80 & 0.00501 & $3.74\,10^{-4}$\\
\bottomrule
\end{tabular}
\end{table}
For each band we minimized the norm of the trial gradient over its sign box. Only at zero bandwidth are all admissible signs exact subgradient signs. We normalized the resulting direction and allowed 40 halvings with the same actual decrease floor. An independent dyadic calculation checked that all 320 trial slopes were strictly negative. With exact signs, only one search met the decrease floor, and none improved the objective by $10^{-6}$ relatively. The prespecified band gave measurable decreases at all 80 points (Table~\ref{tab:band-stop}). Setting every free sign to zero instead attained the $10^{-6}$ threshold on 15/80. The median full box check took 0.592 ms on one BLAS thread, compared with 1.881 ms to reach the checkpoint. On the 40 corrupted cases, the band correction increased the error to the clean kernel in 11 cases despite reducing the fitted loss. These are objective-descent measurements, not evidence of improved recovery.

\begin{table}[t]
\centering\small\setlength{\tabcolsep}{3pt}
\caption{Sign selection at 80 matched checkpoints (seeds 100--109). Timing starts at the common checkpoint. The last column gives median additional relative gain within 20 ms, allowing at most eight correction rounds.}
\label{tab:holdout-stop}
\begin{tabular}{lrrr}
\toprule
Sign selection & Certified & First call (ms) & Gain (\%)\\
\midrule
Free signs set to zero & 34/80 & 0.028 & 0\\
Projected + fallback & 80/80 & 0.144 & 0.0266\\
Full box least squares & 80/80 & 0.495 & 0.0277\\
\bottomrule
\end{tabular}
\end{table}
On subsets drawn with seeds 100--109, we compared three sign choices at the same bandwidth. The projected variant starts at zero, takes at most eight box-projected gradient iterations with step $1/(\|M\|_1\|M\|_\infty)$ when this denominator is positive, tests Proposition~\ref{prop:band-descent} with $\theta=1/2$, and falls back to full bounded least squares if needed. All methods use the same actual-value backtracking and resume ordinary one-sided proposals between corrections. We rotated method order across instances. The early test succeeded without a full solve on 75/80 first calls and 373/640 calls overall. The full check obtained a slightly greater median decrease under this budget (Table~\ref{tab:holdout-stop}). The speed difference concerns small rank-one stopping checks. Section~\ref{sec:masked-rank} tests masked fits at higher rank against a direct prox-linear solver.

\subsection{Masked absolute loss at ranks two and four}
\label{sec:masked-rank}
Does the stopping check improve a fit when similar objects have unobserved entries, and is the one-sided model competitive with a direct symmetric update? We used the same two feature collections and Gram scaling as in Section~\ref{sec:robust-checkpoints}. Each problem has 64 objects, rank two or four, and off-diagonal observation probability $p=0.25$ or $0.60$. Every diagonal is observed. For four subset seeds (200--203), we tested clean observed similarities and a condition that independently changes observed off-diagonal pairs by $\pm0.5$ with probability 0.1. This gives 64 matched configurations. The leading positive eigenpairs of the observed matrix, with off-diagonals divided by $p$, supply a shared start. All scores on hidden entries use the clean Gram matrix and exclude the diagonal. No hidden value enters initialization, fitting, stopping or step selection.

\paragraph{A common optimization budget.}
We compared the models $m_o,m_s$ of \eqref{eq:two-absolute-models} on a single symmetric anchor. Each uses $\lambda=|\Omega|/64$, counting ordered observations, and warm-started accelerated projected ascent on its dual box. We used a common dual-gradient Lipschitz bound $4\|V\|_2^2/\lambda$, 120 inner iterations at most and a primal--dual tolerance of $10^{-5}\max(1,\Phi(V))$. A common halving search accepts a strictly measurable reduction of the observed objective. We also tested two additions to $m_o$: the fixed-band check from Section~\ref{sec:robust-checkpoints}, and the finite-radius check in Section~\ref{sec:zero-step}. A check was eligible after two successive relative decreases of at most $10^{-3}$ or a rejected proposal. Each run could check up to six times and continued taking ordinary steps after these checks. The radius variant tested $\max(1,\|V\|_F)$ times $10^{-2},10^{-3},10^{-4},10^{-5}$. All methods used the same 48 outer rounds; independent readouts also retained the last completed update within 100 or 250 ms of a common start. Setup and evaluation of the hidden entries were outside each solver timer. The implementation used dense $64\times64$ masked inner products and a single BLAS thread.

\begin{table}[t]
\centering\small\setlength{\tabcolsep}{2.5pt}
\caption{Masked fits, 64 matched configurations. Observed and hidden columns give the median paired percentage reduction in mean absolute error versus the unguarded one-sided solver at 48 rounds. Time is the median elapsed time for 48 model calls. Split AM reports its midpoint after 24 sweeps.}
\label{tab:masked-rank}
\begin{tabular}{lrrr}
\toprule
Method & Fit (\%) & Test (\%) & ms\\
\midrule
One-sided & 0 & 0 & 168.6\\
Split AM (midpoint) & 0.3418 & 0.8003 & 167.8\\
Fixed band & 0.0123 & 0.0050 & 172.8\\
Radius test & 0.0086 & 0.0179 & 184.8\\
Prox-linear & 1.8086 & 0.8766 & 203.3\\
\bottomrule
\end{tabular}
\end{table}
Relative to the one-sided baseline, the symmetric prox-linear update gave smaller training error on 63/64 configurations and smaller hidden-entry error on 42/64 at 48 rounds (Table~\ref{tab:masked-rank}). With just 100 ms it improved the two metrics on 61/64 and 45/64, with median paired reductions of 3.32\% and 2.66\%. At 250 ms the respective counts were 63/64 and 42/64. At 48 rounds the fixed and radius checks each took a decreasing extra step on 258/258 attempts across 53 problems. The checks accounted for 0.66\% and 8.18\% of aggregate time in their respective solvers. Some resulting trajectories ended worse than the plain solver even on observed entries. A local decrease does not order subsequent iterates.

\paragraph{Ordinary two-factor AM.}
We also ran the original alternating updates on these 64 cases, accepting a candidate against the split objective at the previous free factor. This control maintains distinct factors and reports their midpoint after each full sweep. After 48 block calls (24 sweeps) its median relative factor disagreement was 0.0821. The midpoint improved observed and hidden MAE over the single-anchor baseline on 44/64 and 42/64 problems. In the direct paired comparison, prox-linear had the smaller observed MAE on 62/64 and the smaller hidden MAE on 35/64. Thus the difference on unobserved entries is much less decisive against ordinary split AM. The AM timing was collected separately with the same single-thread settings.

\paragraph{Effect of inner accuracy.}
The main comparison measures bounded inner work. Only 9 of 3072 one-sided calls and 18 of 3072 symmetric calls met the dual tolerance. We therefore repeated a matched 24-round comparison on every seed-200 configuration (16 problems), with inner caps of 120 and 1440. The counts meeting tolerance increased from 0/384 for each method to 247/384 and 152/384, respectively. The symmetric solver had lower observed error on 16/16 at both caps. Its hidden error was lower on 15/16 with the smaller cap and 14/16 with the larger one.

We also compared the two models at the \emph{same} starting factor and again at the same factor after 24 ordinary one-sided rounds. At all 32 anchors, the feasible symmetric comparison step ruled out symmetric model accuracy $10^{-4}\max(1,\Phi(V))$ for the computed one-sided proposal. The lower bound on its symmetric suboptimality had median 5.81\% of $\max(1,\Phi(V))$ at initialization and 1.05\% after 24 rounds. Using the separate primal--dual gaps to bound both optimizer errors, the lower bound on the distance between the two \emph{exact} steps was positive on all 32 anchors (minimum $0.0206$ in Frobenius norm). This check isolates the model discrepancy from the inexact solves. The data support the local descent certificate. They favor the direct prox-linear baseline on these small fits and leave comparisons with optimized sparse or parallel row solvers open.

\subsection{Paired test of certificate acceptance}
\label{sec:margin-comparison}
We compared the two margins at the \emph{same} exact-AM endpoint, with $h=0$, $\gamma=0.5$ and no penalty changes. We stopped each trajectory the first time both residuals satisfied $\|e_\pm\|_F\le10^{-5}=\tau$. We then scored the directional rule $2\gamma-q_D\ge\underline m$, the retained-pairing rule \eqref{eq:directional-margin-adaptive}, and the direct test $\|D\|_F\le\tau/\underline m$. The thresholds were 0.25, 0.75 and 1.25, with seeds $0,\ldots,11$. All 36 trajectories reached the common checkpoint within the 2500-sweep budget.

\begin{table}[t]
\centering
\caption{Accepted endpoints out of 12 per family. Each triple gives the counts for the directional margin without transpose curvature, the margin with that curvature, and the direct disagreement test, respectively.}
\label{tab:margin-comparison}
\small
\begin{tabular}{lccc}
\toprule
Minimum margin & $0.25$ & $0.75$ & $1.25$\\
\midrule
Four-cycle & 0/12/12 & 0/12/12 & 0/0/7\\
Masked & 12/12/12 & 12/12/12 & 0/5/6\\
Top-$r$ & 12/12/12 & 12/12/12 & 0/0/11\\
\bottomrule
\end{tabular}
\end{table}

The four-cycle uses the fixed positive definite instance from Section~\ref{sec:consensus-geometry}, varying the initial displacement along $v$; its 12 rows of raw data represent starts on one constructed instance. The masked group uses independent $36$-by-$36$ Gaussian Gram instances of rank three, observes each off-diagonal pair with probability $0.35$, and includes the diagonal. The Top-$r$ group uses rank-three factors and 36-dimensional PSD matrices with independent random eigenbases, leading eigenvalues $(3,2,1.5)$, and remaining eigenvalues from $0.75$ down to $0.05$. The ancillary files include the generator and all recorded endpoints.

At the strictest margin, including transpose curvature certified five additional masked endpoints. It did not change any Top-$r$ decision, while 11 Top-$r$ endpoints already satisfied the direct disagreement bound. In all three families the direct test accepted at least as often, as $\|e_-\|_F\le\tau$ and any accepted positive margin implies $\|D\|_F\le\tau/\underline m$. For $h=0$ the direct test is enough if the target is factor disagreement. For shared nonsmooth $h$, the retained margin also bounds the subgradient enlargement in \eqref{eq:epsilon-h-residual-bound}. Each comparison uses the same recorded endpoint and fixed penalty.

\subsection{Verification of the inexact block criterion}
We used five generated sparse weighted least-squares examples ($n=48$, $r=7$, off-diagonal sampling probability $0.22$, observed diagonal). Each example ran 40 fixed-penalty sweeps with unconstrained factors, $\gamma=0.7$ and $\eta=0.15$. Independent row-CG recurrences started from the old free factor; the first iterate satisfying the aggregate true block residual test was accepted. Across 400 block updates the median inner count in each instance was three and the maximum was three. All decreases exceeded $(\gamma/2-\eta)\|s\|_F^2$, and the largest residual-to-motion ratio was $0.139$ (rounded). This is a check of the stopping certificate on small systems. It measures neither runtime improvement over Cholesky nor convergence rates on imaging data.

\subsection{Time to a common tolerance with inexact row solves}
\label{sec:sparse-inner}
For sparse quadratic loss, which part of a one-sided update saves time? We used unit-normalized Digits images and fitted their cosine similarities on a regular graph. Each image had 24 observed off-diagonal neighbors plus its diagonal; eight other neighbors were held out. Disjoint shifts of a random cyclic ordering generated the symmetric masks. No labels entered the experiment. For each row, let $B$ contain the observed neighbor factors, $a$ their target similarities, and $v$ the fixed factor row as a column vector. The fitted row is $v+\Delta$. Its exact displacement has two implementations:
\begin{equation*}
\begin{split}
 \Delta&=(B^\top B+\gamma I)^{-1}B^\top(a-Bv)\\
       &= B^\top(BB^\top+\gamma I)^{-1}(a-Bv).
\end{split}
\end{equation*}
Thus $r$ and the observed degree determine the two direct system sizes. A third implementation applies $p\mapsto B^\top(Bp)+\gamma p$ through the observed edges in Jacobi-preconditioned conjugate gradients (PCG). This avoids the row Gram matrices. All methods in this study solve the same fixed-penalty split problem and start a block from its old free factor. The PCG test is $\|g_{\rm block}\|_F\le c\gamma\|s\|_F$, with $s$ the actual change across the call.

\paragraph{Accuracy and penalty.} We crossed $n\in\{256,1024\}$, $r\in\{8,16\}$, five paired seeds and fixed $\gamma\in\{0.5,4\}$. Each of four implementations ran for 20 sweeps: batched Cholesky, a second batched direct solve, and Gram-based PCG with $c=0.05$ or $0.25$. Timers included row-system construction, solves and the common online check $R(U)=\|\nabla\Phi(U)\|_F/\max(1,\|\nabla\Phi(U_0)\|_F)$ at the midpoint; data setup, independent edge-based block verification and held-out scoring were outside all timers. All 160 runs reached $R\le0.10$. At $\gamma=0.5$ all four methods reached $R\le0.02$ in all 20 cases; PCG at $c=0.25$ took a median $1.02$ times the Cholesky time to this tighter target. At $\gamma=4$ the median of the per-run mean PCG counts fell from $2.93$ to $1.00$ for $c=0.25$. Both PCG variants missed the tighter target on all 20 cases; each direct implementation reached it once. Computing a cheaper block step can come with slower progress in the original stationarity measure.

\begin{table}[t]
\centering
\caption{Observed-edge PCG time divided by each comparator's time to $R\le0.10$ (median of five paired ratios). All five pairs reached this target in each cell. The last column gives the median edge-PCG time in milliseconds. The two direct solves use general batched linear solves in dimensions $r$ and 25, respectively.}\label{tab:edge-pcg}
\small
\begin{tabular}{rrcccc}
\toprule
$r$ & $\gamma$ & $r$-direct & 25-direct & Gram PCG & ms\\
\midrule
16 & 0.5 & 0.92 & 0.51 & 1.04 & 6.5\\
16 & 4.0 & 0.73 & 0.37 & 0.99 & 19.2\\
32 & 0.5 & 0.49 & 0.61 & 0.71 & 8.9\\
32 & 4.0 & 0.35 & 0.41 & 0.58 & 22.9\\
64 & 0.5 & 0.21 & 0.75 & 0.36 & 14.9\\
64 & 4.0 & 0.16 & 0.51 & 0.30 & 37.2\\
\bottomrule
\end{tabular}
\end{table}

\paragraph{Varying rank at fixed degree.} We fixed $n=512$ and varied the rank over $16,32,64$, using seeds 5--9, both penalties, and 16 sweeps. We compared both direct formulations with Gram-based and observed-edge PCG at $c=0.25$. At $r=64$ and $\gamma=0.5$, charging the stronger observation-space direct solve reduces the apparent gain: the time ratio is $0.75$ against that solve and $0.21$ against the factor-space solve (Table~\ref{tab:edge-pcg}). At $r=16$ Gram-based PCG was comparably fast. At the tighter target, edge PCG succeeded on 23 of 30 conditions and each direct method on 25. Each implementation pair (the two exact solves, or the two PCGs) agreed in its original residual ratios to within $10^{-9}$ throughout. At $r=64$, $\gamma=0.5$, the final median held-out RMSE was $0.0855$ for PCG and $0.0857$ for the direct solves. The time differences are clearer than the changes in prediction error.

Across the two experiments, independent checks of all 10,240 block updates satisfied the applicable decrease and residual tests up to numerical precision; neither PCG implementation needed its direct fallback. These timings characterize one-thread NumPy implementations on an AMD EPYC 9V74, one image collection, and regular degree-25 observations. They do not compare split AM with sBMF-GN, and they do not establish fixed-penalty exactness on these masks.

\subsection{Full-Observation \texorpdfstring{Top-$r$}{Top-r} PSD Approximation}
Consider the problem
\[
 \min_{W\in\mathbb R^{n\times r}}
 \Phi(W)=\tfrac12\|A-WW^\top\|_F^2,\qquad A\succeq0.
\]
Table~\ref{tab:topr-results} uses the trial data of Hu et al.~\cite{hu2026abmf}. At each of two matrix sizes, four methods were run on five paired trials. The means and sample standard deviations of runtime are computed from these 40 records.

All methods receive the same matrix--block multiplication interface and matched initial factor. Planted eigenvectors are not exposed to the solvers. An independent reference eigenspace is computed after the solver runs. Every final output is evaluated using the common factor $U=(X+Y)/2$, with $X=Y$ for symmetric methods. The relative objective gap is
\[
 \frac{|\Phi(U)-\Phi_\star|}{\max\{1,|\Phi_\star|\}},
\]
and the subspace error is $\|(I-P_U)Q_*\|_F$, where $P_U$ projects onto the numerical range of $U$ and $Q_*$ is an orthonormal basis for the reference subspace.

\begin{table*}[t]
\centering
\caption{Top-$r$ PSD approximation, using the runs of Hu et al.~\cite{hu2026abmf}. Times are means $\pm$ sample standard deviations over five paired trials. Errors and penalties are means. For AM-adp, the last column gives the penalty after the final update. For AM-mix, it gives the penalty used in the final sweep. Each solver uses its own stopping rule.}
\label{tab:topr-results}
\small
\setlength{\tabcolsep}{6pt}
\begin{tabular}{llrrrr}
\toprule
$(n,r)$ & Method & Time (s) & Relative objective gap & Subspace error & Final $\gamma$\\
\midrule
$(10^4,50)$
 & sBMF-GD & $3.979\pm0.945$ & $1.98\times10^{-5}$ & $1.25\times10^{-2}$ & --\\
 & sBMF-GN & $0.472\pm0.088$ & $2.61\times10^{-7}$ & $1.45\times10^{-3}$ & --\\
 & AM-adp & $1.551\pm0.325$ & $3.28\times10^{-6}$ & $5.16\times10^{-3}$ & $33.77$\\
 & AM-mix & $0.678\pm0.098$ & $4.62\times10^{-7}$ & $1.95\times10^{-3}$ & $6.90$\\
\midrule
$(5\times10^4,1000)$
 & sBMF-GD & $2332.627\pm211.053$ & $1.99\times10^{-5}$ & $5.57\times10^{-2}$ & --\\
 & sBMF-GN & $225.145\pm19.155$ & $2.85\times10^{-7}$ & $6.68\times10^{-3}$ & --\\
 & AM-adp & $2586.158\pm249.521$ & $1.48\times10^{-5}$ & $4.82\times10^{-2}$ & $143.91$\\
 & AM-mix & $738.091\pm59.289$ & $2.79\times10^{-6}$ & $2.10\times10^{-2}$ & $30.51$\\
\bottomrule
\end{tabular}
\end{table*}

AM-mix uses the practical penalty update of Hu et al.~\cite{hu2026abmf}. It is faster than AM-adp in both configurations. The specialized sBMF-GN solver has the lowest mean time, objective gap and subspace error in both.

The benchmark uses relative objective change for the symmetric methods and additionally tests factor disagreement for the asymmetric methods. The reported times therefore correspond to different endpoint criteria. Solver timers exclude the independent reference calculation, and initialization caching is method-dependent. MATLAB R2020a and the Windows platform are recorded, but processor and thread metadata are absent.

\paragraph{Numerical checks of the identities.}
With seed 20261118 we checked 120 random quadratic-loss instances with shared $\ell_1$-plus-ridge regularization or nonnegative constraints, and 30 rank-one weighted-graph instances. All certificate inequalities held. The largest relative error in the midpoint identity was $4.35\times10^{-16}$, and the largest discrepancy in the graph-spectrum identity was $1.11\times10^{-15}$. For unrestricted factors the checks covered 120 general Gram identities, 120 least-squares cubic and sublevel bounds, and 20 shifted-subspace bounds. The scripts and their outputs accompany the trial data.

\section{Discussion}
\label{sec:discussion}
Whether positive penalties suffice depends on the quantifiers. A fully observed PSD target guarantees agreement at every unrestricted critical point for every $\gamma>0$. Incomplete observations preserve that guarantee uniformly over targets precisely for the block masks in Theorem~\ref{thm:mask-universal}. For a given target outside that class, the input spectrum tests one explicit obstruction. Its absence leaves other unequal critical points possible. At a fixed midpoint Theorem~\ref{thm:fiber} instead compares every feasible disagreement, including active nonnegativity constraints. A stationary split may have a midpoint that is not stationary for the symmetric problem.

The trace hinge isolates a modeling decision. Two extensions can agree on every PSD prediction yet differ on negative-trace products; the minimal hinge coefficient follows from that difference. A projected factor model with nonnegative factors already has nonnegative trace. For this model a midpoint contraction respects feasibility, but the digit-kernel runs show why it is expensive to apply a correction at every sweep. Direct monitoring of the original projected-gradient residual and held-out prediction error remains necessary.

The choice of extension also matters at equal factors. With entrywise absolute loss, ordered entries on an exactly fitted pair permit two different subgradient signs. Moving a common factor ties those choices together. Corollary~\ref{cor:edge-slope} identifies the affected observations. The symmetric subgradient test can serve as a fallback when a block solver returns zero; its cost depends on how many observed residuals are exactly zero. In the rank-one tests in Section~\ref{sec:robust-checkpoints}, actual block steps were tiny, yet none of the returned points satisfied exact zero-block optimality. Enlarging the set of admissible signs gave directions with a larger guaranteed step and measurable decreases. The extra fit improvement was modest. At ranks two and four, the held-out entries in Section~\ref{sec:masked-rank} gave smaller median gains for these corrections than for a direct prox-linear update. The finite sign-conflict identity explains how the update models can differ even at equal factors. Scaling beyond these dense small instances remains open.

The opposite-factor examples used prescribed starting points. To estimate how often an alternating solver approaches them would require a distribution over initializations. The squared-loss timings also depend on the graph and the implementation. The direct comparisons at fixed degree and the fully observed Top-$r$ benchmark answer different computational questions.

\section{Conclusion}
Factor agreement depends on which entries of a PSD target are observed. For unrestricted least squares, a binary mask preserves agreement at all critical points for every target and every positive penalty exactly when it observes disjoint full principal blocks. For a given target outside this class, the smallest eigenvalue of the masked matrix determines whether opposite critical factors exist. At a computed pair, the difference of the block residuals and the curvature between a product and its transpose give a quantitative test for disagreement.

The same distinction matters in computation. For an absolute loss, a one-sided step can miss descent at a symmetric anchor. For smooth losses and fixed penalty, a relative residual test allows inexact block solves while preserving whole-sequence convergence under the hypotheses of Theorem~\ref{thm:fixed-kl}. At rank 64 and $\gamma=0.5$ on the degree-25 digit graphs, applying the row operators on observed entries took a median of 0.75 times the time of the observation-space direct solver to the same gradient tolerance. Irregular masks and larger data sets remain to be tested.

\section*{Acknowledgments}
This work was supported by the National Natural Science Foundation of China (No.~62266055).

\clearpage
\appendices

\section{Finite-Accuracy Guarantees for Symmetric Outputs}
\label{sec:finite-accuracy}
The spectral criterion in~\cite{hu2026abmf} assumes exact joint criticality. At a finite iterate, a small change in objective value does not certify either factor agreement or stationarity for the symmetric problem. We therefore retain the two residuals separately. For
\[
Z=XY^\top,\quad U=\frac{X+Y}{2},\quad D=X-Y,
\]
choose a \emph{common} $G\in\partial f(Z)$ and $H_X\in\partial h(X)$, $H_Y\in\partial h(Y)$, and define
\begin{align}
e_X&=GY+\gamma D+\tfrac12H_X,\nonumber\\
e_Y&=G^\top X-\gamma D+\tfrac12H_Y,\qquad
e_\pm=e_X\pm e_Y.                         \label{eq:residual-definition}
\end{align}
These are residuals at the \emph{same current pair}; stale gradients from different alternating half-steps do not define this certificate. The loss gradient is unique when $f$ is differentiable. For nonsmooth $f$, unrelated blockwise subgradients are not interchangeable with the common selection in \eqref{eq:residual-definition}.

\subsection{Consensus from the residual difference}
\begin{theorem}[Finite-residual consensus bound]
\label{thm:consensus-residual}
Suppose $f$ is finite, convex, transpose-symmetric, and $\sigma_f$-strongly convex, and $h$ is proper, closed, and $\sigma_h$-strongly convex, where $\sigma_f,\sigma_h\ge0$. For the selections in \eqref{eq:residual-definition}, put $S=\operatorname{sym}G$. Then
\begin{equation}
\begin{split}
&(2\gamma+\sigma_h/2)\|D\|_F^2-\langle D,SD\rangle\\
&\quad+\frac{\sigma_f}{2}\|Z-Z^\top\|_F^2
\le \langle D,e_-\rangle .
\end{split}                                      \label{eq:consensus-residual}
\end{equation}
Consequently, if
\begin{equation}
m_\sigma=2\gamma+\sigma_h/2-\lambda_{\max}(S)>0,
                                                        \label{eq:residual-margin}
\end{equation}
then
\begin{equation}
\|D\|_F\le \frac{\|e_-\|_F}{m_\sigma}.       \label{eq:consensus-bound}
\end{equation}
For $D\ne0$, the same conclusion holds with the possibly larger directional margin
\begin{equation}
m_D=2\gamma+\sigma_h/2-
       \frac{\langle D,SD\rangle}{\|D\|_F^2}>0
                                                        \label{eq:directional-margin}
\end{equation}
in place of $m_\sigma$.
\end{theorem}
The margin determines how an error in the residual difference propagates to factor disagreement. If the margin is nonpositive, the theorem is silent; it does not assert that disagreement must occur. A short power iteration supplies a parameter proposal but is not, by itself, a certified upper bound on $\lambda_{\max}(S)$. The directional alternative requires only the current quadratic form.

\paragraph{Curvature along the transpose difference.}
For Frobenius-loss SymNMF, the proof of Zhu et al.~\cite{snmf-nips} contains the term $-\|XY^\top-YX^\top\|_F^2/2$ (Eq.~(11) in arXiv:1811.05642). For general convex losses, the spectral estimate of Hu et al.~\cite[Lemma~1]{hu2026abmf} follows by dropping a nonnegative pairing. To evaluate this pairing, write $d=\|D\|_F$, $B=Z-Z^\top$, and, for $D\ne0$, define
\[
 q_D=\langle D,SD\rangle/d^2,\qquad
 b_f^{\rm tr}=\tfrac12\langle G-G^\top,B\rangle.
\]
Transpose symmetry and strong convexity~\cite[Assumption~2(c)]{hu2026abmf} give $b_f^{\rm tr}\ge\sigma_f\|B\|_F^2/2$. The smoothness constant $l_f$ in~\cite[Assumption~2(d)]{hu2026abmf} gives an upper bound for the same pairing. For example, $f(Z)=\tfrac12\|\operatorname{sym}Z\|_F^2$ has $l_f=1$ and $b_f^{\rm tr}=0$ for every $Z$, including nonsymmetric products.

At a joint critical point, \eqref{eq:consensus-residual} yields
\begin{equation}
 (2\gamma+\sigma_h/2)d^2+\tfrac12\sigma_f\|B\|_F^2
 \le \langle D,SD\rangle.
 \label{eq:transpose-critical-bound}
\end{equation}
For example, when $2\gamma+\sigma_h/2>0$, the upper bound on $d^2$ in~\cite[Theorem~1]{hu2026abmf} decreases by exactly $\sigma_f\|B\|_F^2/[2(2\gamma+\sigma_h/2)]$ at the same critical pair.

Define the two margins
\begin{equation}
\begin{aligned}
 m_{\rm curv}&=m_D+\frac{\sigma_f\|B\|_F^2}{2d^2},\\
 m_{\rm pair}&=m_D+\frac{b_f^{\rm tr}}{d^2},\qquad
 m_D=2\gamma+\frac{\sigma_h}{2}-q_D.
\end{aligned}
\label{eq:transpose-margins}
\end{equation}
The identity \eqref{eq:proof-loss-monotonicity} and strong monotonicity of $\partial h$ give
\[
 m_Dd^2\le m_{\rm curv}d^2\le m_{\rm pair}d^2
 \le\langle D,e_-\rangle.
\]
For either positive refined margin $m$, this proves $d\le\|e_-\|_F/m$. In particular, at a joint critical point any candidate with $D\ne0$ must satisfy
\begin{equation}
 \gamma\le\tfrac12 q_D-\tfrac14\sigma_h
             -\tfrac14\sigma_f\frac{\|B\|_F^2}{d^2}.
\label{eq:refined-necessary-offdiagonal}
\end{equation}
Its strict reverse excludes unequal critical factors at that candidate. To obtain a single penalty for a region one needs an upper bound on the entire right-hand side over its off-diagonal pairs. At $D=0$ the quotient is unnecessary.

The improvement depends on the factor geometry. Taking $Y=cX$ with $X\ne0$ and $c\ne1$ gives $B=0$ and $d>0$. Thus the best unrestricted lower bound on $\|B\|_F^2/d^2$ is zero. For rank one, putting $u=(x+y)/2$ gives $\|B\|_F^2/d^2=2\|(I-P_D)u\|_2^2$, where $P_D$ projects onto the line spanned by $x-y$. Equivalently, with $Q_u=\|u\|_2^2 I-uu^\top\succeq0$, one has $\|B\|_F^2/2=D^\top Q_uD$. Substitution in \eqref{eq:transpose-critical-bound} gives the sufficient condition for $x=y$ at a critical pair
\[
  2\gamma+\sigma_h/2>\lambda_{\max}(S-\sigma_f Q_u).
\]
This test does not divide by $d$. When $\sigma_f>0$ and $u\ne0$, its right-hand side is strictly smaller than $\lambda_{\max}(S)$ precisely when $u$ is outside the top eigenspace of $S$: equality in both Rayleigh bounds would require a top eigenvector in $\ker Q_u=\operatorname{span}(u)$.

The measured pairing can also help when $\sigma_f=0$. For the symmetric weighted squared loss, $G=W\odot(Z-A)$ with symmetric $A,W$ and entrywise nonnegative $W$. Then $b_f^{\rm tr}=\tfrac12\sum_{ij} w_{ij} B_{ij}^2$, which retains the discrepancy on observed entries even if other entries have zero weight. All these tests apply at the evaluated factors and the specified penalty. Using a measured correction to propose a subsequent penalty requires fresh evaluation at the new pair.

The midpoint gives a PSD matrix of rank at most $r$. Expanding its product gives~\cite{hu2026abmf}
\begin{equation}
UU^\top-\operatorname{sym}(XY^\top)=\tfrac14DD^\top.
\label{eq:midpoint-geometry}
\end{equation}
Thus its perturbation of the symmetric part is at most $\|D\|_F^2/4$ in Frobenius norm. This measures an output correction, not error relative to an unknown ground truth.

\subsection{Computing the spectral improvement}
\label{sec:consensus-geometry}
Theorem~\ref{thm:geometry} identifies when the transpose difference yields a smaller spectral threshold. The operator in \eqref{eq:allrank-threshold} acts on $n\times r$ factors. The following estimate uses an $n\times n$ matrix.

\begin{proposition}[A compressed spectral bound and an angle estimate]
\label{prop:compressed-spectral}
Use the hypotheses and notation of Theorem~\ref{thm:geometry}, and let $E$ be the full top eigenspace of $S$.
For full column rank $U$, a cheaper certified upper bound is
\begin{equation}
 \kappa_U\le\lambda_{\max}\!\big(S-\alpha(I-P_U)\big),
 \qquad \alpha=\sigma_f s_{\min}(U)^2,
 \label{eq:compressed-threshold}
\end{equation}
where $P_U$ is the projector onto $\operatorname{range}(U)$. For $\sigma_f>0$, this $n$-dimensional bound has the same strict-improvement criterion. If $E\ne\mathbb R^n$, write $g$ for the gap to the largest eigenvalue below the top one. Then
\begin{equation}
 \lambda_{\max}(S)-\kappa_U
 \ge \min(g,\alpha)\big(1-\|P_EP_U\|_2\big).
 \label{eq:angle-guarantee}
\end{equation}
\end{proposition}
Thus the possible gain in the penalty threshold is half the spectral drop. Equation~\eqref{eq:angle-guarantee} also explains why strict improvement alone can be numerically negligible. The eigenvalue tests use an exact eigenvalue or a certified upper bound. The directional implementation in Appendix~\ref{sec:adaptive} uses neither.

\paragraph{Masked observations.}
For symmetric weighted least squares, define an edge map with one row per observed unordered off-diagonal pair,
\[
 (\mathcal M_UH)_{ij}=\sqrt{w_{ij}}\,(h_i^\top u_j-u_i^\top h_j),
 \quad i<j,
\]
where $h_i^\top,u_i^\top$ are rows of $H,U$. The exact pairing satisfies $b_f^{\rm tr}=\|\mathcal M_UD\|_2^2$. Consequently Theorem~\ref{thm:geometry}'s residual bound holds with
\[
 \kappa_{W,U}=\lambda_{\max}(\mathcal S-\mathcal M_U^*\mathcal M_U).
\]
It improves strictly on $\lambda_{\max}(S)$ precisely when $\mathcal M_U$ is injective on the matrices with columns in $E$. Indeed, equal Rayleigh bounds require both membership in that subspace and zero observed discrepancy. In particular at least $r\dim E$ observed off-diagonal pairs are necessary for strict improvement. Gram-completion rigidity~\cite{singer2010rigidity} instead differentiates observed symmetric inner products, giving a sum of two terms on each edge. Here the edge differences measure disagreement of split factors.

\paragraph{Exact curvature along the disagreement direction.}
The global modulus can lose information even for a smooth loss. For a twice continuously differentiable convex $f$, put $B=Z-Z^\top$ and let
\[
 \mathcal H_Z=\int_0^1\nabla^2 f(Z^\top+sB)\,ds.
\]
The fundamental theorem of calculus gives $G-G^\top=\mathcal H_Z[B]$ and $b_f^{\rm tr}=\langle B,\mathcal H_Z[B]\rangle/2$. Thus the pointwise residual bound of Theorem~\ref{thm:geometry} holds with the exact secant operator
\[
 \kappa_Z=\lambda_{\max}\bigl(\mathcal S-
                       \tfrac12 L_U^*\mathcal H_Z L_U\bigr).
\]
It specializes to $\kappa_{W,U}$ for weighted least squares. Its strict improvement requires injectivity of $\mathcal H_Z^{1/2}L_U$ on the top factor eigenspace, by the same Rayleigh argument.

At a diagonal pair $Z=UU^\top$, there is a converse with a local optimization meaning. Let $h=0$, $G=\nabla f(UU^\top)$, and $\varphi_H(t)=F_\gamma(U+tH/2,U-tH/2)$. By transpose symmetry $\varphi_H$ is even. The product on this path equals $UU^\top+tL_UH/2-t^2HH^\top/4$. Two differentiations therefore give
\begin{equation}
\begin{aligned}
 2\varphi_H''(0)
 &=2\gamma\|H\|_F^2-\langle H,GH\rangle\\
 &\quad+\tfrac12\langle L_UH,\nabla^2 f(UU^\top)[L_UH]\rangle .
\end{aligned}
\label{eq:normal-hessian}
\end{equation}
In particular $\min_{\|H\|_F=1}2\varphi_H''(0)=2\gamma-\kappa_{UU^\top}$. Above this threshold all nonzero disagreement directions have positive curvature; below it a diagonal critical point is a strict saddle of the split objective. At equality the smallest second derivative is zero and higher-order terms decide stability along its null directions. For twice differentiable $h$, the right-hand side gains $\langle H,\nabla^2 h(U)[H]\rangle/2$. This interpretation concerns directions with the midpoint fixed; optimization within the diagonal still depends on the original symmetric objective.

\paragraph{An explicit separation on positive definite data.}
Take $r=1$, $h=0$, and observe the four edges $12,23,34,41$ of a cycle in both orientations with unit weights. Put $A_{ii}=3$, $A_{23}=A_{32}=A_{41}=A_{14}=2$, with all remaining entries zero. This is a permutation of two positive definite $2$-by-$2$ blocks. For $u=\mathbf1$, the symmetric gradient at $X=Y=u$ has successive edge values $+1,-1,+1,-1$ and annihilates $u$. Its top eigenvalue is $2$. The matrix $\mathcal M_u^*\mathcal M_u$ is the cycle Laplacian. Subtracting it leaves two blocks with diagonal $-2$ and off-diagonal $2$, so $\kappa_{W,u}=0$. The original pointwise spectral test requires $\gamma>1$, while the observed-entry test holds for every $\gamma>0$ at this pair.

The separation also occurs at unequal factors. Put $v=(1,1,-1,-1)^\top$, $X=u+tv/2$, and $Y=u-tv/2$. For $0<|t|<\sqrt{2}$ at $\gamma=1/2$, the two directional margins, computed without $\sigma_h$, are
\begin{equation}
 m_{\rm old}=-1+t^2/2,\qquad
 m_{\rm obs}=1+t^2/2.
 \label{eq:cycle-gates}
\end{equation}
Indeed $Sv=2(1-t^2/4)v$ and $\|\mathcal M_uv\|_2^2=2\|v\|_2^2$. Both residuals tend to zero as $t\to0$. Thus for any positive residual tolerance the two acceptance rules can disagree even at arbitrarily accurate pairs. This example concerns a neighborhood of one critical point; it gives no global all-positive-penalty result for weighted losses.

\paragraph{A boundary on full-observation Top-$r$.}
For $h=0$, $A\succeq0$ and $f(Z)=\|Z-A\|_F^2/2$, every diagonal critical pair has $S=UU^\top-A$ and $SU=0$. If $U\ne0$, $S$ vanishes on $\operatorname{range}(U)$ and is negative semidefinite on its orthogonal complement. Hence $\lambda_{\max}(S)=0$ and its top eigenspace contains $\operatorname{range}(U)$. By \eqref{eq:allrank-strict}, $\kappa_U=0$ as well. At $U=0$, $L_U=0$ directly. At these critical pairs the two spectral thresholds coincide, although the evaluated directional test can change along a finite trajectory. The positive-penalty Top-$r$ result in Appendix~\ref{sec:hzero-full-ls} instead follows from Gram balance.

\subsection{The residual sum and midpoint stationarity}
Midpoint cancellation of second-order disagreement terms is already present in the approximate KKT analysis of Han et al.~\cite{han2026lorads}. We use that mechanism with observable sum and difference residuals and a finite exactness margin, and subsequently extend the certificate to shared nonsmooth regularizers.

Write $\Phi(V)=f(VV^\top)+h(V)$ and
\[
K=\tfrac12(DU^\top-UD^\top),\qquad
Z=UU^\top-\tfrac14DD^\top+K.
\]
All operator norms on linear maps below are induced by the Frobenius norm.

\begin{theorem}[Quadratic midpoint identity]
\label{thm:midpoint-quadratic}
Let
\[
f(W)=\tfrac12\langle W,\mathcal Q W\rangle-\langle B,W\rangle+c,
\]
where $\mathcal Q$ is self-adjoint positive semidefinite,
$\mathcal Q(W^\top)=(\mathcal Q W)^\top$, $B=B^\top$, and
$\|\mathcal Q\|\le L_f$. Suppose $h$ is a convex quadratic function, including $h=0$, and use its gradients in \eqref{eq:residual-definition}. Then
\begin{equation}
\nabla\Phi(U)-e_+
 =\tfrac12\mathcal Q(DD^\top)U+(\mathcal Q K)D,
\label{eq:midpoint-quadratic-identity}
\end{equation}
and hence
\begin{equation}
\|\nabla\Phi(U)\|_F
 \le \|e_+\|_F+\tfrac32L_f\|U\|_2\|D\|_F^2.
\label{eq:midpoint-quadratic-bound}
\end{equation}
\end{theorem}
The quadratic order cannot generally be improved: in the scalar case
$f(z)=\tfrac12(z-a)^2$, $h=0$, and $U=u\ne0$, the difference in
\eqref{eq:midpoint-quadratic-identity} is exactly $uD^2/2$.

\begin{theorem}[Smooth midpoint transfer]
\label{thm:midpoint-smooth}
Suppose $f$ is transpose-symmetric and twice continuously differentiable, with Hessian norm bounded by $L_f$ and Hessian Lipschitz constant $M_f$ on a convex, transpose-invariant neighborhood containing $UU^\top$, $Z$, $Z^\top$, and the segments used between these points. Suppose $h$ has Hessian Lipschitz constant $M_h$ on a convex neighborhood of $X,Y$. If $\|U\|_2\le R$, then
\begin{equation}
\begin{aligned}
\|\nabla\Phi(U)-e_+\|_F&\le C_R\|D\|_F^2,\\
C_R&=\tfrac32L_fR+M_fR^3+\tfrac18M_h.
\end{aligned}
\label{eq:midpoint-smooth-constant}
\end{equation}
\end{theorem}
The smooth transfer uses differentiability and local Hessian bounds. The consensus estimate in Theorem~\ref{thm:consensus-residual} also uses convexity. In particular, a fixed positive margin and small $e_\pm$ suffice for a finite-penalty stationarity estimate. This is conditional on that margin, not a guarantee that any prescribed finite penalty has it.

\begin{corollary}[Smooth finite-accuracy stopping]
\label{cor:finite-stopping}
Under Theorem~\ref{thm:consensus-residual} and either smooth transfer theorem, let $C$ denote the applicable constant and let $m>0$ be an available margin from \eqref{eq:residual-margin} or \eqref{eq:directional-margin}. Then
\begin{equation}
\|\nabla\Phi(U)\|_F
 \le \|e_+\|_F+C\bigl(\|e_-\|_F/m\bigr)^2.
\label{eq:finite-stopping-smooth}
\end{equation}
For $C>0$, the sufficient tests
\[
\|e_+\|_F\le\varepsilon/2,\qquad
\|e_-\|_F\le m\sqrt{\varepsilon/(2C)}
\]
certify $\|\nabla\Phi(U)\|_F\le\varepsilon$.
\end{corollary}
\begin{proof}
In the midpoint bound of Theorem~\ref{thm:midpoint-quadratic} or~\ref{thm:midpoint-smooth}, substitute $\|D\|_F\le\|e_-\|_F/m$ from Theorem~\ref{thm:consensus-residual}. The two tests bound the terms in \eqref{eq:finite-stopping-smooth} by $\varepsilon/2$ each. For $C=0$, $\|e_+\|_F\le\varepsilon$ alone suffices.
\end{proof}
The different tolerances are justified by the two residuals' different roles; they do not prove that every solver can attain those tolerances independently or faster. Actual savings must include the cost of evaluating the certificate.

\subsection{Shared nonsmooth regularizers and constraints}
For a convex $h$, averaging subgradients at different points need not yield a subgradient at their midpoint. The following enlarged-subgradient argument avoids this invalid step. Define
\begin{equation}
\begin{split}
\bar H&=\tfrac12(H_X+H_Y),\\
\epsilon_h&=h(U)-\tfrac12\{h(X)+h(Y)\}
              +\tfrac14\langle H_X-H_Y,D\rangle .
\end{split}                                                \label{eq:epsilon-h}
\end{equation}
All three function values are finite when $X,Y\in\operatorname{dom}\partial h$. Write $g(V)=f(VV^\top)$ and, for any $t>0$, define
\begin{equation}
\mathcal G_t(U)
 =\frac{U-\operatorname{prox}_{th}(U-t\nabla g(U))}{t}.
\label{eq:prox-gradient-map}
\end{equation}
This is a stationarity measure for the original factorized problem, not a certificate of its global optimum.

\begin{theorem}[Proximal certificate with a shared convex regularizer]
\label{thm:midpoint-proximal}
Let $h$ be proper, closed, convex and let $f$ be differentiable and transpose-symmetric. Then $\epsilon_h\ge0$ and
$\bar H\in\partial_{\epsilon_h}h(U)$, where
\[
\begin{aligned}
&v\in\partial_\epsilon h(U)\\
&\quad\Longleftrightarrow\quad
h(V)\ge h(U)+\langle v,V-U\rangle-\epsilon\\
&\hspace{6em}\text{for every }V.
\end{aligned}
\]
Suppose a valid bound $b_f$ satisfies
\begin{equation}
\|\nabla g(U)+\bar H-e_+\|_F\le b_f.          \label{eq:loss-transfer-budget}
\end{equation}
With $a=\|e_+\|_F+b_f$, one has
\begin{equation}
\|\mathcal G_t(U)\|_F
 \le \frac{a+\sqrt{a^2+4\epsilon_h/t}}{2}
 \le a+\sqrt{\epsilon_h/t}.
\label{eq:proximal-certificate}
\end{equation}
For a quadratic loss, take $b_f=\tfrac32L_f\|U\|_2\|D\|_F^2$.
For a $C^{2,1}$ loss satisfying Theorem~\ref{thm:midpoint-smooth}, take
$b_f=(\tfrac32L_fR+M_fR^3)\|D\|_F^2$.
If, in addition, $f$ is convex and
$m_0=2\gamma-\lambda_{\max}(\operatorname{sym}G)>0$, then
\begin{equation}
0\le\epsilon_h\le\frac{\|e_-\|_F^2}{8m_0}.
\label{eq:epsilon-h-residual-bound}
\end{equation}
For $D\ne0$, this last assertion also holds with
$m_0=2\gamma-\langle D,SD\rangle/\|D\|_F^2>0$, or with the larger positive value $m_0=2\gamma-q_D+b_f^{\rm tr}/\|D\|_F^2$.
\end{theorem}
Notice that the last bound deliberately omits the $\sigma_h/2$ enhancement: it controls an enlargement of the original $h$, not just consensus. It follows, for either second-order loss budget with constant $C_f$, that
\begin{equation}
\|\mathcal G_t(U)\|_F
 \le \|e_+\|_F+
       C_f(\|e_-\|_F/m_0)^2+
       \frac{\|e_-\|_F}{\sqrt{8tm_0}} .
\label{eq:finite-stopping-nonsmooth}
\end{equation}
The ordinary nonsmooth stationarity residual cannot replace this proximal measure without extra regularity. For example, with scalar $f=0$, $h(x)=|x|$, $X=3s$, $Y=-s$, and $s>0$, valid endpoint subgradients give $e_+=0$ and $D=4s$, whereas $\operatorname{dist}(0,\partial h(U))=1$ at $U=s$. This example rules out a universal bound using only $\|e_+\|+O(\|D\|^2)$.

If $f$ has only an $L_f$-Lipschitz gradient on the required neighborhood, a valid, weaker budget is
\begin{equation}
b_f=2L_fR^2\|D\|_F+\tfrac32L_fR\|D\|_F^2 .
\label{eq:lip-gradient-budget}
\end{equation}
Thus the proximal certificate remains available for such losses, but the quadratic transfer must not be asserted across a Huber breakpoint. Smooth robust losses or a fixed local quadratic region meet stronger assumptions. Proofs of all statements are provided in Appendix~\ref{app:finite-proofs}.

\section{Certified Adaptive Penalties with Finite Inner Accuracy}
\label{sec:adaptive}
The practical spectral rule of Hu et al.~\cite{hu2026abmf} changes the objective being minimized. Its convergence does not follow from a fixed-penalty descent argument. We introduce a stagewise algorithm whose acceptance test concerns the original symmetric problem. It can use an arbitrary spectral estimate to propose a penalty, while a separate computable test certifies the returned point. The guarantee is finite termination at a requested accuracy and stationarity of accumulation points as the requested accuracy decreases; it does not assert convergence of every intermediate iterate.

\subsection{An obstruction to cross-parameter descent}
\begin{proposition}
\label{prop:penalty-obstruction}
For $f(z)=\frac12(z+1)^2$, $h=0$, and $0<\gamma<1/2$,
\[
 \min_{x,y}F_\gamma(x,y)=2\gamma-2\gamma^2,
\]
attained at $x=-y=\sqrt{1-2\gamma}$ and its simultaneous sign reversal. Increasing $\gamma$ within $(0,1/2)$ from such a minimizer makes
$F_{\gamma_{\rm new}}(x_{\rm new},y_{\rm new})
 \le F_{\gamma_{\rm old}}(x_{\rm old},y_{\rm old})$
impossible, regardless of the new factors.
\end{proposition}
Thus increasing the penalty until the changing objective decreases need not terminate. We instead control each fixed-penalty inner solve by a common reference level and accept a stage using an original-problem certificate.

\subsection{Assumptions and safeguards}
Fix $\gamma_{\min}>0$ and a feasible reference factor $V_0$, and set
\[
 E_0=\Phi(V_0)=F_\gamma(V_0,V_0).
\]
This identity holds for every $\gamma$.
\begin{assumption}
\label{ass:adaptive}
The loss $f$ is finite-valued, convex, differentiable, transpose symmetric, and has an $L$-Lipschitz gradient in the Frobenius norm. The function $h$ is proper, closed, convex, with a computable proximal map. A known $R>0$ bounds both factor Frobenius norms throughout
\[
 \mathcal S=\{(X,Y):F_{\gamma_{\min}}(X,Y)\le E_0\}.
\]
A known finite lower bound $F_*$ satisfies $F_{\gamma_{\min}}\ge F_*$. No new constraint or regularizer is added by this assumption.
\end{assumption}
These conditions do not require a strongly convex or coercive loss. For example, if the original model already has
\[
 h(V)=\frac{\mu}{2}\|V\|_F^2+\widetilde h(V),
 \quad \mu>0,\quad \widetilde h\ge h_0,\quad f\ge f_0,
\]
one may take
\[
 R^2=\frac{4(E_0-f_0-h_0)}{\mu},\qquad F_*=f_0+h_0.
\]
An existing compact convex factor constraint gives another direct bound. Adding a constraint solely to meet this assumption would change the model.

Choose a margin $\underline m>0$, a retry factor $\beta>1$, and an original-problem proximal-gradient parameter $t>0$. Put
\begin{align}
 B&=\|\nabla f(0)\|_2+LR^2,\nonumber\\
 \Gamma&=\max\{\gamma_{\min},(B+\underline m)/2\},
 \label{eq:gamma-certified-cap}\\
 \overline L_b&=LR^2+\Gamma,\qquad
 \alpha_\gamma=\frac{1}{2(LR^2+\gamma)}.\nonumber
\end{align}
For every pair in $\mathcal S$, $\|\nabla f(XY^\top)\|_2\le B$. Hence $\Gamma$ guarantees the required margin. It is a termination safeguard, not a penalty that must be used at every stage.

For $D=X-Y\ne0$ and the current $G=\nabla f(XY^\top)$, use the directional margin
\begin{equation}
\begin{aligned}
 m_{\rm dir}&=2\gamma-q_D+b_f^{\rm tr}/\|D\|_F^2\\
 &=2\gamma+\langle D,GY-G^\top X\rangle/\|D\|_F^2 .
\end{aligned}
\label{eq:directional-margin-adaptive}
\end{equation}
Equation~\eqref{eq:proof-loss-monotonicity} gives the second form. Both gradient products are already required for the current residuals. Implementation can compare $\langle D,GY-G^\top X\rangle$ with $(\underline m-2\gamma)\|D\|_F^2$, using $O(nr)$ extra arithmetic without an eigensolve or a dense product $XY^\top$. At $D=0$ the test passes directly. Since $b_f^{\rm tr}\ge0$, $2\gamma-q_D\ge\underline m$ implies $m_{\rm dir}\ge\underline m$ at the same endpoint. For $h=0$, the entire test reduces to $\langle D,e_-\rangle\ge\underline m\|D\|_F^2$; measuring the disagreement and midpoint gradient directly is also feasible.

\subsection{Inner updates and acceptance}
Write
$\psi_\gamma(X,Y)=f(XY^\top)+\frac{\gamma}{2}\|X-Y\|_F^2$.
At a fixed penalty trial, write $\alpha=\alpha_\gamma$, so a smaller trial penalty permits a larger inner step while the proof retains uniform safeguards. One inner iteration consists of
\begin{align}
 X^+&=\operatorname{prox}_{\alpha h/2}
       \bigl(X-\alpha\nabla_X\psi_\gamma(X,Y)\bigr),\nonumber\\
 Y^+&=\operatorname{prox}_{\alpha h/2}
       \bigl(Y-\alpha\nabla_Y\psi_\gamma(X^+,Y)\bigr).
 \label{eq:certified-inner}
\end{align}
These are finite block updates, not exact minimizers of the aBMF block objectives. The proximal maps themselves are assumed exact; numerical errors in them require additional residual control. Their optimality equations provide
\begin{align}
 H_X^+&=-2\nabla_X\psi_\gamma(X,Y)-2(X^+-X)/\alpha,\nonumber\\
 H_Y^+&=-2\nabla_Y\psi_\gamma(X^+,Y)-2(Y^+-Y)/\alpha.
 \label{eq:prox-subgradient-selections}
\end{align}
These belong to $\partial h(X^+)$ and $\partial h(Y^+)$. Evaluate $e_X,e_Y$ of \eqref{eq:residual-definition} at the same final pair using these selections, not gradients from different half-steps.

For target $\varepsilon>0$, define
\begin{equation}
 C_{\rm stop}
 =1+\frac{5LR^2}{\underline m}
       +\frac{1}{\sqrt{8\underline m t}},
 \qquad \tau=\varepsilon/C_{\rm stop}.
 \label{eq:stage-tolerance}
\end{equation}
The conservative inner stopping test is
$\max\{\|e_+\|_F,\|e_-\|_F\}\le\tau$.
Then accept if $D=0$ or $m_{\rm dir}\ge\underline m$, and return $U=(X+Y)/2$. Otherwise increase $\gamma$ to $\min\{\beta\gamma,\Gamma\}$ and repeat the fixed-penalty solve. The sharper posterior certificates of Appendix~\ref{sec:finite-accuracy} may permit earlier acceptance; the conservative test gives a uniform worst-case bound.

At every trial, use a proposed warm start only if its value of the new $F_\gamma$ is at most $E_0$; otherwise reset to $(V_0,V_0)$. This avoids an unproved boundedness assumption under repeated penalty increases.

\begin{algorithm}[t]
\caption{Stagewise certified adaptive aBMF}
\label{alg:certified-adaptive}
\begin{enumerate}
\item Given $\varepsilon$, propose any $\gamma\in[\gamma_{\min},\Gamma]$, optionally using warm-started power iterations.
\item Check the warm start against $F_\gamma\le E_0$; reset to $(V_0,V_0)$ if necessary.
\item Apply \eqref{eq:certified-inner} until
$\max\{\|e_+\|_F,\|e_-\|_F\}\le\tau$.
\item If $D=0$ or $m_{\rm dir}\ge\underline m$, return $U=(X+Y)/2$ and the measured residuals.
\item Otherwise set $\gamma\leftarrow\min\{\beta\gamma,\Gamma\}$ and return to Step~2.
\end{enumerate}
\end{algorithm}

\subsection{Termination and original-problem accuracy}
\begin{theorem}[Certified output and total inner work]
\label{thm:adaptive-certified}
Under Assumption~\ref{ass:adaptive}, Algorithm~\ref{alg:certified-adaptive} terminates and returns a factor $U$ satisfying
\[
 \|\mathcal G_t(U)\|_F\le\varepsilon,\qquad
 \|X-Y\|_F\le\tau/\underline m,
\]
where $\mathcal G_t$ is defined in \eqref{eq:prox-gradient-map}. The output $UU^\top$ is PSD of rank at most $r$. A convex factor constraint encoded by $h$ is preserved at $U$.

Let
\[
 J=1+\left\lceil\frac{\log(\Gamma/\gamma_{\min})}{\log\beta}\right\rceil,
 \qquad a=(LR^2+\gamma_{\min})/2,
\]
and
\[
 C_r=2(2LR^2+B+2\Gamma+2\overline L_b).
\]
The total number of inner iterations, including rejected penalty trials, is at most
\begin{equation}
 J\left[1+\left\lceil
      \frac{(E_0-F_*)C_r^2}{a\tau^2}
     \right\rceil\right].
 \label{eq:adaptive-complexity}
\end{equation}
For fixed problem constants this is $O(J\varepsilon^{-2})$ inner iterations, not a wall-clock bound; actual gradient and proximal-map costs must also be counted.
\end{theorem}

\begin{corollary}[Decreasing requested accuracies]
\label{cor:adaptive-cluster}
Run Algorithm~\ref{alg:certified-adaptive} with $\varepsilon_s\downarrow0$, common reference level and constants, and arbitrary penalty proposals at each stage. Every accumulation point of the returned factors is critical for $\Phi$, and the accepted factor differences tend to zero. Monotonicity or convergence of the accepted penalties is not required.
For $\varepsilon_s=\varepsilon_0q^s$, $0<q<1$, cumulative inner work through the first stage with accuracy at most $\varepsilon$ is $O(J\varepsilon^{-2})$.
\end{corollary}
The corollary concerns the factors returned at accepted stages. Penalties may increase or decrease between stages. The per-iteration update of Hu et al.~\cite{hu2026abmf} has no such acceptance test. A short power iteration can propose $\gamma$, while a certified finite cap such as \eqref{eq:gamma-certified-cap} supplies the termination guarantee.

\section{The Unregularized Case}
\label{sec:unregularized}
Here $h\equiv0$ means that neither a factor regularizer nor a factor constraint is present. In particular, nonnegativity encoded by an indicator is not an instance of $h=0$. This specialization removes the enlarged-subgradient error and exposes additional algebraic structure. We distinguish conclusions valid for general losses from the stronger ones specific to full-observation squared loss.

\subsection{Residuals without a shared regularizer}
\label{sec:hzero-general}
With the same $U,D,G$ as above and a differentiable loss,
\begin{equation}
\begin{aligned}
 e_X&=GY+\gamma D,& e_Y&=G^\top X-\gamma D,\\
 e_+&=GY+G^\top X,&
 \nabla\Phi(U)&=2\nabla f(UU^\top)U.
\end{aligned}
\label{eq:hzero-residuals}
\end{equation}
In particular, $H_X=H_Y=0$, $\epsilon_h=0$, and
$\mathcal G_t(U)=\nabla\Phi(U)$ for every $t>0$. There is no nonsmooth term or proximal-map parameter in the output certificate.

\begin{corollary}[Unregularized finite-accuracy certificate]
\label{cor:hzero-certificate}
Let $f$ be convex, differentiable, and transpose symmetric. Suppose that, at the current pair, either
\[
 m=2\gamma-\lambda_{\max}(\operatorname{sym}G)>0
\]
or, for $D\ne0$,
\[
 m=2\gamma-\frac{\langle D,GD\rangle}{\|D\|_F^2}>0.
\]
Then $\|D\|_F\le\|e_-\|_F/m$. Under the quadratic or smooth hypotheses of Theorems~\ref{thm:midpoint-quadratic}--\ref{thm:midpoint-smooth}, with $h=0$ and $\|U\|_2\le R$,
\begin{equation}
\begin{aligned}
 \|\nabla\Phi(U)\|_F
 &\le\|e_+\|_F+C_0\|D\|_F^2\\
 &\le\|e_+\|_F+C_0(\|e_-\|_F/m)^2,
\end{aligned}
\label{eq:hzero-certificate}
\end{equation}
where $C_0=\frac32L_fR$ for a quadratic loss, and
$C_0=\frac32L_fR+M_fR^3$ under the stated Hessian-Lipschitz assumptions. At $D=0$, $\nabla\Phi(U)=e_+$ without a margin test.
\end{corollary}
The bound using the measured $D$ may be tighter than replacing it by its residual upper bound. With only a Lipschitz loss gradient, the weaker budget \eqref{eq:lip-gradient-budget} remains necessary; setting $h=0$ does not itself justify a quadratic transfer rate.

\subsection{Simpler inner updates and stopping tolerances}
The proximal maps in \eqref{eq:certified-inner} become the identity, so the inner method is alternating gradient descent on the fixed trial objective. The same-point residuals \eqref{eq:hzero-residuals} require no subgradient recovery. Under Assumption~\ref{ass:adaptive}, the general Lipschitz-gradient variant can replace \eqref{eq:stage-tolerance} by
\begin{equation}
 C_{\mathrm{stop},0}=1+\frac{5LR^2}{\underline m},\qquad
 \tau=\varepsilon/C_{\mathrm{stop},0}.
\label{eq:hzero-lipschitz-stop}
\end{equation}
All other energy, margin, and retry safeguards are retained.

When a uniform quadratic transfer constant $C_0>0$ is available, a less restrictive consensus tolerance is justified:
\begin{equation}
 \|e_+\|_F\le\varepsilon/2,\qquad
 \|e_-\|_F\le\underline m\sqrt{\frac{\varepsilon}{2C_0}}.
\label{eq:hzero-asymmetric-stop}
\end{equation}
After the same margin acceptance test, these inequalities certify
$\|\nabla\Phi(U)\|_F\le\varepsilon$. The inner loop may check these two thresholds separately. For its worst-case analysis, set
\[
 \tau_0=\min\left\{\frac{\varepsilon}{2},
          \underline m\sqrt{\frac{\varepsilon}{2C_0}}\right\}.
\]
Theorem~\ref{thm:adaptive-certified}'s iteration bound holds with $\tau_0$ replacing $\tau$, because a joint residual at most $\tau_0$ suffices for both tests. Thus the total work is
\[
 O\!\left(J\max\left\{\varepsilon^{-2},
                \frac{C_0}{\underline m^2\varepsilon}\right\}\right),
\]
with the remaining problem constants fixed. It remains $O(J\varepsilon^{-2})$ as $\varepsilon\downarrow0$: the square-root consensus tolerance does not improve the overall worst-case exponent. If $C_0=0$, requiring both residuals to be at most $\varepsilon$ suffices and also forces consensus in the limit. Direct evaluation of $\nabla\Phi(U)$ remains a competing stopping strategy.

\subsection{Bounded sublevels without adding a regularizer}
\begin{proposition}[Loss-controlled factor bounds]
\label{prop:hzero-sublevel}
Suppose $h=0$, $f\ge f_*>-\infty$, and a known $M_E<\infty$ satisfies
\[
 f(Z)\le E_0\quad\Longrightarrow\quad\|Z\|_F\le M_E.
\]
For every $\gamma\ge\gamma_{\min}>0$ and pair with $F_\gamma(X,Y)\le E_0$,
\begin{equation}
 \|X\|_F^2+\|Y\|_F^2
 \le 2\sqrt r\,M_E+\frac{2(E_0-f_*)}{\gamma_{\min}}.
\label{eq:hzero-factor-bound}
\end{equation}
The right side therefore provides an admissible $R^2$ in Assumption~\ref{ass:adaptive}, with $F_*=f_*$.
\end{proposition}
A coercive finite continuous loss ensures that such a sublevel radius exists; an executable penalty cap additionally requires a known bound on it. For the unweighted loss $f(Z)=\frac12\|Z-A\|_F^2$, one may take $f_*=0$ and $M_E=\|A\|_F+\sqrt{2E_0}$. No ridge penalty is inserted into the model. By contrast, $f(Z)=\frac12 Z_{11}^2$ on $2\times2$ matrices has the unbounded zero-energy family $X=Y=(0,t)^\top$. A positive consensus penalty alone cannot bound unobserved equal-factor directions.

\subsection{PSD data alone do not settle the general-loss case}
\begin{proposition}[A weighted PSD-data counterexample]
\label{prop:hzero-weighted-counterexample}
Let
\[
 A=\begin{pmatrix}1&1\\1&1\end{pmatrix}\succeq0,\qquad
 W=\begin{pmatrix}1/3&1\\1&1/3\end{pmatrix},
\]
and $f(Z)=\frac12\sum_{i,j}w_{ij}(Z_{ij}-A_{ij})^2$, with $h=0$.
For $\gamma=1/6$, the factors
$X=(1/2,-1/2)^\top$ and $Y=-X$ are an unequal critical pair of $F_\gamma$.
\end{proposition}
All weights are positive; the loss is coercive, smooth, strongly convex, and transpose symmetric. Therefore the every-positive-penalty conclusion for unweighted PSD approximation, established below, must not be inferred from $h=0$ and PSD data alone or transferred unchanged to weighted observations.

\subsection{Gram balance without a shared regularizer}
\label{sec:hzero-gram}
For the residuals in \eqref{eq:residual-definition}, the following identity uses only the common loss-subgradient selection; it does not require a quadratic loss.
\begin{lemma}[Residual Gram balance]
\label{lem:hzero-balance}
Let $h=0$, put $C=X^\top X-Y^\top Y$, and define $e_X,e_Y$ using the same $G$. Then
\begin{equation}
\gamma C=X^\top e_X-e_Y^\top Y.
\label{eq:hzero-balance}
\end{equation}
For $\gamma\ne0$,
\begin{equation}
\|C\|_F\le
\frac{\|X\|_2\|e_X\|_F+\|Y\|_2\|e_Y\|_F}{|\gamma|}.
\label{eq:hzero-balance-bound}
\end{equation}
Consequently, every joint critical point with $\gamma\ne0$ has equal factor Gram matrices.
\end{lemma}
Gram balance at stationary points of regularized bilinear factorizations is an established mechanism~\cite{li2019nonconvexgeometry}. We use its residual form to obtain a specialization that avoids estimating the largest eigenvalue of the current loss gradient. Equal Gram matrices alone do not imply $X=Y$.

\subsection{Full-observation least squares: a sharp exactness boundary}
\label{sec:hzero-full-ls}
Consider
\begin{equation}
F_\gamma(X,Y)=\tfrac12\|XY^\top-A\|_F^2
                  +\tfrac\gamma2\|X-Y\|_F^2,
\quad A=A^\top,
\label{eq:hzero-ls}
\end{equation}
with $h=0$. Define $U=(X+Y)/2$, $D=X-Y$, and
\[
T=\tfrac12(X^\top X+Y^\top Y),\qquad C=X^\top X-Y^\top Y.
\]
The two residuals satisfy the exact identity
\begin{equation}
e_-=(A+2\gamma I)D+DT-UC.
\label{eq:hzero-sylvester}
\end{equation}
Unlike a sufficient bound involving $\lambda_{\max}(\operatorname{sym}G)$, the following result uses the input matrix and factor Gram matrices.

\begin{theorem}[Finite residuals and sharp critical-point exactness]
\label{thm:hzero-sharp}
Let $r\ge1$, $\gamma>0$, and
\[
\delta=\lambda_{\min}(A)+2\gamma,\qquad
\eta=\|e_-\|_F+\|UC\|_F.
\]
If $\delta\ge0$, then, with $d=\|D\|_F$,
\begin{equation}
\delta d+\frac{d^3}{4r}\le\eta.
\label{eq:hzero-cubic-certificate}
\end{equation}
In particular, $d\le\eta/\delta$ when $\delta>0$, and
$d\le(4r\eta)^{1/3}$ when $\delta=0$. A fully residual-based upper bound is
\begin{equation}
\begin{split}
\eta\le\|e_-\|_F+\frac{\|U\|_2}{\gamma}
\bigl(\|X\|_2\|e_X\|_F+\|Y\|_2\|e_Y\|_F\bigr).
\end{split}
\label{eq:hzero-eta-residual}
\end{equation}
Moreover, every critical point of \eqref{eq:hzero-ls} has $X=Y$ if and only if
\begin{equation}
A+2\gamma I\succeq0.
\label{eq:hzero-exact-threshold}
\end{equation}
Thus every $\gamma>0$ suffices for a PSD input $A$, including rank-deficient inputs.
\end{theorem}
The nonstrict boundary in \eqref{eq:hzero-exact-threshold} is included: the quartic factor term enforces consensus even when the linear spectral margin vanishes. At that boundary the displayed finite-residual bound is H\"older rather than Lipschitz. This order is attained in the scalar family $A=-2\gamma$, $X=t$, $Y=-t$: one has $C=0$, $d=2|t|$, and $\eta=2|t|^3$. This statement concerns full-observation Frobenius least squares; it does not follow merely from setting $h=0$ in an arbitrary transpose-symmetric loss.

Positive-penalty sublevels of this least-squares model are bounded, as shown below. Consequently minima exist, and under \eqref{eq:hzero-exact-threshold} all global minimizers have equal factors; in particular,
\[
\min_{X,Y}F_\gamma(X,Y)=\min_U\tfrac12\|UU^\top-A\|_F^2.
\]
This model-value equivalence does not make every critical point globally optimal.

For $A\succeq0$, no input eigenvalue computation is needed in
\eqref{eq:hzero-cubic-certificate}: the safe value $\delta=2\gamma$ gives
\begin{equation}
\|D\|_F\le\frac{\|e_-\|_F+\|UC\|_F}{2\gamma}.
\label{eq:hzero-psd-certificate}
\end{equation}
Together with Theorem~\ref{thm:midpoint-quadratic}, this yields
\begin{equation}
\begin{split}
\|\nabla\Phi(U)\|_F
\le\|e_+\|_F+\tfrac32\|U\|_2
       \left(\frac{\eta}{2\gamma}\right)^2.
\end{split}
\label{eq:hzero-psd-output}
\end{equation}
The directly measured $d$ or the positive root of
$2\gamma d+d^3/(4r)=\eta$ can provide a tighter posterior bound. These are stationarity certificates, not guarantees that a stationary factor represents the leading eigenspace or a global optimum.

Still assuming $A\succeq0$, for a finite-accuracy specialization suppose $\|X\|_F,\|Y\|_F\le R$ with $R>0$, and put
\[
a_\gamma=1+2R^2/\gamma.
\]
If $\max\{\|e_+\|_F,\|e_-\|_F\}\le\tau$, then
$\|e_X\|_F,\|e_Y\|_F\le\tau$ and
\[
\|C\|_F\le2R\tau/\gamma,\qquad
\eta\le a_\gamma\tau.
\]
Consequently, the fixed-penalty test
\begin{equation}
\tau\le\min\left\{\frac\varepsilon2,
\frac{2\gamma}{a_\gamma}\sqrt{\frac{\varepsilon}{3R}}\right\}
\label{eq:hzero-fixed-tolerance}
\end{equation}
certifies $\|\nabla\Phi(U)\|_F\le\varepsilon$. For this particular PSD loss, spectral tracking is therefore unnecessary for critical-point exactness or this finite-output guarantee. The penalty can instead be selected to balance conditioning and computational cost; changing it during optimization still requires the appropriate safeguards. The direct $D$-based bound or a direct midpoint-gradient test can be less conservative.

The Gram term cannot simply be omitted. In the scalar PSD instance
$A=0$, $\gamma=1$, $X=2$, $Y=1$, one has $e_X=e_Y=3$ and $D=1$.
Thus $e_-=0$ despite unequal factors. Exactness for every positive penalty is not a uniform bound of the form $\|D\|\le\|e_-\|/(2\gamma)$ at arbitrary iterates. Nor is it uniform as $\gamma\downarrow0$: the residual balance estimate itself contains $1/\gamma$.

\subsection{Why zero and negative penalties are different}
\label{sec:hzero-nonpositive}
At $\gamma=0$, the transformation $(X,Y)\mapsto(cX,c^{-1}Y)$ preserves the product, and unequal critical factors can occur even for PSD data. For example, with scalar $A=1$, $(X,Y)=(c,c^{-1})$ is critical for every $c\ne0$. The positive-penalty conclusion must therefore not be extended to zero by continuity.

The distinction also matters for a vanishing positive sequence. Keep $A=1$, $X=2$, and $Y=1/2$, and let $\gamma_k\downarrow0$. Then
\[
e_X=\tfrac32\gamma_k,\qquad e_Y=-\tfrac32\gamma_k,
\]
yet $D=3/2$ and $\nabla\Phi((X+Y)/2)=45/32$ remain nonzero. This is a sequence of approximate critical points of changing objectives, not a claimed AM trajectory. It disproves a uniform finite-accuracy inference from $\gamma_k>0$ and $e_X,e_Y\to0$ alone. For this least-squares model, a uniform finite-residual guarantee can use a positive penalty floor together with a bound on the factor norms. Disagreement and the midpoint gradient may also be tested separately.

More decisively, for any $h=0$ loss finite at zero and any $\gamma<0$,
\begin{equation}
F_\gamma(tV,0)=f(0)+\tfrac\gamma2t^2\|V\|_F^2
\longrightarrow-\infty
\label{eq:hzero-negative-unbounded}
\end{equation}
for nonzero $V$. Hence a negative penalty is not a well-posed unconstrained aBMF model, even when the input is PSD. A particular quadratic block may remain locally convex if its Gram matrix dominates $-\gamma I$; that local fact does not remedy global unboundedness. Negative damping in a separately safeguarded local model would be a different algorithm and requires its own analysis.

\subsection{Fixed-penalty AM and subspace progress for PSD data}
\label{sec:hzero-am}
For \eqref{eq:hzero-ls} with $A\succeq0$ and fixed $\gamma>0$, exact alternating minimization uses
\begin{align}
X^{k+1}&=(A+\gamma I)Y^k\bigl((Y^k)^\top Y^k+\gamma I\bigr)^{-1},
\nonumber\\
Y^{k+1}&=(A+\gamma I)X^{k+1}
       \bigl((X^{k+1})^\top X^{k+1}+\gamma I\bigr)^{-1}.
\label{eq:hzero-am-updates}
\end{align}
The inverses exist for every starting pair. The sublevel sets are bounded without adding a ridge regularizer: if $F_\gamma\le E$, set
\[
d_E=\sqrt{2E/\gamma},\qquad M_E=\|A\|_F+\sqrt{2E}.
\]
Then both factor Frobenius norms are bounded by
\begin{equation}
R_E=\sqrt{\sqrt r\,(M_E+d_E^2/4)}+d_E/2.
\label{eq:hzero-sublevel-radius}
\end{equation}
Thus the general adaptive framework has a concrete boundedness witness for this loss when a positive $\gamma_{\min}$ is maintained.

\begin{proposition}[Fixed-penalty AM]
\label{prop:hzero-am}
For the PSD full-observation problem, exact AM with any fixed $\gamma>0$ generates a bounded sequence. The completed sweeps have finite length and converge to $(U_*,U_*)$, where $U_*$ is stationary for the symmetric objective. Both endpoint residuals tend to zero.
\end{proposition}
The limit can depend on the initialization. For instance, zero initial factors stay zero even if $A$ has positive eigenvalues.

The exact updates also implement shifted subspace iteration~\cite{hu2026abmf}. More explicitly, let $1\le r<n$, order the PSD eigenvalues as $\lambda_1\ge\cdots\ge\lambda_n\ge0$, assume $\lambda_r>\lambda_{r+1}$, and suppose the projection of $Y^0$ onto the leading $r$ eigenvectors is nonsingular. With $V_1,V_2$ the leading and remaining orthonormal eigenvectors, respectively, define $\mathcal T_k=V_2^\top Y^k(V_1^\top Y^k)^{-1}$. Then
\begin{equation}
\begin{split}
\operatorname{range}(Y^k)&=\operatorname{range}\bigl((A+\gamma I)^{2k}Y^0\bigr),\\
\|\mathcal T_k\|_2
&\le\left(\frac{\lambda_{r+1}+\gamma}{\lambda_r+\gamma}\right)^{2k}
\|\mathcal T_0\|_2.
\end{split}
\label{eq:hzero-subspace-rate}
\end{equation}
This quantifies column-space progress per complete sweep. Smaller positive $\gamma$ improves this particular ratio, but can worsen Gram-system conditioning, residual certification, and numerical stability. The ratio alone neither compares complete runtimes with specialized sBMF-GN nor controls the scaling of the factors. Proofs are given in Appendix~\ref{app:hzero}.

\section{How the Extension Determines Computational Curvature}
\label{sec:curvature}

A symmetric objective specifies the loss only on symmetric matrices.
Its extension away from this subspace is an additional modeling choice:
two extensions can agree on every \(XX^\top\) while inducing different
alternating subproblems. We characterize this choice without identifying
alternating minimization with a Gauss--Newton (GN) iteration.

\subsection{An exact model decomposition}
Fix \(X\), write \(Y=X+2P\), and define
\begin{equation}
\begin{split}
S&=XX^\top,\qquad J_X[P]=XP^\top+PX^\top,\\
K_X[P]&=XP^\top-PX^\top,\qquad M=S+J_X[P].
\end{split}
\label{eq:curv-operators}
\end{equation}
Here \(P=(Y-X)/2\) is a candidate midpoint step, not the disagreement
\(D=X-Y\) used in the finite-accuracy analysis. Let
\[
\mathcal D_f(Z,M)=f(Z)-f(M)-\langle\nabla f(M),Z-M\rangle
\]
denote the Bregman divergence, and define
\[
\mathcal J_h(X,P)=\tfrac12h(X)+\tfrac12h(X+2P)-h(X+P).
\]

\begin{proposition}[Implicit Bregman and Jensen terms]
\label{prop:curv-bregman}
Suppose \(f\) is differentiable, convex, and transpose symmetric,
and \(h\) is convex. For \(X,X+2P\in\operatorname{dom}h\),
\begin{equation}
\begin{split}
F_\gamma(X,X+2P)
={}& f(S+J_X[P])+h(X+P)+2\gamma\|P\|_F^2\\
&+\mathcal D_f(M+K_X[P],M)+\mathcal J_h(X,P).
\end{split}
\label{eq:curv-bregman}
\end{equation}
Both additional terms are nonnegative. If \(f\) is
\(\sigma_f\)-strongly convex and \(L_f\)-smooth on the relevant segment,
the Bregman term lies between
\(\frac{\sigma_f}{2}\|K_X[P]\|_F^2\) and
\(\frac{L_f}{2}\|K_X[P]\|_F^2\).
\end{proposition}

Thus, a block subproblem is exactly a prox-linear model of the symmetric
objective plus extension-dependent curvature and a regularizer Jensen
gap. Prox-linear methods themselves are established tools for low-rank
recovery~\cite{charisopoulos2021composite}; the identity specifies the
additional terms induced by this particular splitting. If \(h\) is an
indicator, its endpoint feasibility condition \(X+2P\in\operatorname{dom}h\)
must be retained. Midpoint feasibility alone does not give an equivalent
subproblem.

For \(f(Z)=\frac12\|Z-A\|_F^2\), \(A=A^\top\), and \(h=0\),
\eqref{eq:curv-bregman} reduces to
\begin{equation}
F_\gamma(X,X+2P)=
\tfrac12\|R+J_X[P]\|_F^2+
\tfrac12\|K_X[P]\|_F^2+2\gamma\|P\|_F^2,
\label{eq:curv-full-lm}
\end{equation}
where \(R=XX^\top-A\). This is a GN model with positive semidefinite
structural damping \(K_X^*K_X\) and scalar LM damping \(4\gamma I\).
Its minimizer is
\(P_\gamma=-\frac12 RX(X^\top X+\gamma I)^{-1}\), whenever the
right-hand matrix is positive definite. This direction coincides with
the inexact LM direction in~\cite[Eq.~(7)]{behling2026pilm} with their damping parameter $\mu=\gamma$.
The original block update uses \(X+2P_\gamma\), whereas a midpoint
iteration uses \(X+P_\gamma\). Their accepted steps and convergence
properties therefore require separate analyses.

\subsection{Weighted observations: exact cancellation of coupling}
Let \(W=W^\top\) have nonnegative entries, let
\(\|Z\|_W^2=\sum_{i,j}w_{ij}Z_{ij}^2\), and take \(A=A^\top\).
Weights may vanish, so this notation can denote a seminorm.
Consider the extensions
\begin{equation}
f_\tau(Z)=\tfrac12\|\operatorname{sym}Z-A\|_W^2
+\tfrac{\tau}{2}\|\operatorname{skew}Z\|_W^2,\qquad \tau\ge0.
\label{eq:curv-extension}
\end{equation}
All choices give the same weighted symmetric objective. In this
subsection \(h=0\). The half-step submodel is
\begin{equation}
m_{\tau,\gamma}(P)=
\tfrac12\|R+J_X[P]\|_W^2+
\tfrac{\tau}{2}\|K_X[P]\|_W^2+2\gamma\|P\|_F^2.
\label{eq:curv-weighted-model}
\end{equation}
Writing \(\mathcal W[Z]=W\odot Z\), define
\(H_J=J_X^*\mathcal WJ_X\), \(H_K=K_X^*\mathcal WK_X\), and
\(B_\gamma=H_J+H_K+4\gamma I\).

\begin{proposition}[Separable curvature and an inner-system bound]
\label{prop:curv-cancellation}
Let \(x_i^\top,p_i^\top\) denote the rows of \(X,P\), and put
\(G_j=\sum_i w_{ij}x_ix_i^\top\). Then
\begin{equation}
\bigl(B_\gamma[P]\bigr)_j^\top=4(G_j+\gamma I)p_j .
\label{eq:curv-blockdiag}
\end{equation}
If \(B_\gamma\succ0\), \(\gamma\ge0\), and \(0<\tau\le1\), the Hessian
\(H_{\tau,\gamma}=H_J+\tau H_K+4\gamma I\) satisfies
\begin{equation}
\tau B_\gamma\preceq H_{\tau,\gamma}\preceq B_\gamma,\qquad
\kappa(B_\gamma^{-1/2}H_{\tau,\gamma}B_\gamma^{-1/2})\le\tau^{-1}.
\label{eq:curv-precondition}
\end{equation}
\end{proposition}

At \(\tau=1\), the GN coupling is canceled exactly, and the subproblem
separates into independent \(r\times r\) systems. At \(\tau=0\), it is
the standard weighted LM subproblem, which generally remains coupled.
Intermediate \(\tau\) provides a quantified condition-number bound for
a block-preconditioned inner solve. This elementary bound concerns
the linear system; it does not establish faster outer convergence.

If \(m\) entries have positive weight, forming and directly solving all
row systems costs \(O(mr^2+nr^3)\) per block. A matrix-free GN
Jacobian/adjoint application costs \(O(mr+nr)\), before preconditioning
or other overhead. Consequently, an inner GN solve with \(q\) such
iterations must be compared with the actual work of the selected
alternating solver, rather than with dense inversion of an \(nr\)-square
matrix. GNMR and scaled-gradient methods already exploit inexpensive
linear algebra and preconditioning~\cite{zilber2022gnmr,tong2021scaledgd}.

The observation pattern matters even at rank one. For zero-diagonal
\(W\), nonzero \(x_i\), and positive weighted degrees, set
\(c_{ij}=w_{ij}x_i^2x_j^2\),
\(C=[c_{ij}]\), and \(T=\operatorname{diag}(C\mathbf1)\).
At \(\gamma=0\),
\begin{equation}
\operatorname{spec}(B_0^{-1}H_J)
=\operatorname{spec}\!\left[
\tfrac12(I+T^{-1/2}CT^{-1/2})\right].
\label{eq:curv-graph-spectrum}
\end{equation}
The right-hand side is half the normalized signless Laplacian spectrum.
In particular, a connected bipartite observation graph gives a zero
GN eigenvalue although every row system in \(B_0\) is positive definite.
The missing curvature corresponds to a first-order ambiguity on the bipartite graph. The proof
and a nearly bipartite example appear in
Appendix~\ref{app:curvature}.

\subsection{A boundary where splitting gives no cheaper system}
For \(f(Z)=\frac12\|\mathcal A(Z)-b\|^2\), suppose that every measurement
matrix is symmetric, equivalently
\(\mathcal A(Z)=\mathcal A(Z^\top)\). Then
\(\mathcal A(K_X[P])=0\) and
\begin{equation}
F_\gamma(X,X+2P)=
\tfrac12\|\mathcal A(S)-b+\mathcal A(J_X[P])\|^2
+2\gamma\|P\|_F^2
\label{eq:curv-no-discount}
\end{equation}
when \(h=0\). The aBMF half-step and LM solve the same system up to
scaling. Hence the separability benefit in
Proposition~\ref{prop:curv-cancellation} is a property of the chosen
extension, not of splitting alone. Likewise, full-observation Top-\(r\)
has a specialized inexpensive GN direction~\cite{liu2015gn}.
These boundaries motivate reporting time to a common original-problem
accuracy, with all inner iterations and certificate costs included.

\section{Proofs of the Finite-Accuracy Guarantees}
\label{app:finite-proofs}
The arguments use Frobenius inner products. Matrix spectral norms are written $\|\cdot\|_2$, whereas the norm of a linear operator on matrices is induced by the Frobenius norm.

\subsection{Proof of Theorem~\ref{thm:consensus-residual}}
\begin{proof}
Since $f(W)=f(W^\top)$, a common selection $G\in\partial f(Z)$ satisfies $G^\top\in\partial f(Z^\top)$. Subtracting the two residual definitions and taking an inner product with $D=X-Y$ yields
\begin{align}
\langle D,e_-\rangle
 &=\langle D,GY-G^\top X\rangle
       +2\gamma\|D\|_F^2
       +\tfrac12\langle D,H_X-H_Y\rangle.       \label{eq:proof-residual-diff}
\end{align}
The identity
\[
DY^\top-XD^\top=Z-Z^\top-DD^\top
\]
shows that
\begin{align}
\langle D,GY-G^\top X\rangle
 &=\langle G,Z-Z^\top\rangle-\langle D,SD\rangle\nonumber\\
 &=\tfrac12\langle G-G^\top,Z-Z^\top\rangle
       -\langle D,SD\rangle.                  \label{eq:proof-loss-monotonicity}
\end{align}
Strong monotonicity gives
\[
\begin{aligned}
\langle G-G^\top,Z-Z^\top\rangle
 &\ge\sigma_f\|Z-Z^\top\|_F^2,\\
\langle D,H_X-H_Y\rangle&\ge\sigma_h\|D\|_F^2.
\end{aligned}
\]
Substitution proves \eqref{eq:consensus-residual}. If $m_\sigma>0$, dropping its nonnegative skew term gives
$m_\sigma\|D\|_F^2\le\langle D,e_-\rangle
\le\|D\|_F\|e_-\|_F$.
For $D=0$ the claimed bound is immediate; otherwise divide by
$m_\sigma\|D\|_F$. Retaining the actual quadratic form instead of its largest-eigenvalue bound proves the directional version.
\end{proof}

\subsection{Proof of Theorem~\ref{thm:geometry}}
\label{app:proof-spectral-improvement}
\begin{proof}
The Rayleigh quotient in \eqref{eq:allrank-threshold} bounds
$\langle D,SD\rangle-\sigma_f\|L_UD\|_F^2/2$ from above by $\kappa_U\|D\|_F^2$. Substitution in \eqref{eq:consensus-residual} and Cauchy--Schwarz give the residual bound, including $D=0$. At joint criticality the positive margin forces $X=Y=U$. Convexity and transpose symmetry give $S=(G+G^\top)/2\in\partial f(UU^\top)$, and $(H_X+H_Y)/2\in\partial h(U)$. Adding the zero block residuals gives $2SU+(H_X+H_Y)/2=0$, the stationarity condition for the symmetric objective. The inequality $\kappa_U\le\lambda_{\max}(S)$ follows from $L_U^*L_U\succeq0$ and $\sigma_f\ge0$. For $\sigma_f>0$ compactness of the unit sphere implies that the old and new largest eigenvalues coincide precisely when a nonzero $H$ has all columns in $E$ and $L_UH=0$.

If $U$ is column-rank deficient, take $H=va^\top$ for nonzero $v\in E$ and nonzero $a\in\ker U$. For full column rank $U$, we claim that $\ker L_U=\{UC:C=C^\top\}$. Left multiplication of $HU^\top=UH^\top$ by $I-P_U$ gives $(I-P_U)HU^\top=0$ and hence $(I-P_U)H=0$. Writing $H=UC$ then gives $U(C-C^\top)U^\top=0$, which forces $C=C^\top$. The converse is direct. If $v=Ua$ is a nonzero vector in $E$, $H=Uaa^\top$ satisfies the two equality conditions. Conversely, any such $H$ in the kernel has a nonzero column in $E\cap\operatorname{range}(U)$. This proves \eqref{eq:allrank-strict}.

\end{proof}

\subsection{Proof of Proposition~\ref{prop:compressed-spectral}}
\begin{proof}
Write $H=H_\parallel+H_\perp$ with $H_\parallel=P_UH$. The within-subspace block and the two off-diagonal blocks of $L_UH$ are Frobenius-orthogonal. Thus
\[
 \tfrac12\|L_UH\|_F^2
 =\tfrac12\|L_UH_\parallel\|_F^2+\|H_\perp U^\top\|_F^2
 \ge s_{\min}(U)^2\|H_\perp\|_F^2 .
\]
This proves \eqref{eq:compressed-threshold}; equality with the original top eigenvalue again requires a unit vector in both $E$ and $\operatorname{range}(U)$ when $\alpha>0$. For \eqref{eq:angle-guarantee}, $\lambda_{\max}(S)I-S\succeq g(I-P_E)$. Adding $\alpha(I-P_U)$ gives a lower bound by $\min(g,\alpha)(2I-P_E-P_U)$. For orthonormal bases $V_E,V_U$, the Gram matrix of $[V_E,V_U]$ has identity diagonal blocks and off-diagonal block $V_E^\top V_U$. Its largest eigenvalue is $1+\|V_E^\top V_U\|_2=1+\|P_EP_U\|_2$, also the largest eigenvalue of $P_E+P_U$. Taking minimum eigenvalues proves the stated gap.
\end{proof}

\subsection{Proof of Theorem~\ref{thm:midpoint-quadratic}}
\begin{proof}
Set $M=UU^\top$, $H=M-DD^\top/4$, and
$K=(DU^\top-UD^\top)/2$. Then $H$ is symmetric, $K$ is skew-symmetric, and $Z=H+K$. The residual sum satisfies
\begin{equation}
e_+=2(\operatorname{sym}G)U
          -(\operatorname{skew}G)D
          +\tfrac12\{\nabla h(X)+\nabla h(Y)\}.
\label{eq:proof-residual-sum}
\end{equation}
The gradient of $h$ is affine, so its average in
\eqref{eq:proof-residual-sum} equals $\nabla h(U)$.
The transpose-commuting property of $\mathcal Q$ maps symmetric matrices to symmetric matrices and skew-symmetric matrices to skew-symmetric matrices. Since $B$ is symmetric,
\[
\operatorname{sym}G=\mathcal QH-B,\qquad
\operatorname{skew}G=\mathcal QK.
\]
Also, $\nabla\Phi(U)=2(\mathcal QM-B)U+\nabla h(U)$.
Subtracting \eqref{eq:proof-residual-sum} therefore gives exactly
\[
\nabla\Phi(U)-e_+
 =\tfrac12\mathcal Q(DD^\top)U+(\mathcal QK)D.
\]
For the norm estimate use
\[
\begin{aligned}
\|DD^\top\|_F&\le\|D\|_F^2,\\
\|K\|_F&\le\|U\|_2\|D\|_F,\\
\|D\|_2&\le\|D\|_F.
\end{aligned}
\]
The first term is bounded by $L_f\|U\|_2\|D\|_F^2/2$, and the second by $L_f\|U\|_2\|D\|_F^2$. Their sum proves
\eqref{eq:midpoint-quadratic-bound}. For the scalar example,
$\mathcal Q$ is the identity, $K=0$, and the identity reduces to
$\nabla\Phi(u)-e_+=uD^2/2$, proving that the order is attained.
\end{proof}

\subsection{Proof of Theorem~\ref{thm:midpoint-smooth}}
\begin{proof}
Retain $M,H,K$ from the preceding proof. Differentiating transpose symmetry gives
\[
\nabla f(W^\top)=\nabla f(W)^\top.
\]
Since $H^\top=H$ and $K^\top=-K$,
\begin{align}
\operatorname{sym}G
 &=\tfrac12\{\nabla f(H+K)+\nabla f(H-K)\},\nonumber\\
\operatorname{skew}G
 &=\tfrac12\{\nabla f(H+K)-\nabla f(H-K)\}.
\label{eq:proof-transpose-gradient}
\end{align}
The Hessian bound makes $\nabla f$ $L_f$-Lipschitz on the stated segments, so
\begin{equation}
\|\operatorname{skew}G\|_F\le L_f\|K\|_F.
\label{eq:proof-skew-gradient}
\end{equation}
The Lipschitz Hessian gives the Taylor remainder estimate
\[
\|\nabla f(H+V)-\nabla f(H)-\nabla^2 f(H)[V]\|_F
 \le\tfrac12M_f\|V\|_F^2 .
\]
Apply it to $V=K$ and $V=-K$ and average. The linear terms cancel, yielding
\begin{equation}
\|\operatorname{sym}G-\nabla f(H)\|_F
 \le\tfrac12M_f\|K\|_F^2.
\label{eq:proof-central-cancellation}
\end{equation}
Furthermore,
\[
\|\nabla f(M)-\nabla f(H)\|_F
 \le\tfrac14L_f\|DD^\top\|_F.
\]
Using \eqref{eq:proof-residual-sum}, the smooth objective gradient obeys
\begin{align}
\nabla\Phi(U)-e_+
={}&2\{\nabla f(M)-\operatorname{sym}G\}U
       +(\operatorname{skew}G)D \nonumber\\
 &+\nabla h(U)-\tfrac12\{\nabla h(X)+\nabla h(Y)\}.
\label{eq:proof-midpoint-error}
\end{align}
The first line of \eqref{eq:proof-midpoint-error} is bounded in Frobenius norm by
\[
\left(\tfrac32L_fR+M_fR^3\right)\|D\|_F^2,
\]
using $\|K\|_F\le R\|D\|_F$ and the norm inequalities from the preceding proof. Applying the same Taylor-remainder argument to $\nabla h$ at $U$ with offsets $\pm D/2$ bounds the second line by
$M_h\|D\|_F^2/8$. Summing proves the claim.

For the weaker budget \eqref{eq:lip-gradient-budget}, only gradient Lipschitz continuity is assumed. In that case
\[
\|\operatorname{sym}G-\nabla f(H)\|_F\le L_f\|K\|_F
\]
replaces \eqref{eq:proof-central-cancellation}. The loss part of
\eqref{eq:proof-midpoint-error} is consequently bounded by
$2L_fR^2\|D\|_F+(3/2)L_fR\|D\|_F^2$, as asserted.
\end{proof}

\subsection{Proof of Theorem~\ref{thm:midpoint-proximal}}
\begin{proof}
Convexity implies $U\in\operatorname{dom}h$ and $h(U)$ is finite. Subgradient inequalities at the two endpoints give, for every $V$,
\begin{align*}
h(V)
 &\ge\tfrac12\{h(X)+h(Y)\}\\
 &\quad+\tfrac12\langle H_X,V-X\rangle
      +\tfrac12\langle H_Y,V-Y\rangle\\
 &=h(U)+\langle\bar H,V-U\rangle-\epsilon_h .
\end{align*}
Moreover,
\begin{align*}
2\epsilon_h
={}&\bigl[h(U)-h(X)-\langle H_X,U-X\rangle\bigr]\\
 &+\bigl[h(U)-h(Y)-\langle H_Y,U-Y\rangle\bigr]\ge0.
\end{align*}
Thus $\bar H\in\partial_{\epsilon_h}h(U)$.

Let $V=\operatorname{prox}_{th}(U-t\nabla g(U))$ and
$p=(U-V)/t$. The proximal minimizer exists uniquely because $h$ is proper, closed, and convex. Its optimality condition gives
\[
H_V=p-\nabla g(U)\in\partial h(V).
\]
Combining the $\epsilon_h$-subgradient inequality at $U$ with the exact subgradient inequality at $V$ yields
\[
\langle\bar H-H_V,U-V\rangle\ge-\epsilon_h.
\]
Since $U-V=tp$, this becomes
\[
\|p\|_F^2
 \le\langle\nabla g(U)+\bar H,p\rangle+\epsilon_h/t
 \le a\|p\|_F+\epsilon_h/t.
\]
The nonnegative root of the resulting quadratic inequality is
\eqref{eq:proximal-certificate}; its second inequality follows from
$\sqrt{a^2+4b}\le a+2\sqrt b$ for $a,b\ge0$.
The stated budgets $b_f$ follow from the loss terms of
\eqref{eq:proof-midpoint-error}; those terms do not use differentiability of $h$.

It remains to prove \eqref{eq:epsilon-h-residual-bound}. Convexity of $h$ gives the Jensen bound
\[
h(U)-\tfrac12\{h(X)+h(Y)\}\le0,
\]
and hence
$\epsilon_h\le\langle H_X-H_Y,D\rangle/4$.
The exact residual identity
\eqref{eq:proof-residual-diff}--\eqref{eq:proof-loss-monotonicity} gives
\begin{align*}
\tfrac14\langle H_X-H_Y,D\rangle
={}&\tfrac12\langle D,e_-\rangle-\gamma\|D\|_F^2
       +\tfrac12\langle D,SD\rangle\\
 &-\tfrac14\langle G-G^\top,Z-Z^\top\rangle.
\end{align*}
Keeping the last inner product and using $D\ne0$ gives the quadratic upper bound below with $m_0=2\gamma-q_D+b_f^{\rm tr}/\|D\|_F^2$. Dropping the nonnegative pairing gives the simpler margins. Consequently, for $m_0>0$,
\[
\epsilon_h
 \le\tfrac12\|D\|_F\|e_-\|_F-\tfrac12m_0\|D\|_F^2
 \le\frac{\|e_-\|_F^2}{8m_0}.
\]
The last inequality completes the square in $\|D\|_F$. Using the actual directional quadratic form instead of its eigenvalue bound proves the alternative expression for $m_0$. Theorem~\ref{thm:consensus-residual} with $\sigma_h=0$ also gives $\|D\|_F\le\|e_-\|_F/m_0$, which together with
\eqref{eq:proximal-certificate} proves
\eqref{eq:finite-stopping-nonsmooth}.
\end{proof}

\section{Proofs for the Stagewise Penalty Algorithm}
\label{app:adaptive-proofs}

\subsection{Proof of Proposition~\ref{prop:penalty-obstruction}}
Set $p=xy$. If $p\ge0$, the objective is at least
$\frac12(p+1)^2\ge1/2$. If $p<0$, then
$(x-y)^2=(x+y)^2-4p\ge-4p$, with equality if $x=-y$. Hence
\[
 F_\gamma(x,y)\ge \frac12(p+1)^2-2\gamma p.
\]
For $0<\gamma<1/2$, the latter quadratic is minimized over $p<0$
at $p=2\gamma-1$, with value $2\gamma-2\gamma^2<1/2$.
Equality is attained by the factors stated in the proposition.
The value is strictly increasing on $(0,1/2)$, since its derivative is
$2-4\gamma>0$. Raising the penalty from an old global minimizer
therefore raises the smallest possible new objective value.

\subsection{Uniform bounds and descent}
Write $z=(X,Y)$ and use the product Frobenius norm. The reset test
ensures that each trial starts with $F_\gamma(z^0)\le E_0$. Since
$\gamma\ge\gamma_{\min}$,
\[
 F_{\gamma_{\min}}(z)\le F_\gamma(z).
\]
Any decrease in the current $F_\gamma$ consequently keeps the iterates
in $\mathcal S$.

For fixed $Y$, the gradient of $\psi_\gamma(\cdot,Y)$ has Lipschitz
constant at most $L\|Y\|_2^2+\gamma\le LR^2+\gamma=:L_b(\gamma)$. This bound is global in
the updated block, so the argument does not presuppose that the next
factor lies in $\mathcal S$. The proximal minimization and the descent
lemma give
\[
\begin{aligned}
 &F_\gamma(X,Y)-F_\gamma(X^+,Y)\\
 &\quad\ge \left(\frac{1}{2\alpha_\gamma}
                  -\frac{L_b(\gamma)}{2}\right)
       \|X^+-X\|_F^2\\
 &\quad=\frac{L_b(\gamma)}2\|X^+-X\|_F^2\\
 &\quad\ge a\|X^+-X\|_F^2.
\end{aligned}
\]
The intermediate pair thus lies in $\mathcal S$. The same argument
applies to the $Y$ update with $X^+$ fixed. Therefore
\begin{equation}
 F_\gamma(z^k)-F_\gamma(z^{k+1})
 \ge a\|z^{k+1}-z^k\|_F^2,
 \label{eq:adaptive-inner-descent}
\end{equation}
where $a=(LR^2+\gamma_{\min})/2$.
The lower bound $F_\gamma\ge F_{\gamma_{\min}}\ge F_*$ implies
\begin{equation}
 a\sum_{k=0}^{N-1}\|z^{k+1}-z^k\|_F^2\le E_0-F_*.
 \label{eq:adaptive-step-sum}
\end{equation}

The gradient bound follows from
$\|XY^\top\|_F\le\|X\|_F\|Y\|_2\le R^2$:
\[
 \|\nabla f(XY^\top)\|_2
 \le\|\nabla f(0)\|_2+L\|XY^\top\|_F\le B.
\]
This also holds on the product of the two radius-$R$ factor balls,
which will be used to bound gradients between consecutive iterates.

\subsection{Residuals generated by the inner updates}
The optimality equations of the two proximal maps give the valid
selections in \eqref{eq:prox-subgradient-selections}. At the common
endpoint $z^+=(X^+,Y^+)$, the residuals are
\begin{align*}
 e_X^+&=\nabla_X\psi_\gamma(z^+)
       -\nabla_X\psi_\gamma(z)
       -(X^+-X)/\alpha,\\
 e_Y^+&=\nabla_Y\psi_\gamma(X^+,Y^+)
       -\nabla_Y\psi_\gamma(X^+,Y)
       -(Y^+-Y)/\alpha.
\end{align*}
On the product of radius-$R$ factor balls, a joint gradient
Lipschitz constant is
\[
 L_\psi=2LR^2+B+2\Gamma.
\]
Indeed, writing $d_X=\|X_1-X_2\|_F$ and
$d_Y=\|Y_1-Y_2\|_F$, one has
\begin{align*}
 \|\nabla_X\psi_\gamma(z_1)-\nabla_X\psi_\gamma(z_2)\|_F
 &\le (LR^2+\Gamma)d_X\\
 &\quad +(LR^2+B+\Gamma)d_Y.
\end{align*}
The analogous inequality interchanges the two coefficients.
The spectral norm of this symmetric $2$-by-$2$ coefficient matrix
is $L_\psi$, proving the claimed joint bound.
Within this trial $\alpha=\alpha_\gamma$, so
$\alpha^{-1}\le2\overline L_b$. It follows, conservatively, that
\[
 \|(e_X^+,e_Y^+)\|_F
 \le \sqrt{2}(L_\psi+\alpha^{-1})\|z^+-z\|_F,
\]
and hence
\begin{equation}
 \max\{\|e_+^+\|_F,\|e_-^+\|_F\}
 \le C_r\|z^+-z\|_F,
 \quad C_r=2(L_\psi+2\overline L_b).
 \label{eq:adaptive-residual-step}
\end{equation}
Combining \eqref{eq:adaptive-step-sum} and
\eqref{eq:adaptive-residual-step}, the inner residual test must be met
within
\[
 N_\tau=1+\left\lceil
 \frac{(E_0-F_*)C_r^2}{a\tau^2}
 \right\rceil
\]
iterations. Otherwise each of the first $N_\tau$ decreases would
exceed $a\tau^2/C_r^2$, contradicting the available energy decrease.

\subsection{Proof of Theorem~\ref{thm:adaptive-certified}}
For a nonzero disagreement, the directional form of
\eqref{eq:transpose-margins}, using $\sigma_h=0$ and retaining the exact pairing, gives
\[
 \|D\|_F\le \|e_-\|_F/m_{\rm dir}.
\]
On an accepted trial, this is at most $\tau/\underline m$.
The exact-pairing form of Theorem~\ref{thm:midpoint-proximal} gives
$\epsilon_h\le\|e_-\|_F^2/(8m_{\rm dir})$.
The loss budget \eqref{eq:lip-gradient-budget} and
$\|D\|_F\le\|X\|_F+\|Y\|_F\le2R$ imply
\begin{align*}
 b_f
 &\le 2LR^2\|D\|_F+\tfrac32 LR\|D\|_F^2\\
 &\le5LR^2\|D\|_F.
\end{align*}
Consequently,
\begin{align*}
 \|\mathcal G_t(U)\|_F
 &\le\|e_+\|_F+5LR^2\|D\|_F
       +\frac{\|e_-\|_F}{\sqrt{8m_{\rm dir}t}}\\
 &\le
 \tau\left(1+\frac{5LR^2}{\underline m}
                +\frac1{\sqrt{8\underline m t}}\right)
 =\varepsilon.
\end{align*}
If $D=0$, both endpoints coincide with $U$,
$\bar H=(H_X+H_Y)/2\in\partial h(U)$, and
$e_+=\nabla g(U)+\bar H$.
The same proximal argument with $\epsilon_h=0$ gives
$\|\mathcal G_t(U)\|_F\le\|e_+\|_F\le\tau$.
This case uses no directional quotient.

For every nonzero $D$ in $\mathcal S$,
\[
 \frac{\langle D,\nabla f(XY^\top)D\rangle}{\|D\|_F^2}
 \le\|\nabla f(XY^\top)\|_2\le B.
\]
Thus at $\gamma=\Gamma$,
$m_{\rm dir}\ge2\Gamma-B\ge\underline m$.
The geometric retry sequence reaches this cap after at most $J$
trials, including its initial trial. Each trial takes at most
$N_\tau$ inner iterations by the preceding argument. This proves
termination and \eqref{eq:adaptive-complexity}, including all rejected
trials and all resets.

Since $U=(X+Y)/2$, the matrix $UU^\top$ is PSD of rank at most $r$.
Convexity of $\operatorname{dom}h$ places $U$ in the same factor
constraint set when constraints are encoded by $h$.

\subsection{Proof of Corollary~\ref{cor:adaptive-cluster}}
The accepted factors have norms at most $R$, and
$\|X_s-Y_s\|_F\le\tau_s/\underline m\to0$.
Each returned midpoint satisfies
$\|\mathcal G_t(U_s)\|_F\le\varepsilon_s\to0$.
For proper closed convex $h$, $\operatorname{prox}_{th}$ is continuous;
$\nabla g$ is also continuous. Therefore the mapping
$\mathcal G_t$ is continuous on the factor space. Along any convergent
subsequence $U_{s_j}\to U_\star$, one has
$\mathcal G_t(U_\star)=0$. The proximal optimality condition gives
$0\in\nabla g(U_\star)+\partial h(U_\star)$.
This is original-problem criticality and in particular implies
$U_\star\in\operatorname{dom}h$; no separate continuity assumption on
the values $h(U_s)$ is needed.

For geometric accuracies, if $s_*$ is the first stage with
$\varepsilon_{s_*}\le\varepsilon$, then
$q\varepsilon<\varepsilon_{s_*}\le\varepsilon$, apart from the
trivial initial-stage case, and
\[
 \sum_{s=0}^{s_*}\varepsilon_s^{-2}
 \le \frac{\varepsilon_{s_*}^{-2}}{1-q^2}
 < \frac{\varepsilon^{-2}}{q^2(1-q^2)}.
\]
Summing the per-stage bound proves the stated cumulative complexity.
The additive number of stages is logarithmic and is absorbed by the
same asymptotic bound.

\section{Proofs for the Unregularized Case}
\label{app:unregularized}
\label{app:hzero}

\subsection{Unregularized certificates and stopping rules}
\label{app:hzero-general}
With $h=0$, Corollary~\ref{cor:finite-stopping} gives \eqref{eq:hzero-certificate} with $\sigma_h=M_h=0$. At $D=0$, \eqref{eq:hzero-residuals} gives $\nabla\Phi(U)=e_+$. For the Lipschitz-gradient case, Assumption~\ref{ass:adaptive} gives $\|D\|_F\le 2R$, so \eqref{eq:lip-gradient-budget} yields $b_f\le 5LR^2\|D\|_F$. With $\epsilon_h=0$, the accepted margin bounds the midpoint gradient by $\|e_+\|_F+(5LR^2/\underline m)\|e_-\|_F$, which gives \eqref{eq:hzero-lipschitz-stop}.

For $C_0>0$, the two tests in \eqref{eq:hzero-asymmetric-stop} follow directly from \eqref{eq:hzero-certificate} by allocating $\varepsilon/2$ to each term. Using their smaller tolerance $\tau_0$ in \eqref{eq:adaptive-complexity} gives the stated work bound since
\[
 \tau_0^{-2}=\max\left\{4\varepsilon^{-2},
                    \frac{2C_0}{\underline m^2\varepsilon}\right\}.
\]
At decreasing requested accuracies, the output conclusion follows from Corollary~\ref{cor:adaptive-cluster} with the same energy, margin and retry safeguards.

\subsection{Proof of Proposition~\ref{prop:hzero-sublevel}}
The nonnegative penalty and lower bound on the loss give
\[
 f(XY^\top)\le E_0,\qquad
 \|D\|_F^2\le\frac{2(E_0-f_*)}{\gamma_{\min}}.
\]
Write $Z=XY^\top$. Since $\operatorname{rank}Z\le r$, the trace and nuclear-norm inequalities imply
\[
 \operatorname{tr}Z\le|\operatorname{tr}Z|
 \le\|Z\|_*\le\sqrt r\,\|Z\|_F\le\sqrt r\,M_E.
\]
Finally,
\[
 \|X\|_F^2+\|Y\|_F^2=2\operatorname{tr}(XY^\top)+\|D\|_F^2,
\]
which proves \eqref{eq:hzero-factor-bound}. In the unweighted least-squares example, the triangle inequality gives
$\|Z\|_F\le\|A\|_F+\sqrt{2E_0}$. For the missing-direction example, $XY^\top$ has zero $(1,1)$ entry and $D=0$ for every $t$, while the factor norms diverge.

\subsection{Proof of Proposition~\ref{prop:hzero-weighted-counterexample}}
At the stated factors,
\[
 G=W\odot(XY^\top-A)
   =\begin{pmatrix}-5/12&-3/4\\-3/4&-5/12\end{pmatrix}.
\]
This symmetric matrix satisfies $GX=\frac13X=2\gamma X$.
Since $Y=-X$ and $D=2X$, one has
\[
\begin{aligned}
 GY+\gamma D&=-GX+2\gamma X=0,\\
 G^\top X-\gamma D&=GX-2\gamma X=0.
\end{aligned}
\]
The factors are nonzero and unequal, proving joint criticality without consensus. The loss Hessian is the entrywise multiplication operator by $W$, whose smallest and largest eigenvalues in the Frobenius geometry are $1/3$ and $1$. Thus none of the claimed smoothness, strong convexity, or coercivity properties relies on a missing entry.

\subsection{Proof of Lemma~\ref{lem:hzero-balance}}
For a common $G$ and $h=0$, the two residuals are
$e_X=GY+\gamma(X-Y)$ and $e_Y=G^\top X-\gamma(X-Y)$.
Direct multiplication cancels the loss terms:
\begin{align*}
X^\top e_X-e_Y^\top Y
&=\gamma\{X^\top(X-Y)+(X-Y)^\top Y\}\\
&=\gamma(X^\top X-Y^\top Y).
\end{align*}
Submultiplicativity gives \eqref{eq:hzero-balance-bound}. At a joint critical point both residuals vanish, proving Gram balance for $\gamma\ne0$. This proof remains valid for a nonsmooth loss when the same admissible loss subgradient is used in the two residuals.

\subsection{Proof of Theorem~\ref{thm:hzero-sharp}}
Write $P=X^\top X$, $Q=Y^\top Y$, $T=(P+Q)/2$, and $C=P-Q$.
For the full-observation quadratic loss,
\[
e_-=XQ-YP+(A+2\gamma I)(X-Y).
\]
Since $X=U+D/2$ and $Y=U-D/2$, one has
$XQ-YP=DT-UC$, proving \eqref{eq:hzero-sylvester}.
Furthermore,
\[
T=U^\top U+\tfrac14D^\top D\succeq\tfrac14D^\top D.
\]
Taking the Frobenius inner product of
$e_-+UC=(A+2\gamma I)D+DT$ with $D$ gives
\begin{align*}
\langle D,e_-+UC\rangle
&\ge\delta\|D\|_F^2+\operatorname{tr}(T D^\top D)\\
&\ge\delta\|D\|_F^2+\tfrac14\|D^\top D\|_F^2\\
&\ge\delta d^2+\frac{d^4}{4r}.
\end{align*}
The last inequality uses $(\operatorname{tr}(D^\top D))^2\le r\|D^\top D\|_F^2$.
Cauchy--Schwarz bounds the left side above by $d\eta$.
For $d>0$, division gives \eqref{eq:hzero-cubic-certificate}; for $d=0$ it is immediate. The linear and cubic consequences follow by dropping nonnegative terms. Lemma~\ref{lem:hzero-balance} and $\|UC\|_F\le\|U\|_2\|C\|_F$ give \eqref{eq:hzero-eta-residual}.

At any critical point with $\gamma>0$, Lemma~\ref{lem:hzero-balance} gives $C=0$. If $A+2\gamma I\succeq0$, the cubic inequality forces $D=0$, including the semidefinite boundary. Equivalently, the homogeneous Sylvester equation yields $DT^{1/2}=0$; because $X^\top X=Y^\top Y=T$, the rows of $D$ lie in $\operatorname{range}(T)$, and therefore $D=0$.

For necessity, suppose $A$ has a unit eigenvector $v$ with eigenvalue $\lambda<-2\gamma$. Let $q\in\mathbb R^r$ be a unit vector and set
\[
s=\sqrt{-\lambda-2\gamma},\qquad
X=svq^\top,\quad Y=-svq^\top.
\]
Then $X\ne Y$ and
\[
e_X=s(s^2+\lambda+2\gamma)vq^\top=0,\qquad e_Y=-e_X=0.
\]
This supplies an unequal critical point and proves necessity.
For PSD $A$, replacing the actual $\delta$ by its lower bound $2\gamma$ preserves the inequality. The output bound follows from the quadratic midpoint identity with $L_f=1$.

\subsection{Sublevel boundedness and proof of Proposition~\ref{prop:hzero-am}}
Because both terms in \eqref{eq:hzero-ls} are nonnegative for $\gamma>0$, $F_\gamma\le E$ implies
\[
\|XY^\top\|_F\le M_E,
\qquad \|D\|_F\le d_E.
\]
The midpoint identity gives
\[
UU^\top=\operatorname{sym}(XY^\top)+\tfrac14DD^\top.
\]
Since $UU^\top$ has rank at most $r$,
\begin{align*}
\|U\|_F^2
&=\operatorname{tr}(UU^\top)
 \le\sqrt r\,\|UU^\top\|_F\\
&\le\sqrt r\,(M_E+d_E^2/4).
\end{align*}
The bounds for $X=U+D/2$ and $Y=U-D/2$ follow by the triangle inequality. Thus \eqref{eq:hzero-sublevel-radius} bounds the entire sublevel set, not only a particular AM trajectory. Applying it with $\gamma=\gamma_{\min}>0$ and the common reference level supplies the radius required by the stagewise framework.

For each fixed block, the Hessian in the other block is right multiplication by a Gram matrix plus $\gamma I$. It is uniformly at least $\gamma I$, so the unique minimizers are \eqref{eq:hzero-am-updates}. Strong convexity and exact minimization give
\begin{equation}
\begin{aligned}
&F_\gamma(X^k,Y^k)-F_\gamma(X^{k+1},Y^{k+1})\\
&\quad\ge\frac\gamma2\|X^{k+1}-X^k\|_F^2\\
&\qquad+\frac\gamma2\|Y^{k+1}-Y^k\|_F^2.
\end{aligned}
\label{eq:hzero-am-descent}
\end{equation}
The objective is bounded below by zero. Telescoping proves square summability of the successive block changes and confines every pair to the initial bounded sublevel set. The last block equation gives
$e_Y(X^{k+1},Y^{k+1})=0$. The first block equation gives
$e_X(X^{k+1},Y^k)=0$.
The polynomial map $\nabla_X F_\gamma$ is Lipschitz on a ball containing these pairs. Therefore
\[
\|e_X(X^{k+1},Y^{k+1})\|_F
 \le L_0\|Y^{k+1}-Y^k\|_F\longrightarrow0
\]
for a finite constant $L_0$. Both current residuals tend to zero. Continuity makes every accumulation point critical, and Theorem~\ref{thm:hzero-sharp} gives equal factors for PSD $A$. For exact block solves \eqref{eq:block-dec}--\eqref{eq:block-res} hold with $a=\gamma/2$ and $b=0$. Here $h=0$ and $F_\gamma$ is polynomial, hence KL. Theorem~\ref{thm:fixed-kl} therefore gives finite length and convergence of the full sequence. At its diagonal limit either block equation gives $\nabla\Phi(U_*)=0$.

\subsection{Shifted subspace iteration}
Put $M=A+\gamma I\succ0$. Every right-hand inverse in
\eqref{eq:hzero-am-updates} is invertible, and hence
\[
\operatorname{range}(Y^{k+1})
 =\operatorname{range}(M^2Y^k)
 =\operatorname{range}(M^{2k+2}Y^0).
\]
There is consequently an invertible $r\times r$ matrix $R_k$ such that
$Y^k=M^{2k}Y^0R_k$. Write
$M=V_1\Lambda_1V_1^\top+V_2\Lambda_2V_2^\top$.
The assumed nonsingularity of $V_1^\top Y^0$ persists and
\begin{equation*}
\begin{split}
V_2^\top Y^k(V_1^\top Y^k)^{-1}
=\Lambda_2^{2k}V_2^\top Y^0
 (V_1^\top Y^0)^{-1}\Lambda_1^{-2k}.
\end{split}
\end{equation*}
Taking spectral norms proves \eqref{eq:hzero-subspace-rate}.
For a positive eigengap, the derivative of
$(\lambda_{r+1}+\gamma)/(\lambda_r+\gamma)$ with respect to $\gamma$ is
$(\lambda_r-\lambda_{r+1})/(\lambda_r+\gamma)^2>0$.
This proves the claimed ordering of the subspace ratios and no stronger runtime or factor-scaling conclusion.

\section{Proofs for Computational Curvature}
\label{app:curvature}

\subsection{Proof of Proposition~\ref{prop:curv-bregman}}
\begin{proof}
For differentiable transpose-symmetric \(f\), differentiating
\(f(Z^\top)=f(Z)\) yields
\(\nabla f(Z^\top)=\nabla f(Z)^\top\).
Since \(M=M^\top\), its gradient is symmetric, whereas
\(K_X[P]^\top=-K_X[P]\). Therefore
\(\langle\nabla f(M),K_X[P]\rangle=0\), and
\[
f(M+K_X[P])=f(M)+\mathcal D_f(M+K_X[P],M).
\]
Furthermore,
\[
X(X+2P)^\top=S+2XP^\top=M+K_X[P].
\]
Substituting these identities and the definition of
\(\mathcal J_h\) into \(F_\gamma(X,X+2P)\) proves
\eqref{eq:curv-bregman}. Convexity of \(f\) gives nonnegativity of its
Bregman divergence, and convexity of \(h\) gives nonnegativity of its
midpoint Jensen gap. Both endpoint values of \(h\) are finite by
hypothesis, which also makes the midpoint value finite. The strong
convexity and smoothness inequalities applied from \(M\) to
\(M+K_X[P]\) give the stated bounds.
\end{proof}

\subsection{Proof of Proposition~\ref{prop:curv-cancellation}}
\begin{proof}
Symmetry of \(W\) implies weighted orthogonality of symmetric and
skew-symmetric matrices, and invariance of the weighted seminorm under
transposition. Thus
\[
\begin{split}
&\langle P,(H_J+H_K)[P]\rangle\\
&\quad=\|XP^\top+PX^\top\|_W^2+
  \|XP^\top-PX^\top\|_W^2\\
&\quad=2\|XP^\top\|_W^2+2\|PX^\top\|_W^2\\
&\quad=4\sum_jp_j^\top G_jp_j.
\end{split}
\]
Both sides represent self-adjoint quadratic forms, so polarization
identifies their operators. Adding \(4\gamma I\) yields
\eqref{eq:curv-blockdiag}. Since \(H_J,H_K\succeq0\) and \(\gamma\ge0\),
\[
B_\gamma-H_{\tau,\gamma}=(1-\tau)H_K\succeq0
\]
and
\[
H_{\tau,\gamma}-\tau B_\gamma
=(1-\tau)(H_J+4\gamma I)\succeq0.
\]
Congruence by \(B_\gamma^{-1/2}\) gives all eigenvalues in \([\tau,1]\),
which proves \eqref{eq:curv-precondition}.
\end{proof}

For completeness, at \(\tau=1\), minimization in the original endpoint
variable \(Y=X+2P\) gives the independent equations
\[
(G_j+\gamma I)y_j=\sum_iw_{ij}a_{ij}x_i+\gamma x_j.
\]
A Cholesky factorization is sufficient when the row matrix is positive
definite. The \(O(mr^2+nr^3)\) bound counts construction of all \(G_j\),
right-hand sides, and dense row factorizations. It does not assume that
a proximal or coordinate implementation must solve every block exactly.
For the full GN model,
\((J_X[P])_{ij}=x_i^\top p_j+p_i^\top x_j\) needs \(O(r)\) work per
observed entry; accumulating its adjoint has the same order. Arrays of
size \(nr\) add the displayed vector-operation cost. These counts do
not include row-preconditioner construction, setup, line searches, or
stopping diagnostics, which must be included in runtime comparisons.

\subsection{Proof of the rank-one graph identity}
We prove \eqref{eq:curv-graph-spectrum}. For \(r=1\), regard \(X=x\)
and \(P=p\) as vectors. Let \(E=\operatorname{diag}(x)\), which is
invertible by assumption, and write \(p=Ez\). Define
\(g_j=\sum_iw_{ij}x_i^2\). Positive weighted degrees mean
\(x_j^2g_j>0\), and
\[
E^\top B_0E=4\operatorname{diag}(x_j^2g_j)=4T.
\]
In addition,
\[
\begin{split}
p^\top H_Jp
&=\sum_{i,j}w_{ij}(x_ip_j+p_ix_j)^2\\
&=\sum_{i,j}c_{ij}(z_i+z_j)^2\\
&=2z^\top(T+C)z,
\end{split}
\]
where symmetry and zero diagonal of \(C\) have been used.
Thus the generalized eigenproblem \(H_Jp=\lambda B_0p\) is congruent
to
\(2(T+C)z=4\lambda Tz\). Whitening by \(T^{1/2}\) proves
\eqref{eq:curv-graph-spectrum}.

The quadratic-form representation shows that this matrix is positive
semidefinite. Applying \((z_i+z_j)^2\le2(z_i^2+z_j^2)\) shows its
eigenvalues are at most one. The vector \(z=\mathbf1\) gives the
eigenvalue one. For a connected observation graph, a zero eigenvalue
requires \(z_i=-z_j\) on every edge. A nonzero such vector exists
exactly when the graph is bipartite: it is constant on each part with
opposite signs. An odd cycle forces its entries to be zero, and
connectedness then forces all entries to vanish.

Finally, take \(x_i=1\) for three vertices, with
\(w_{12}=w_{21}=w_{23}=w_{32}=1\) and
\(w_{13}=w_{31}=\varepsilon>0\).
The matrices \(B_0=4\operatorname{diag}(1+\varepsilon,2,1+\varepsilon)\)
remain uniformly positive definite as \(\varepsilon\downarrow0\).
For each positive \(\varepsilon\), the graph is nonbipartite, but its
smallest preconditioned GN eigenvalue tends to zero by continuity,
whereas its largest remains one. Consequently, no positive uniform
lower spectral bound follows from positive row degrees alone.
The loss of curvature in the limit is an identifiability property
of the observation model. Adding \(\tau>0\) controls the inner-system
condition number as in \eqref{eq:curv-precondition}, while changing the
outer direction.

\subsection{Derivation of the symmetric-measurement boundary}
For a linear measurement operator with symmetric measurement
matrices, \(\mathcal A(Z^\top)=\mathcal A(Z)\). Hence
\[
\mathcal A(K_X[P])=0,\qquad
\mathcal A(J_X[P])=2\mathcal A(XP^\top).
\]
Since \(X(X+2P)^\top=S+2XP^\top\), direct substitution into the
quadratic aBMF loss gives \eqref{eq:curv-no-discount}.
Equivalently, define \(\mathcal L_X[P]=\mathcal A(XP^\top)\).
The endpoint-step normal equation is
\[
(\mathcal L_X^*\mathcal L_X+\gamma I)[\Delta]
=-\mathcal L_X^*(\mathcal A(S)-b).
\]
The LM normal equation with scalar damping \(4\gamma\) is
\[
4(\mathcal L_X^*\mathcal L_X+\gamma I)[P]
=-2\mathcal L_X^*(\mathcal A(S)-b).
\]
The substitution \(\Delta=2P\) makes these equations identical.
No conclusion of cheaper aBMF linear algebra follows in this case.

\section{A Rank-Two SymNMF Counterexample}
\label{app:symnmf-counterexample}
The following counterexample is due to Hu et al.~\cite{hu2026abmf}. For symmetric nonnegative matrix factorization, let $h$ be the indicator of the nonnegative orthant and set
\[
\begin{gathered}
A=\begin{pmatrix}0&1\\1&0\end{pmatrix},\qquad
\gamma=\tfrac38,\\
X=\tfrac14\begin{pmatrix}3&1\\1&3\end{pmatrix},\quad
Y=\tfrac14\begin{pmatrix}1&3\\3&1\end{pmatrix}.
\end{gathered}
\]
The matrix $A$ is entrywise nonnegative but indefinite. Both factors are strictly positive and unequal. With
\[
\begin{gathered}
K=\begin{pmatrix}1&-1\\-1&1\end{pmatrix},\\
G=XY^\top-A=\tfrac38K,\quad D=X-Y=\tfrac12K,
\end{gathered}
\]
direct multiplication gives $GY+\gamma D=0$ and $G^\top X-\gamma D=0$. Thus the pair is a joint critical point. Because each block objective is $\gamma$-strongly convex, exact AM initialized at $X_0=Y_0=Y$ reaches this unequal pair after one sweep and remains there.

Its objective is $15/32$. In contrast, for $q=(1,-1)^\top/\sqrt2$, $q^\top Aq=-1$ and
\[
\tfrac12\|WW^\top-A\|_F^2
\ge\tfrac12(q^\top WW^\top q+1)^2\ge\tfrac12
\]
for every nonnegative $W$. Consequently no equal-factor pair is globally optimal in this instance. Hu et al.~\cite{hu2026abmf} give the corresponding examples for hierarchical AM and SymHALS.

This instance rules out a guarantee for all entrywise nonnegative targets and all $\gamma>0$. The negative eigenvalue of $A$ distinguishes it from the PSD targets in Theorem~\ref{thm:hzero-sharp}.

\end{document}